\documentclass[
12pt]{report}
\usepackage[utf8]{inputenc} 
\usepackage[T1]{fontenc} 
\usepackage{graphicx}
\usepackage{setspace}
\usepackage{mathptmx} 
\usepackage[backend=biber,style=numeric,sorting=none]{biblatex} 

\usepackage[autostyle=true]{csquotes} 
\usepackage{lipsum}
\usepackage{array}
\usepackage{tabularx}
\usepackage{amsmath}
\usepackage{amsthm}
\usepackage[bottom]{footmisc}
\usepackage{geometry}
\usepackage[utf8]{inputenc}
\usepackage[english]{babel}
\usepackage{floatrow}
\usepackage{caption,subcaption}
\usepackage{multirow}
\usepackage{algorithm2e}
\usepackage{wrapfig}
\usepackage{upgreek}
\usepackage{enumitem,amssymb}
\newlist{todolist}{itemize}{2}
\setlist[todolist]{label=$\square$}

\addto\captionsenglish{
	\renewcommand{\contentsname}%
	{TABLE OF CONTENTS}%
}

\begin{document}



\begin{titlepage}
\begin{center}

\includegraphics[width=2.5cm, height=2.5cm]{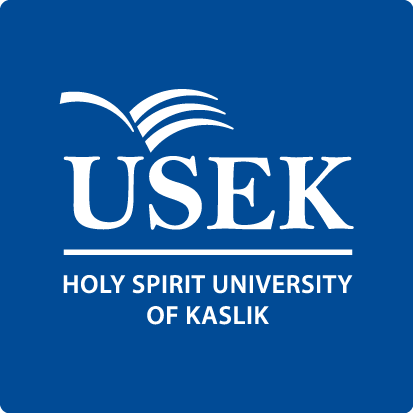} 
\\
\vspace*{0.7cm}

{\huge \bfseries\MakeUppercase{3D Labeling Tool} \par}\vspace{0.4cm}


\vspace{2cm}

by\\[0.2cm]
John \textsc{Rachwan}\\
Charbel \textsc{Zlaket}

\vspace{2cm}
Submitted to the\\
Faculty of Engineering

\vspace{2cm}
In partial fulfillment of the requirement\\
For the degree of Bachelor\\
of Engineering - Computer\\
Engineering\\
at the\\
Holy Spirit University of Kaslik (USEK)


\vspace{4cm}



Kaslik, Lebanon 

\end{center}
\end{titlepage}
\addtocounter{page}{1}
\clearpage 
\begingroup
\null
\newpage
\endgroup
\begin{center}
{\bfseries\MakeUppercase{3D Labeling Tool} \par}\vspace{1.5cm}

by\\[0.2cm]
John \textsc{Rachwan}\\
Charbel \textsc{Zlaket}

\vspace{3cm}
\end{center}
\textbf{\large\textsc{Approval:}}\vspace{0.5cm}\\
Prof. Joseph ZALAKET\\
\hspace*{\fill}{\rule[0.01em]{15em}{0.5pt}}\\
Advisor / Supervisor\\[0.2cm]
Associate Dean \hspace{11.2em}[Signature]\\
\vspace{2.5cm}\\
Dr. Charles YAACOUB\\
\hspace*{\fill}{\rule[0.01em]{15em}{0.5pt}}\\
Internal / External Examiner\\[0.2cm]
Associate Professor \hspace{9.5em}[Signature]
\vspace{1.5cm}\\
Dr. Adib AKL\\
\hspace*{\fill}{\rule[0.01em]{15em}{0.5pt}}\\
Internal / External Examiner\\[0.2cm]
Assistant Professor \hspace{9.7em}[Signature]\\
\vfill
\noindent
Date Defended : July 29, 2019
\newpage
\begin{center}
	\Large\bfseries{Plagiarism Statement}
\end{center}
	\setstretch{1.5}
	\addcontentsline{toc}{section}{\MakeUppercase{Plagiarism Statement}} 
	\hspace{1.1em} I confirm that this final year project is my own work, is not copied from any other person's work (published or unpublished) and has not been previously submitted for assessment either at the Holy Spirit University of Kaslik (USEK) or elsewhere.

	I confirm that I have read and understood the Academic Integrity regulations on plagiarism in the \textit{Academic Rules} and \textit{Student Life Handbook}.\\[1cm]
	\rule{\linewidth}{0.5pt}\\
	\noindent Signature\hfill{Date}
	\clearpage

\begin{center}
	\setstretch{1}
	\large\textbf{Final Year Project}\\
	\large\textbf{Release Form}
\end{center}
I, the under signed, hereby submit this final year project to the Holy Spirit University of Kaslik (USEK)\\
as partial fulfillment of the requirements for a Bachelor of Engineering’s degree.
\begin{todolist}
	\item By signing and submitting this agreement: I grant USEK Library the right to

	(a) reproduce electronic copies of my final year project for the purpose of preservation

	(b) include  the  scholarly  material  final year project in  the  archives  and  digital repositories at the Holy Spirit University of Kaslik “USEK Digital Gate”

	(c) release my final year project or dissertation to ProQuest/UMI

	(d) keep more than one copy of my final year project or dissertation for purposes of security and backup

	(e) reproduce, publicly display, and distribute the material to users world-wide at no cost for academic purposes

	\item I should inform the Main Library at the University if I intend to publish or post my final year project before the transaction is completed and listing the University as my affiliation.
	\item I request an embargo of this final year project for …………. months (maximum of 36 months) from the date of submission of the final year project.
\end{todolist}
\noindent
USEK will clearly identify your name as the author or owner of the final year project, and will not make any alteration, other than as allowed by this license, to your submission.\\[1cm]
\rule{\linewidth}{0.5pt}\\
\noindent Signature\hfill{Date}
\clearpage
\begin{center}
	\Large\bfseries\MakeUppercase{Acknowledgment}
\end{center}
\setstretch{1.5}
\addcontentsline{toc}{section}{\MakeUppercase{Acknowledgment}} 
\noindent
We would like to express our immense gratitude for all the people who supported us during the  period of this project. Beginning with BMW Group for providing us with the necessary hardware and connections to make this project possible.\\  Specifically, we thank Mr. Norman Muller, our supervisor at BMW Group for his continued support and hands on contribution to the project.  We would also like to thank our lab members Miss Joyce Abi Saleh and Mr. Hadi Ayoub for their assistance in certain parts of the project that matched their expertise. Additionally, we thank Logitech’s virtual reality team for providing us with their prototype pen.\\
On a special note, we address our warmest thanks towards our supervisor Prof. Joseph ZALAKET and the head of department Dr. Tilda KARKOUR AKIKI for giving us the opportunity to work with BMW Group. We are also thankful for their continuous support, motivation and assistance during the period of this project.
\clearpage

\begin{center}
	\Large\bfseries\MakeUppercase{Abstract}
\end{center}
\setstretch{2}
\addcontentsline{toc}{section}{\MakeUppercase{Abstract}}
Training and testing supervised object detection models require a large collection of images with ground truth labels. Labels define object classes in the image, as well as their locations, shape, and possibly other information such as pose. The labeling process has proven extremely time consuming, even with the presence of manpower. We introduce a novel labeling tool for 2D images as well as 3D triangular meshes: 3D Labeling Tool (3DLT). This is a standalone, feature-heavy and cross-platform software that does not require installation and can run on Windows, macOS and Linux-based distributions. Instead of labeling the same object on every image separately like current tools, we use depth information to reconstruct a triangular mesh from said images and label the object only once on the aforementioned mesh. We use registration to simplify 3D labeling, outlier detection to improve 2D bounding box calculation and surface reconstruction to expand labeling possibility to large point clouds. Our tool is tested against state of the art methods and it greatly surpasses them in terms of speed while preserving accuracy and ease of use.\\

\vspace*{\fill}
\noindent
\textbf{Keywords:}
Object Detection – Image Labeling – Mesh Labeling – Registration – Outlier Detection – Surface Reconstruction

\clearpage

\begin{center}
	\Large\bfseries\MakeUppercase{Resume}
\end{center}
\setstretch{2}
La formation et la mise à l'essai de modèles de détection d'objets supervisés nécessitent une grande collection d'images avec des étiquettes de vérité (groundtruth). Les étiquettes définissent les classes d'objets dans l'image, ainsi que leur emplacement, leur forme et probablement d'autres informations telles que la pose. Le processus d'étiquetage s'est avéré extrêmement long, même en présence de main-d'œuvre. Nous introduisons un nouvel outil d'étiquetage pour les images 2D ainsi que les maillages triangulaires 3D : 3D Labeling Tool (3DLT). Il s'agit d'un logiciel autonome, riche en fonctionnalités et multiplateforme qui ne nécessite pas d'installation et qui peut fonctionner sous Windows, MacOS et Linux. Au lieu d'étiqueter le même objet sur chaque image séparément comme les outils actuels, nous utilisons les informations de profondeur pour reconstruire un maillage triangulaire à partir desdites images et étiqueter l'objet une seule fois sur ledit maillage. Nous utilisons le concept d’enregistrement pour simplifier l'étiquetage 3D, la détection des valeurs aberrantes pour améliorer le calcul des rectangles de délimitation 2D et la reconstruction de surface pour étendre les possibilités d'étiquetage aux gros nuages de points. Notre outil est comparé à des méthodes de pointe et il les surpasse largement en termes de vitesse tout en conservant la précision et la facilité d'utilisation.

\vspace*{\fill}
\noindent
\textbf{Mots-cl\'{e}s:}
Détection d'Objets – Étiquetage des Images – Étiquetage de Maille – Enregistrement – Détection des Valeurs Aberrantes – Reconstruction de Surface

\clearpage

\setstretch{1.5}
\addcontentsline{toc}{section}{\MakeUppercase{Table of Contents}}
\tableofcontents

\clearpage

\listoffigures
\addcontentsline{toc}{section}{List of Figures}
\clearpage
\listoftables
\addcontentsline{toc}{section}{List of Tables}
\clearpage
\begin{center}
	\Large\bfseries{\MakeUppercase{Acronyms}}
\end{center}

\addcontentsline{toc}{section}{\MakeUppercase{Acronyms}}
\noindent
\textbf{$L_2$} : Euclidean distance\\
\textbf{$L_2E$} : Euclidean distance Estimation\\
\textbf{3DLT} : 3D Labeling Tool\\
\textbf{AABB} : Axis-Aligned Bounding Box\\
\textbf{ANN} : Artificial Neural Networks\\
\textbf{API}: Application Program Interface\\
\textbf{AUC} : Area Under Curve\\
\textbf{BMW} : Bavarian Motor Works\\
\textbf{CBLOF} : Cluster-Based Local Outlier Factor\\
\textbf{CLI} : Command Line Interface\\
\textbf{CPPSR} : Color-Based Point Set Registration\\
\textbf{CPU}: Central Processing Unit\\
\textbf{DLD} : Dependent landmark drift \\
\textbf{DoF} : Degrees of Freedom\\
\textbf{EM} : Expectation Maximization\\
\textbf{FFT} : Fast Fourrier Transform\\
\textbf{GAN} : Generative Adversarial Neural Network\\
\textbf{GLTP} : Global-Local Topology Preservation\\
\textbf{GMM} : Gaussian Mixture Model\\
\textbf{GPU}: Graphics Processing Unit\\
\textbf{GUI} : Graphical User Interface\\
\textbf{HBOS} : Histogram-Based Outlier Score\\
\textbf{ICP} : Iterative Closest Point\\
\textbf{IPDA} : Point Clouds Registration with Probabilistic Data Association\\
\textbf{IR} : Infrared\\
\textbf{JSON} : JavaScript Object Notation\\
\textbf{K-NN} : K Nearest Neighbour\\
\textbf{LLE} : Local Linear Embedding\\
\textbf{LOCI} : Local Correlation Integral\\
\textbf{LOF} : Local Outlier Factor\\
\textbf{LRD} : Local Reachability Density\\
\textbf{LU} : Lower-Upper\\
\textbf{LiDAR} : Laser Imaging, Detection And Ranging\\
\textbf{LoOP} : Local Outlier Probability\\
\textbf{MAP} : Maximum a Priori\\
\textbf{MBB} : Minimum Bounding Box\\
\textbf{ML} : Maximum Likelihood\\
\textbf{MSD} : Micro- structure descriptor\\
\textbf{OCSVM} : One-Class Support Vector Machine\\
\textbf{OS} : Operation System\\
\textbf{PCA} : Principal Component Analysis\\
\textbf{PNG}: Portable Network Graphics\\
\textbf{RAM}: Random-Access Memory\\
\textbf{RBF} : Radial Basis Function\\
\textbf{RGBA} : Red Green Blue Alpha\\
\textbf{RMS} : Root Mean Square\\
\textbf{ROC} : Receiver Operating Characteristic\\
\textbf{RPM-VFC} : Robust Point Matching via Vector Field Consensus\\
\textbf{RPM} : Robust Point Matching\\
\textbf{SDK}: Software Development Kit\\
\textbf{SLAM} : Simultaneous Localization and Mapping\\
\textbf{SSD} : Smooth Signed Distance\\
\textbf{SVGM} : Support Vector-Parametrized Gaussian Mixture\\
\textbf{SVM} : Support Vector Machine\\
\textbf{SVR} : Support Vector Registration\\
\textbf{TOF} : Time of Flight\\
\textbf{TPS} : Thin Plate Splines\\
\textbf{VM} : Virtual Machine\\
\textbf{VR} : Virtual Reality\\
\textbf{XML} : Extensible Markup Language\\
\textbf{rPCA} : robust Principal Component Analysis

\clearpage
\pagenumbering{arabic}

\chapter{Introduction}
\label{Chapter1}
	Manual annotation of digital images is a fundamental processing stage in many research projects and industrial applications. It requires human annotators to define and describe spatial regions associated with an image. Spatial regions are defined using standard axis-aligned minimum bounding rectangles and are described using textual metadata. A manual labeling tool permits human annotators to define these spatial regions around a desired object. Current state of the art tools have made the labeling process a very long and tedious affair, one that delays development of industrial applications and research in object detection. For instance, the process of preparing the necessary training and testing data in order to train a model to detect a given object would span several days.\\
	Current annotation tools are based on the same core concept. The annotator observes a digital image on the screen, defines a special region around a certain object and gives it a textual description. Every tool may implement this in a slightly different way, or may add shortcuts that make the process slightly faster. However, in all of them, the user has to label this object in every image it is present in. In this report, we present a novel labeling tool -- 3D Labeling Tool (3DLT) -- that tackles this problem in a different manner. We use the depth information of a sequence of RGB images to compute the camera trajectory and a sparse 3D mesh reconstruction. Then, labeling a certain object in all of the images it is present in is equivalent to labeling this individual object on the mesh. This is possible because the object present in this mesh is reconstructed from the RGB images we wish to label.\\
	Our tool is a standalone, feature-heavy and cross platform software that does not require installation and can run on Windows, macOS and Linux-based distributions. This is possible because it is developed using Unity; a real-time game engine that supports multiple platforms and contains all the necessary camera features that complement the development of this project, e.g emulating real world cameras' behavior. To handle complex computations such as in Chapters 5, 6, 7 and 9, we created a Python server. Communication between the Unity side and the Python side is done through Websockets, see Section \ref{websockets}.\\
	The data that is going to be labeled using our tool is collected by an RGBD camera that generates corresponding RGB and depth images of the scanned environment. Using this information and Open3D's reconstruction library, a 3D mesh is generated and encoded in PLY format along with a point cloud of each shot (depending on the user's needs). The RGB-D images, the mesh, the point clouds and the camera's information are all stored in a custom format called IRIS format that can be later on imported to the 3D labeling tool. Capturing depth information is now accessible for all users especially with the presence of commodity sensors such as Intel's RealSense camera. More on that in Chapter 3.\\
	3DLT holds a user interface that decomposes the workflow of labeling into two parts: importing the data set and then labeling it. This user interface is intive and aesthetic, also it isolates the complex back-end functionalities of the tool from the end user. The development of this user interface is examined in Chapter 4.\\
	Once the labeling object is placed on the mesh, we need to calculate the equivalent axis-aligned bounding rectangle on the 2D images. First, the labeling object has to be projected on the images; this is done by using single view geometry and the camera's projection matrix. A minimum bounding rectangle can then be calculated by getting the minima and maxima of the pixels on both the Y and Y axes. In some special cases, the object can be hidden by another object in one image shot but clearly visible in another. Since we do not want this blocking object to be present in the minimum bounding rectangle we make use of a fast outlier detection method to remove labeling object pixels that are very far from the others. The bounding box calculation and subsequent improvements are depicted in Chapter 5.
	\\
	Even though it takes one labeling action to label an object in all of its images, the labeling action in 3D space is not as simple as drawing a rectangle on a 2D image plane. A labeling object has to be instantiated and then translated, rotated and scaled by the user so that it can label an object on the mesh. To make this process trivial, we decided to use nonrigid registration to calculate the perfect transformation parameters (translation, rotation and scaling) that place the labeling object exactly in the position we want on the mesh. All the user has to do is define a rectangular cuboid on the mesh by selecting 4 points (2 for each dimension), we then register 4 points on a labeling object to the 4 points chosen by the user. In doing this, we shorten the time taken to label a certain object on a mesh to only a few seconds. We discuss this approach in detail in Chapter 6.\\
	For some special cases where the reconstructed mesh is unavailable or very sparse, we have developed an alternative way of labeling the object using the 2D images while retaining the speed and accuracy described in the previous paragraph, given that we know the exact size of the object we wish to label. This is done by reducing the problem to a system of non-linear equations whose solution would reveal the exact position where the labeling object should be placed. The solution to this problem is discussed in Chapter 7.\\
	Due to the basis of the tool being labeling in 3D, it is only natural to assume that it should support the possibility to label a provided 3D point cloud. This can be useful for object detection applications based on the revolutionary point cloud based neural network PointNet or for normal object detection models by extracting images from these point clouds. Obviously, capturing images from a surface-less point cloud would be useless as the machine learning models need to know what the object looks like in the real world. Therefore, we use surface reconstruction to extract a high quality triangular mesh from the point cloud and use Unity to navigate around the mesh and extract RGBD images to complete the tool's required input. A comprehensive comparison of surface reconstruction techniques, a designed surface reconstruction workflow and a description of the tool used to take the RGBD images are detailed in Chapter 8.\\
	In some rare cases\footnote{The object has very odd shape, missing depth in mesh, the object if very small etc.}, placing the 4 points discussed earlier may be non trivial. Henceforth, we introduced a new feature that snaps the labeling element in its convenient place if	it is more or less close to it and it has the same size. We do this by registering the points of the labeling object viewed by the camera to the points of the mesh viewed by the same camera. For this task, we selected chose an algorithm that is robust against outliers, noise and missing data. Especially for correspondence calculation. The feature and its implementation are discussed in Chapter 9.\\
	Finally, we added the ability to label in virtual reality using Unity's SteamVR plugin and in collaboration with Logitech's VR development team. This idea adds an immersive experience and exploits the 3D aspect of the tool to is farthest limits. It will be explained thoroughly in Chapter 10.\\
	The rest of the report is described as follows. In Chapter 11, we perform a series of tests to compare our proposed tool to current state of the art competitors. Finally, we sum up our work and propose interesting future improvements in Chapter 12.

\chapter{Related Work}
	There are two approaches in computer vision literature that associate textual information with images: annotation and categorization. Annotation is associated with keywords or detailed text descriptions in images, while categorization assigns each image to one of a set of preset categories \cite{Chen:2004:ICL:1005332.1016789}. This can be more common in two categories, for example indoor / outdoor or city / landscape, to more specific categories, such as villages in Africa, dinosaurs, mode ships and fight ships \cite{Chen:2004:ICL:1005332.1016789}. Categorization can be used to guide further image processing as an initial step in image learning. For instance, as a pre-processing step, a categorization into textured / non-textured classes is done in \cite{955109}. Recognition is about identifying specific instances of objects. Annotation-based Object recognition would distinguish between images of two structurally distinct cups \cite{Csurka2004} and category-level object recognition would put them in the same class. This report focuses solely on image annotation.
	\section{Manual Annotation}
	The best method to create ground truth for algorithm evaluation is to first create the required keyword vocabulary and then manually annotate the images using these keywords. This is done by placing a polygon \footnote{usually a rectangle} around the desired object and giving it a descriptive keyword or phrase. This textual description is usually from a dataset of words which the annotator can choose from. The manual annotation of images is a very time-consuming and labour-intensive task. In consequence, most comprehensively annotated datasets contain few images, while more "lightly" annotated datasets are large. One example of the former is the Sowerby database \cite{791401} containing 250 images with manually corrected segmentations and a keyword assigned to each segmentation region. The pictures are all rural or urban outdoor scenes, and this dataset is limited to an 85 word vocabulary. A larger set of manually labeled segmented images was presented in \cite{Barnard2008}: the regions were labeled on 1014 manually segmented images.
	\subsection{LabelImg}
	LabelImg \cite{labelimg} is a simple offline annotation tool that allows the user to import multiple images and to draw bounding rectangles around the desired object. It also lets him describe the drawn box by a textual keyword. Finally, it exports all of the bounding boxes in an XML file.
	\subsection{Labelbox}
	Similarly to LabelImg, Labelbox \cite{labelbox} is a very simple annotation tool that gives users the capability of drawing bounding rectangles with textual description around the object. The only difference is that Labelbox is an online tool that can be accessed by any device with internet browser support. However, one downside to Labelbox is that the user relies on the the tool's server which may slow down labeling time in case of excessive user activity.
	\subsection{Anno-Mage}
	Anno-Mage \cite{annomage} offers the same functionality as LabelImg, but it also possesses RetinaNet as a suggesting algorithm. It suggests 80 class objects from the MS COCO dataset \cite{DBLP:journals/corr/LinMBHPRDZ14}. This allows the user to receive suggested bounding rectangles if any object pertaining to one of these 80 classes is present in the images. In these situations, Anno-Mage speeds up the labeling process drastically.
	\section{Intensive Group Annotation}
	Different approaches and systems have been set up to simplify image annotation by receiving input from a large number of people. The simplest way is to get a group of people together to create the annotations— the PASCAL VOC challenge \cite{Everingham2010} organizes an annual "annotation party" in which a group intensively annotates over 3–4 days. In \cite{Everingham2010}, it is proven that this is more effective than distributed asynchronous annotation.
	\section{Collaborative Annotation Over the Internet}
	LabelMe \cite{Russell2008} is an online annotation tool designed to collect keywords that describe image regions for evaluation of object recognition. The user defines a polygon around an object and then enters a keyword that describes the object. Misspelled keywords often occur because the vocabulary is not controlled. This issue is solved through a verification step by the database administrators.

\chapter{Data Set Importer}
\label{chapter2}
	\section{Introduction}
	To start the labeling process, a data set should be imported in a custom format that contains a collection of RGB and depth images, a mesh, the point clouds and finally the camera's intrinsic and extrinsic for each shot (More on that in Chapter 5).\\
	In this chapter, we will examine this custom format and how the data is being collected and processed. Then, we will shed the light on the PLY file extension in which the mesh and the point clouds are encoded with. We end by illustrating the development of a PLY run-time importer in Unity.
	
	\section{Data Collection}
	Before importing the data in the 3D labeling tool, it is essential that it is collected, processed and encoded into a specific format to be able to receive accurate results when labeling.
	
	\subsection{Camera Specifications}
	The data is collected using an Intel RealSense D435 depth camera also known as RGBD camera. It has a frame rate of up to 90 fps, a depth resolution of $1280\times720$ and an RGB resolution of $1920\times1080$. To set up and run the camera, we used \textit{Librealsense} which is Intel's RealSense python library~\cite{librealsense}. In addition, Matlab's Camera Calibrator app was used to extract the camera's intrinsic parameters that are saved for later use.
	
	\subsection{Recording and Processing}
	When the camera starts to record, the depth and color stream are both enabled at a resolution of $1280\times720$ and a frame rate of 30fps. Note that it is important to set both the depth and color on the same frame rate so that both images match. Each frame is saved separately as a depth and RGB image and a timestamp is created for each image (depth and RGB) along with its name.\\
	Moreover, after the data is collected, blurry frames are eliminated and new depth and RGB text files (timestamps with names of the images) are generated without the blurry data.
	
	\subsection{Camera Trajectory}
	To compute the camera's trajectory, ORB-SLAM 2~\cite{orbslam} was used. It is a real-time SLAM library for RGB-D cameras and requires the RGB and depth images along with an association file that contains the RGB and depth text files. ORB-SLAM iterates over all the images in the sequence and generates the camera trajectory file that contains the camera's transformation (position and rotation) along with the frame's timestamp. After the trajectory is computed, the system iterates over every camera's position and calculates the difference between frames; if the frames are too close to each other, the second one is eliminated. This process is repeated for every frame and finally a camera trajectory file is generated.
	
	\subsection{Mesh Construction}
\begin{figure}[!hbt]
		\centering
		\includegraphics[width=\linewidth]{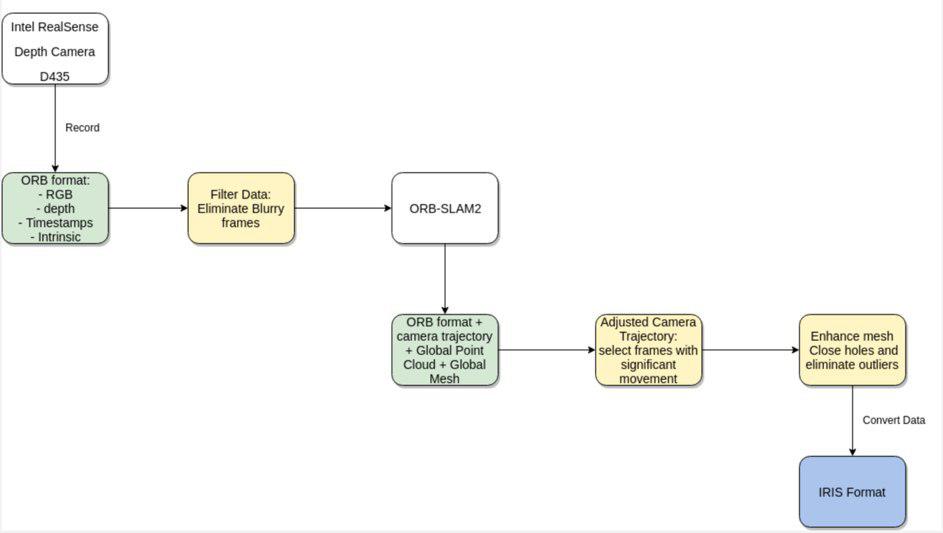}
		\caption{Data Collection Workflow Diagram}
		\label{fig:datacollectionworkflow}
	\end{figure}

	To build the mesh, the system uses Open3D reconstruction method~\cite{meshreconstruction} that uses the camera's intrinsic parameters and the depth scale. This function iterates over all the RGB and depth images, creates RGBD images and integrates volume to them. As a result, the mesh is extracted from the volume.\\
	Please refer to Figure~\ref{fig:datacollectionworkflow} for a diagram presentation of the data collection and processing workflow.
	
	\subsection{IRIS Format}
\begin{figure}[!hbt]
		\centering
		\includegraphics[scale=1]{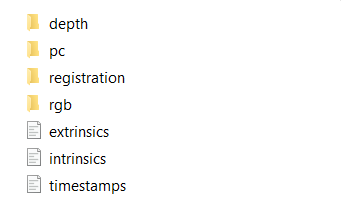}
		\caption{IRIS Format Example}
		\label{fig:irisformatex}
	\end{figure}

	After collecting the needed data, we convert it into a custom format called IRIS that is acceptable by the 3D labeling tool. In this format, the RGB and depth images are renamed and numbered from 0 onward. At this stage we should have obtained two folders, one contains the RGB images and the second one contains the depth images. In addition, this format consists of a ``timestamps.json" file that provides the correspondence between the old timestamps and the new image names. It also contains an ``intrinsics.json" file that holds information about the calibration of the camera. Also, this format contains a file called ``extrinsics.json" that is basically the camera trajectory file produced by ORB-SLAM.\\
	Finally, the constructed mesh is encoded in a PLY format and saved in a folder called ``registration" and the point clouds are saved also in the same format in a folder named ``pc". Figure~\ref{fig:irisformatex} shows an example of a data set in IRIS format.
	
	\section{PLY File Importer}
	To import the RGB and depth images to Unity, we need a byte reader in C\# to parse the PNG images. However, for the mesh and point clouds, we had to develop a custom parser for the PLY file format, which will be explained thoroughly in this section.
	
	\subsection{Polygon File Format}
	The polygon file format(ply), also known as the Stanford Triangle Format, is used for storing a graphical collection of polygons, more specifically, a collection of polygons. The file extension has two representations: an ASCII representation and a binary version used for fast loading and saving.
	The PLY format defines the object as a group of vertices, faces and other elements (surface normals, texture coordinates, colors, etc.). However, it does not include transformation matrices, object subpart or modeling hierarchies. Therefore, this format is not a general scene description language.\\
	Moreover, this file extension is typically intended to represent the object by a list of “X, Y, Z” triples for the vertices and a list of indices. The latter indicates the sequence of the vertices that describes the face of a polygon. The vertices and faces are an example of “elements” in a PLY file, each element has a fixed number of “properties”, e.g. the “vertices” element can be associated with the properties “red, green and blue” that indicate the color of each vertex. In addition, one can create a new element type and define its corresponding properties, elements like “material, edges, etc.”. Note that added elements could be discarded by programs that don’t understand them.
	The PLY format is divided into two complementary sections: the header and the body. The start and end of the header are indicated respectively by the keywords: “ply” and “end\_header”. The header is a description that defines the remainder of the file. First, it states whether the file is binary or ASCII, then it defines the elements of that object by indicating the element’s name, the number of such element in the object and a list of properties associated with each element. These properties are defined by the keyword “property”, followed by the data type of the property, this list also defines the order in which the property appears for each element. Then, the body contains a list of elements for each element type, presented in the order described in the header.
	\begin{figure}[!hbt]
		\centering
		\includegraphics[width=0.5\linewidth]{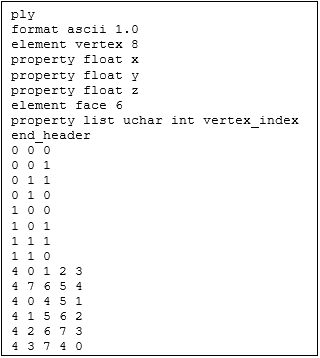}
		\caption{PLY File Example}
		\label{fig:plyexample}
	\end{figure}
	\noindent
	Figure~\ref{fig:plyexample} represents a basic example of a \textbf{cube} written in PLY format. This example illustrates a basic structure of the header, describing how the object is only a collection of vertices and faces (\textit{element vertex and element face}), and how each element has its own properties. Moreover, this example shows the two data types a property may have: scalar and list. The scalar is the data type of the vertex’s properties (\textit{float, int, uchar, etc.}). However, the property “vertex\_index” of the element “face” contains an unsigned char that indicates the number of indices the property contains, followed by a list containing that many integers. Each integer is an index to a vertex.
	
	\subsection{PLY Parser in Unity}
	In the process of developing the 3D labeling tool, a parser was needed to be able to import any PLY file at run-time. The idea was to give the user the ability to import any 3D object represented by a PLY file in a few clicks. This 3D object could be the global mesh or a point cloud that is generated from the RGBD images taken by the camera.
	First, the parser starts with the header of the PLY file and checks the number of the vertices and/or the faces. Then, for each element, the parser takes into consideration the data type of each property and stores it in a custom class that contains only the properties in the same order as given in the header of the file. This method allows the binary reader to parse the body of the PLY file in the same order of the header by considering the data type of each property.
	However, the import program must be designed to take into consideration meshes and point clouds. In fact, the only difference between a mesh and a point cloud is that point clouds don’t have faces, they are only constructed using vertices. So, to take this note into consideration, the parser checks in the header of the PLY file, if the latter contains only vertices or not, to be able to call the corresponding function to render the 3D object as a mesh or as a point cloud in Unity.
	
	\subsubsection{Import the 3D object as a Mesh}\label{meshimporting}
	
\begin{figure}[!hbt]
		\centering
		\includegraphics[width=\linewidth,height=0.24\textheight]{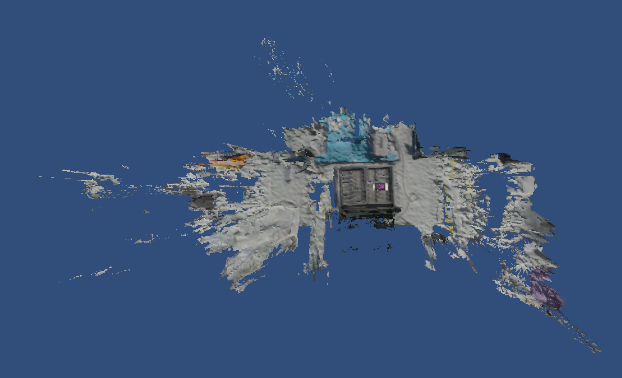}
		\caption{Imported mesh in the tool}
		\label{fig:importedmesh}
	\end{figure}

	After collecting the vertices and the faces of the 3D object, the mesh could be constructed by using Unity’s built-in mesh class that contains the functions that set the vertices of the mesh and the same for the faces. The triangles/faces are enough to define the basic shape of the object, but extra information that describe the detail of the object need to be added. In the application, there’s always a global light in the scene that hits all the objects including the imported mesh. Therefore, to allow the object to be shaded correctly for lighting, a normal vector should be defined for each vertex. During the shading calculations, each vertex normal is compared with the direction of the light: if both vectors are aligned then this point is receiving full light and it’s being shaded with full brightness. Typically, the light will arrive at an angle to the normal and the shading will be somewhere between full brightness and complete darkness, depending on the angle. However, not all imported 3D objects contain information about the normal of each vertex and a recalculation of the vertices' normals is done using the Mesh class function: \textit{Mesh.RecalculateNormals()}, after constructing the mesh. Finally, the 3D object is imported, and the mesh is rendered in the application as shown in Figure~\ref{fig:importedmesh}.
	
	\subsubsection{Import the 3D object as a Point Cloud}
	
\begin{figure}[!hbt]
		\centering
		\includegraphics[width=\linewidth,height=0.4\textheight]{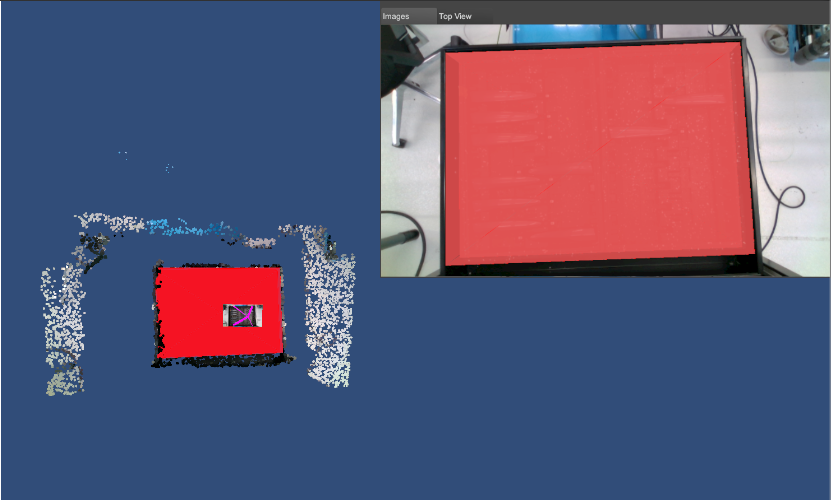}
		\caption{Point Cloud example with labeling object}
		\label{fig:pcwithlabobj}
	\end{figure}

	The input data set may contain a collection of point clouds generated from the RGB images using Open3D libraries as a pre-processing algorithm before handing the data as an input for the tool; therefore, each image has its own point cloud. Adding this type of object representation could enhance the labeling action for the user, especially if the data set does not contain a 3D mesh. In this case, the user takes advantage of the presence of the point clouds to ensure an accurate labeling of the object in all angles as shown in Figure~\ref{fig:pcwithlabobj}.
	\noindent
	Point clouds are a collection of independent spatial points, which are used to represent a 3D object. The point clouds that are being imported into the tool are stored in PLY format. However, Unity does not provide a built-in renderer for this type of objects. Thus, a method for this problem has been given in~\cite{pcx2019}  that implements two classes to ensure the rendering of the point clouds into the application. The class “PointCloudData” implements the constructor to store the data in an object which uses a “ComputeBuffer” to store the points. This latter is a class in Unity that supports GPU data buffer. As discussed in section \ref{meshimporting}, the same goes for the point clouds; the buffer is fed with an array of points, where each point contains two attributes: position and color. After defining the point cloud data, the author implements another class responsible for rendering the point cloud already stored in the point cloud data object. The class “PointCloudRenderer” is added as a component for the imported point cloud. The main idea behind this class is to create a material from a custom shader that is responsible for the mathematical calculations that computes the color of each pixel rendered based on the material configuration and the lighting input. Before drawing the points, the material sets the input for the shader calculation (points transform matrix, GPU instancing, colors and the point buffer). Finally, Unity’s “Graphics” class sends a draw call to the GPU, setting the attribute “MeshTopology” to “Points” because, the user is importing point clouds and not meshes with triangles in this case. Furthermore, the RGB images and the point clouds are coupled using their identifications.

	\chapter{User Interface}
	\label{chap:userinterface}
	\section{Introduction}
	An application's user interface ensures that the services of the tool are offered to the user in a clear way without ambiguity. In this chapter, we examine the developed workflow of the 3D labeling tool. Then we detail how we enhanced the user interface throughout the project to facilitate the labeling experience along with the usability of this application.
	
	\section{Workflow of the 3D Labeling Tool}\label{sec:workflow}
	When the application is launched, an import window pops up and is responsible for loading the IRIS data set that we mentioned in the previous chapter. This tool proposes as well a main labeling step that holds all the main functionalities that will be discussed thoroughly in what follows.
	
	\subsection{Import Window}\label{importwindow}
\begin{figure}[!hbt]
		\centering
		\includegraphics[width=\linewidth]{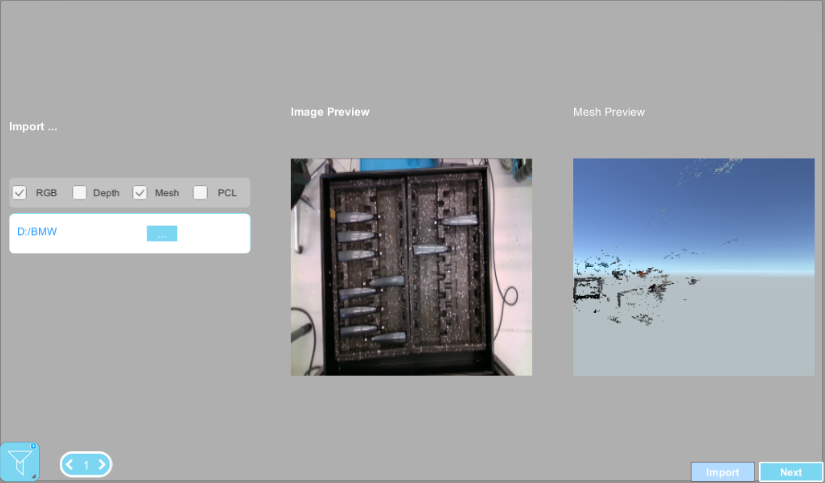}
		\caption{Import Window Example}
		\label{fig:importwindowexample}
	\end{figure}

	This window displays the first workflow phase of the tool where the user uploads the data set using a file browser, decides whether to include only the RGB data or adds the mesh and/or the point cloud data of the imported images. Moreover, the user has the capability to filter some of the images. This was done by including a slider to the import window such that its value is used as follows: if the key of the dictionary that contains the camera positions is a multiple of the slider value then the image is loaded. We The idea behind this filter is to decrease the load on the tool in case the user doesn't want to label the whole data set.\\
	Finally, two preview panels were added where the user can check if the imported data is the right one or not, hence giving him the opportunity to re-import another set before starting to label (see Figure~\ref{fig:importwindowexample}).
	
	\subsection{Main Labeling Window}\label{mainwindow}
\begin{figure}[!hbt]
		\centering
		\includegraphics[width=\linewidth]{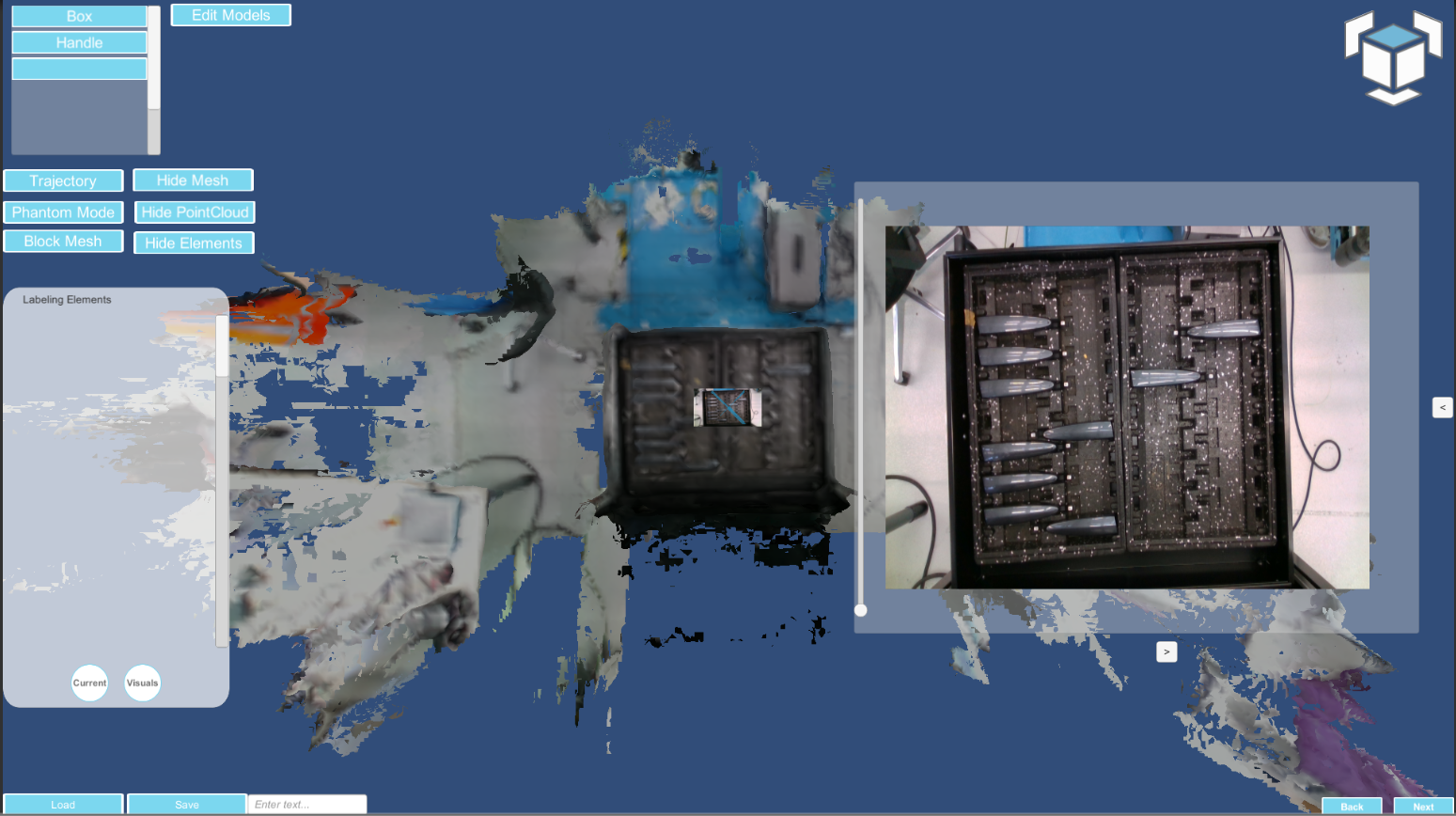}
		\caption{Main Labeling Window Example}
		\label{fig:mainlabelingwindow}
	\end{figure}

	The main functionalities of the tool lie in this window. The latter allows the user to add a labeling object (Primitive Cube or a CAD model). After that, the user carefully positions that object on top of the imported mesh while checking the images on the right side of the screen (see Figure~\ref{fig:mainlabelingwindow}) to verify that the object is in the right place.
	
	\subsubsection{Navigation}
	To ensure that the user can navigate with freedom in the scene and especially around the mesh, we added a camera control script and attached it to the main camera that renders the imported mesh. This script makes the arrow keys and the mouse input available for the user to navigate and orbit around the mesh. The user can also control the speed of the navigation.
	
	\subsubsection{The Labeling Object's Transformation}
\begin{figure}[h]
		\centering
		\includegraphics[width=\linewidth]{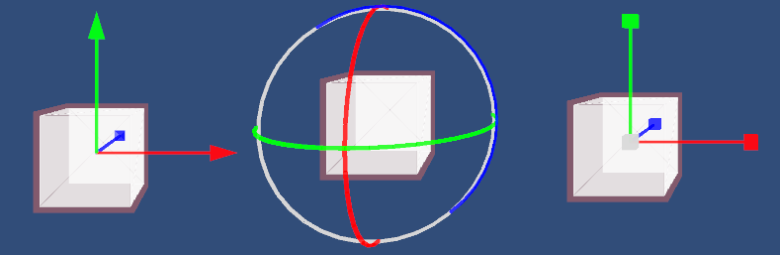}
		\caption{Run-Time Gizmo Tool}
		\label{fig:gizmo}
	\end{figure}

	One of the essential functionalities in this phase is the gizmo\footnote{Gizmos in Unity are used to give visual aids to move, rotate and scale the object.} tool that is responsible for changing the transformation of the labeling element at run-time. This is done with the help of this repository~\cite{gizmotool} that implements a run-time gizmo tool so when the user clicks on the labeling element they will have the option to change its position, rotation(see Figure~\ref{fig:gizmo}). Note that the gizmo is always placed according to the center of geometry of the labeling element.
	
	\subsubsection{Color Picker}
\begin{figure}[!hbt]
		\centering
		\includegraphics[scale=0.75]{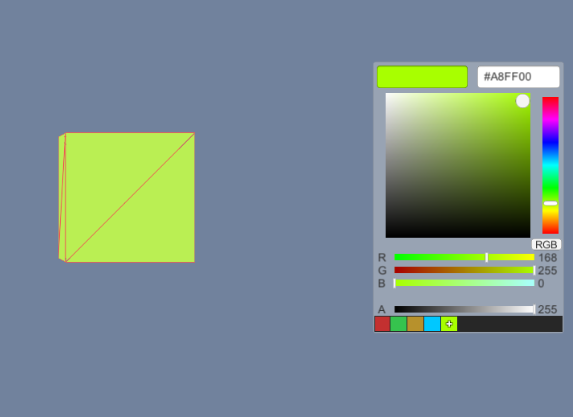}
		\caption{Color Picker}
		\label{fig:colorpicker}
	\end{figure}

	Changing the color of the labeling element is a key feature in this tool because it will affect the labeling mechanism that is discussed in the following chapter. To implement a color picker in the tool at run-time, we used the following open-source repository~\cite{colorpicker} in which the user clicks on the object and pops out a color picker panel (see Figure~\ref{fig:colorpicker}) to change the color and transparency of the object.
	
	\subsubsection{Extra Features}
	As shown in Figure~\ref{fig:mainlabelingwindow}, the main labeling window features many buttons that trigger the functionalities implemented in the tool:\\
	First, the user has the ability to hide/show the mesh, the point clouds and the labeling elements. Second, the trajectory of the camera can be also shown in the main window by clicking on the ``Trajectory" button in case the user is curious about how the images where taken (More on that feature in the next chapter). Also, if the user needs to block the mesh\footnote{Enabling the camera to project the labeling element before rendering the mesh. More on that in the next chapter} when labeling to ensure a better placement of the labeling element they can click on the ``Block Mesh" button. Finally, by clicking ``Phantom Mode", the user can execute a key feature that consists in transforming the labeling element onto its position on the mesh using a registration algorithm and that is thoroughly explained in Chapter 6.\\
	Moreover, when adding the labeling element, the user can enter the \textit{Edit Model} mode and alter the size of the object along with its color and save it for later use, in case this element is redundant in the environment where they are labeling. These elements are then showed on the top left corner of the screen to be later on instantiated.\\
	In case of a large mesh to label, the user has the option to save the session and load it back again to resume the labeling process later. In addition, the labeling action is triggered when the user hits the ``Current" button in the ``Labeling Elements" panel (Figure~\ref{fig:mainlabelingwindow}) where the labeling is triggered.\\
	Finally, after the user finishes the labeling, they can export the annotations by clicking on \textit{Next} button and exporting the files to the data set location as explained in details in the next chapter.
	
	\section{User Interface Enhancements}
	The version shown in Section~\ref{sec:workflow} showcases the main workflow of the tool with an old version of the application that was later on modified to make the user experience more intuitive and aesthetic. In this section, we are going to examine the major changes that were done to the user interface design and functionalities.
	
	\subsection{Import Window Enhancements}
\begin{figure}[!hbt]
		\centering
		\includegraphics[width=\linewidth]{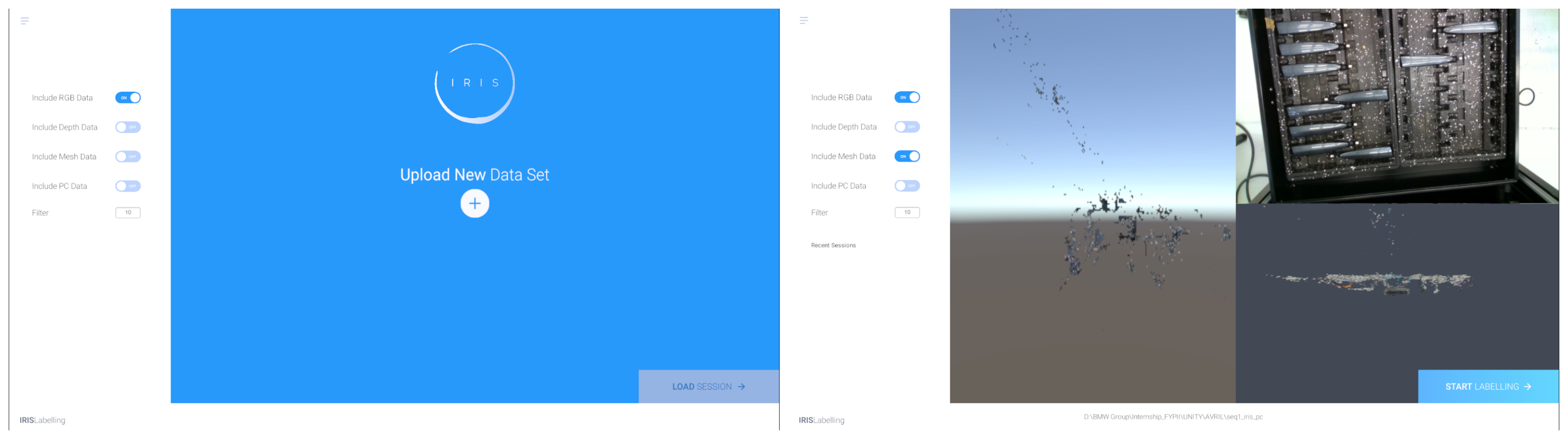}
		\caption{New Import Window Design}
		\label{fig:importwindow}
	\end{figure}

	The main idea behind the upgrade to the import window is transitioning to a project-based approach. For instance, after the import actions, the user can decide whether to start labeling on an already saved session of this data set or to start a new session. The main functionalities of Section~\ref{importwindow} were preserved, with the exception of the data set filter which is now a number rather than a slider. We also added the possibility of loading a recent session as shown in Figure~\ref{fig:importwindow}. In addition, a preview of the top view of the mesh is also added to the import window.

	\noindent
	If the user is satisfied with the preview of the data set, they can start the labeling process by clicking the ``Start Labeling" button on the bottom right corner of the screen. Notice that even before the session starts, the path of the data set is presented in the footer. This can help the user to check if the data path is right and give the opportunity to re-import a new data set.
	
	\subsection{Main Window Enhancements}
	Improving the user interface mainly focused on making things simpler in the main labeling window by removing the obsolete buttons or that did not have any effect on the labeling process. In the simplified interface, the user can chose either one or two ways to add a labeling element: (1) Add a simple primitive cube or (2) Load a CAD model from the inventory (Explained in details in Chapter 7).\\
	
	\subsubsection{Dynamic Panels}
\begin{figure}[!hbt]
		\centering
		\includegraphics[scale=1]{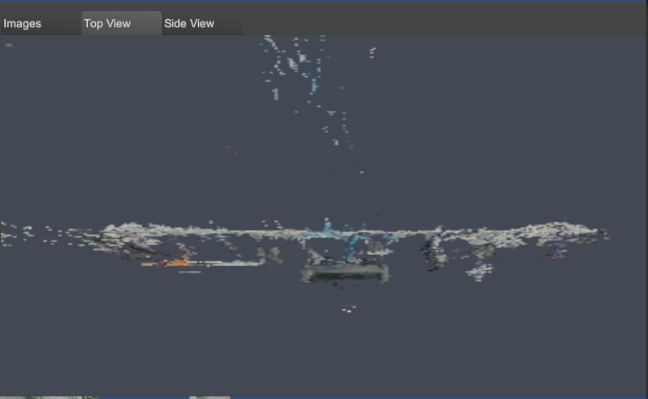}
		\caption{Mesh Top View Example}
		\label{fig:topview}
	\end{figure}

	We added a dynamic panel for the images that lets the user move and minimize it. We also added two extra panels that show the mesh's top and side view. This allows the user to keep track of every side of the mesh while labeling, ensuring the object's position is correct from all angles. In addition, the user can navigate and zoom on the top and side view panels (see Figure~\ref{fig:topview}).
	
	\subsubsection{Labeling Elements}
\begin{figure}[!hbt]
		\centering
		\includegraphics[scale=0.4]{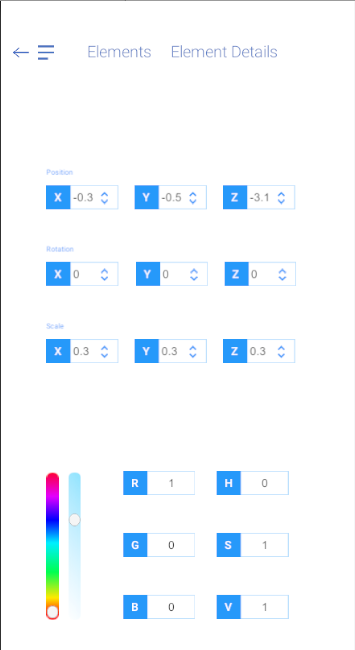}
		\caption{Element Details}
		\label{fig:elementsdetails}
	\end{figure}

\begin{figure}[!hbt]
		\centering
		\includegraphics[width=\linewidth]{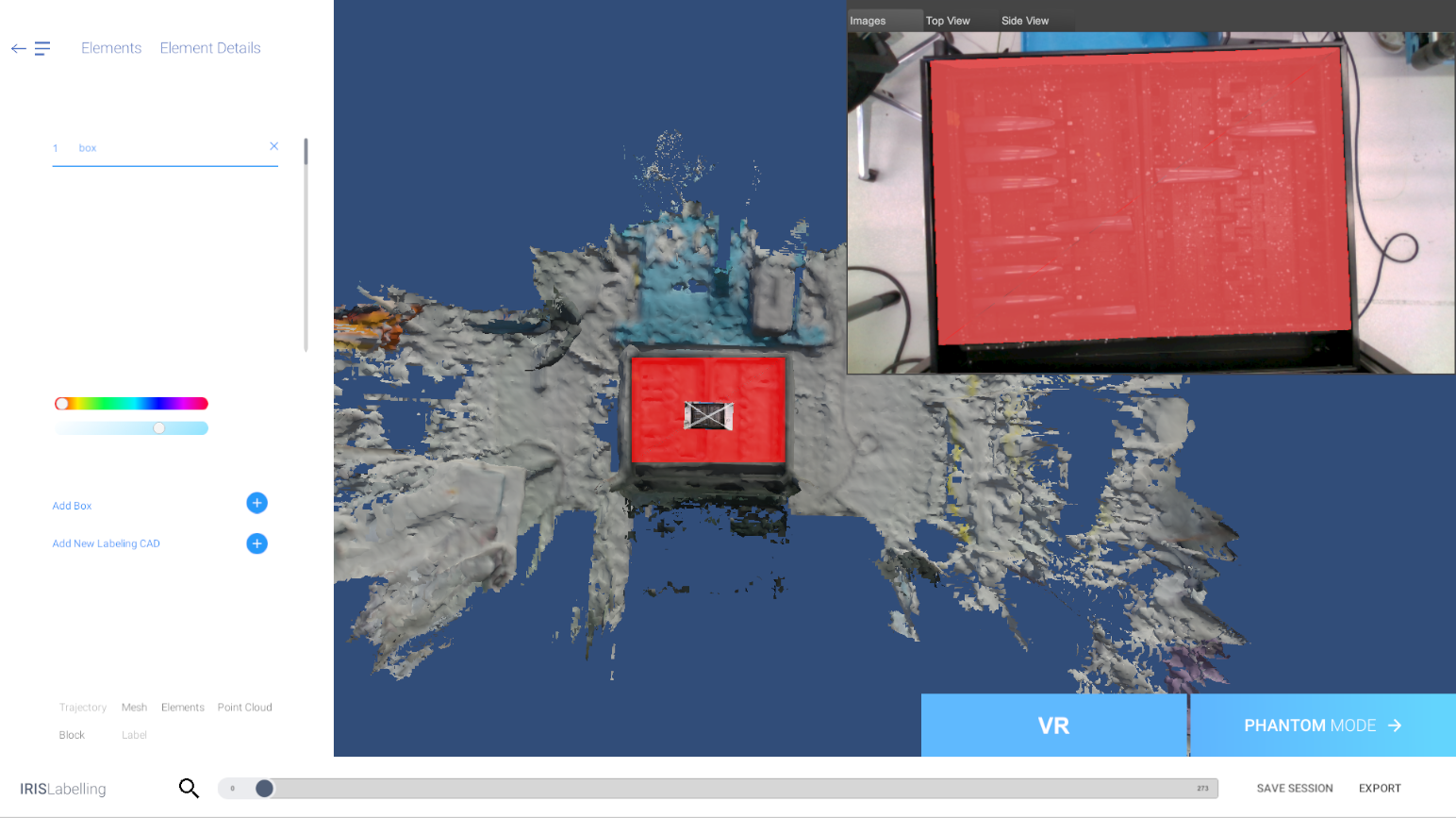}
		\caption{New Main Labeling Window Example}
		\label{fig:mainlabelingwin2}
	\end{figure}

	The labeling element's information is shown in the ``Elements Details" on the top left part of the screen (see Figure~\ref{fig:elementsdetails}). This panel showcases the transform\footnote{Information about the position, rotation and scale in Unity} of the labeling element along with detailed information on its color. This gives the user the ability to make fine adjustments to the object's color, position, rotation and scale.\\
	Furthermore, in case the user wants to identify a certain labeling element's position on the mesh, they can simply double click on its name in the ``Elements" panel on the left (see Figure~\ref{fig:mainlabelingwin2}) and the camera will directly change its position to point at that object and navigate the user to it. 
	
	\subsubsection{New Features}
	In this version of the tool, the user can now copy and paste the labeling elements to ease the labeling in case of redundant objects on the mesh. Also, the ``Save Session" button will write the positions of the placed labeling elements along with their color in a JSON file. This file is saved in the data set's folder.\\
	Moreover, the camera shot can also be identified in the scene by clicking the search button on the bottom left corner of the tool.\\
	The slider in the footer helps the user to navigate between the imported images. Also, the ``VR" button launches the labeling into virtual reality. This is explained and examined in details in Chapter 10.\\
	Finally, we managed to separate the user interface from the main and core functionalities by using C\#'s \textit{delegates} functions that manage events. In other words, a button calls an event that enables a functionality which is completely isolated from the user interface. As a result, this reinforces the scalability of our application since it allows a seamless addition of functionalities.
\chapter{Bounding Box}
\label{chap:boundingbox}
	\section{Minimum Bounding Box}
	Before we define the type of box used to annotate 2D images that train object detection models, we will define its more general counterpart; the minimum bounding box. In geometry, the box with the smallest measure (area, volume, or hypervolume in higher dimensions) within which all the points are located is the minimum bounding or enclosing box for a point set (S) in N dimensions.\\
	A point set's minimum bounding box is the same as its convex hull's minimum bounding box \footnote{The convex hullof a set X of points in the Euclidean plane or in a Euclidean space (or, more generally, in an affine space over the reals) is the smallest convex set that contains X.}, a fact that can be heuristically used to accelerate computation. The name "box"/"hyperrectangle" derives from its use in the Cartesian coordinate system, where it can be actually visualized as a rectangle (two-dimensional case), horizontal parallelepiped (three-dimensional case), etc. The arbitrarily oriented minimum bounding box is the minimum bounding box, calculated without any limitations regarding the result orientation.
	\section{Axis-Aligned Minimum Bounding Box}
	For a given point set, the axis-aligned minimum bounding box (AABB) is its minimum bounding box subject to the constraint that the box edges are parallel to the Cartesian coordinate axes.
	Axis-aligned minimal bounding boxes are used as an easy descriptor of the form of an object at an estimated place. For example, when it is necessary to find intersections in the set of objects in computational geometry and its applications, the initial check is the intersections between their MBBs. Because it is generally much less costly than checking the actual intersection (since it involves only coordinate comparisons), it rapidly enables the exclusion of distant pair checks.
	As a result, they are used in object detection neural networks to identify a specific object. They are also paired with a small text descriptor that identifies what the annotated object is.
	\section{Calculating the Minimum Bounding Rectangle}
	In this section, we will describe how the labeling object is being projected onto the images and how the minimum bounding rectangle is being calculated. It is done on the Python side to allow for complex mathematical calculations.
	\subsection{Projection of Labeling Object onto the Image}
	The data set that is being loaded in Unity contains two files that enclose details about the camera that captured the images. These files are the camera's extrinsics and intrinsics. However, to be able to mimic the real camera into a virtual one in Unity, we had to create a game object that describes a shot view which is formed of a camera and its image frame (children of the shot view game object).
	\begin{figure}[!hbt]
		\centering
		\includegraphics[width=0.8\linewidth]{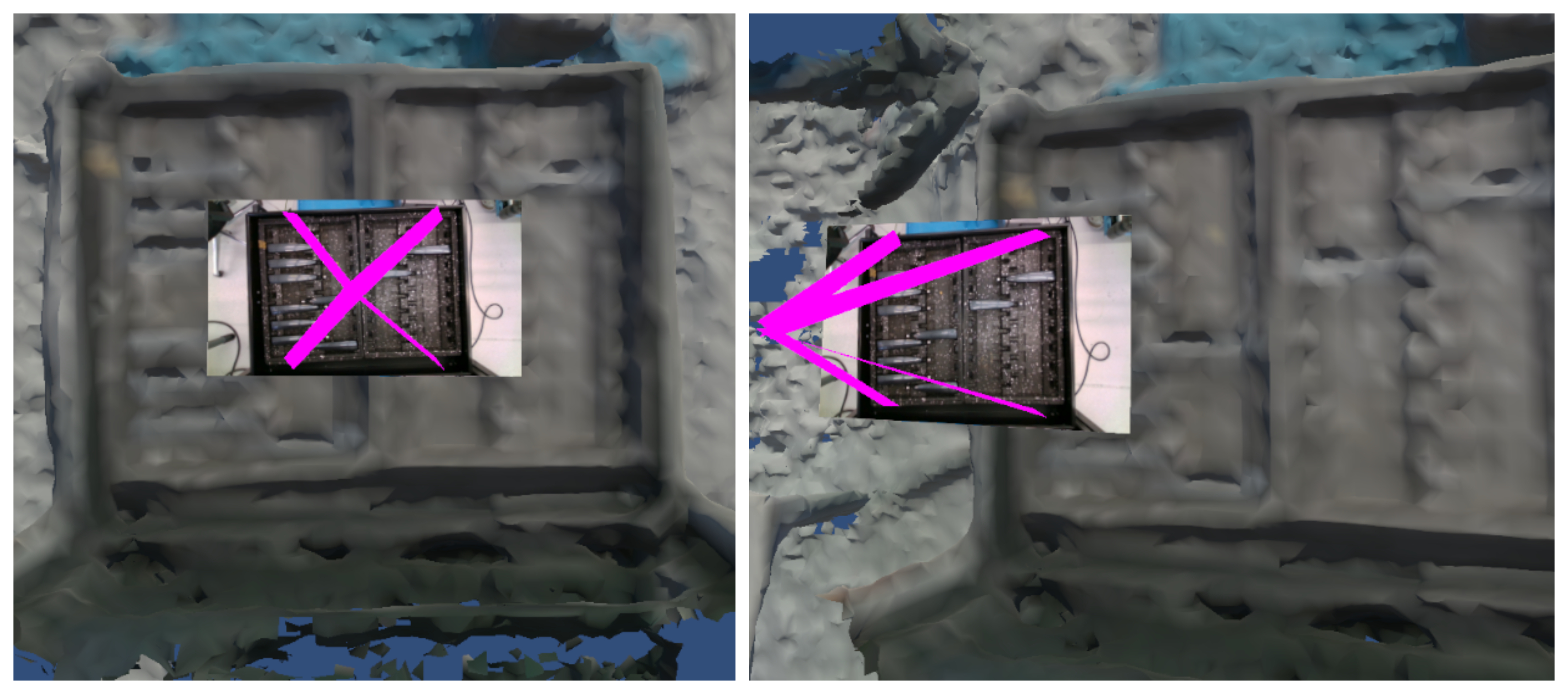}
		\caption{Shot View Example}
		\label{fig:shotView}
	\end{figure}
	
	\subsubsection{The Extrinsic Camera Matrix}
	The extrinsic matrix $[R|t]$ describes how the world transforms according to the camera. The inverse of the extrinsic matrix describes how the camera transforms according to the world origin, in other words it defines the camera's position in the world. In order to construct it, we start with a $3\times3$ rotation matrix on the left. A $3\times1$ translation column vector on the right, where $t=-RC$ such that C is a column vector describing the camera's location in world coordinates. An extra row of (0,0,0,1) is also added to make it a square matrix making the calculation of the inverse easy. The extrinsic matrix looks as follows:
	\begin{equation}
	\centering
	[R|t]=
	\quad
	\begin{bmatrix}
	r_{1,1} & r_{1,2} & r_{1,3} & t_{1} \\
	r_{2,1} & r_{2,2} & r_{2,3} & t_{2} \\
	r_{3,1} & r_{3,2} & r_{3,3} & t_{3} \\
	0 & 0 & 0 & 1
	\end{bmatrix}
	\quad
	\end{equation}
	
	\noindent
	We use the inverse of the camera's extrinsics to position the shot views.
	
	\subsubsection{The Intrinsic Camera Matrix}
	The intrinsic camera matrix $K$, is a $3\times3$ upper triangular matrix that transforms the 3D camera coordinates to 2D image coordinates:
	
	\begin{equation}
	\centering
	K=
	\quad
	\begin{bmatrix}
	f_{x} & s & x_{0}\\
	0 & f_{y} & y_{0} \\
	0 & 0 & 1
	\end{bmatrix}
	\quad
	\end{equation}
	
\begin{figure}[!hbt]
		\centering
		\includegraphics[scale=0.7]{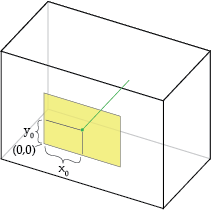}
		\caption{Principal Point Offset}
		\label{fig:pp_offset}
	\end{figure}

	\noindent
	$f_{x}$ and $f_{y}$ represent the components of the focal length of the camera, $s$ is the axis skew which is responsible for the shear distortion of the image and finally $x_{0}$ and $y_{0}$ are the principal point offset\footnote{The principal point offset is the location of the point relative to the frame's origin, which in our case, on the bottom-left of the image}, see Figure~\ref{fig:pp_offset}.
	 We use the camera's intrinsics to calibrate the placement of the image frame and obtain an accurate projection that corresponds to the real camera's image frame, see Figure~\ref{fig:shotView}.
	\subsubsection{Projection}
	The projection matrix is a $3\times4$ matrix which describes the mapping of a pinhole camera\footnote{A simple camera represented as a light-proof box with a small hole on one side. The light of the scene passes through the hole and projects an inverted image on the opposite side of the box.} from 3D points in world coordinates to 2D points in image coordinates.
	We multiply the intrinsic matrix and the extrinsic matrix to obtain the projection matrix $P$: 
	
	\begin{equation}\label{projectionmatrix}
		P = K[R|t]
	\end{equation}
	
	\noindent
	Let $w$ be a $4\times1$ column vector containing world coordinates of a certain point $[x\;y\;z\;1]$ and let $W$ be a $3\times1$ column vector containing the coordinates of the same point in image coordinates $[u\;v\;1]$, the equation transforming the former into the latter is:
	\begin{equation}\label{projectt}
	W = Pw 
	\end{equation}
	 Finally, P is assigned to the virtual camera's projection matrix property in Unity so that this camera can render according to the projection matrix. As a result the 3D labeling elements appear in the corresponding position on the images. Note that, the image frame position is also calculated using the projection matrix: by defining the position on the x and y axis using the camera's offset and focal length on the corresponding axis. 
	
	\subsubsection{Rendering Selectively}
\begin{figure}
		\begin{subfigure}[b]{0.4\textwidth}
			\centering
			\includegraphics[width=\linewidth]{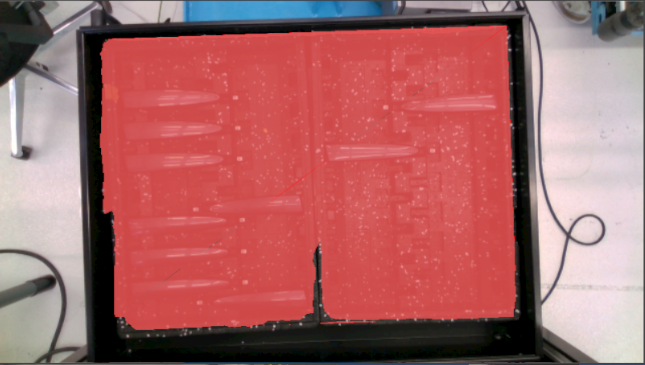}
			\caption{Non-Blocking Mesh}

		\end{subfigure}
		\begin{subfigure}[b]{0.4\textwidth}
			\centering
			\includegraphics[width=\linewidth]{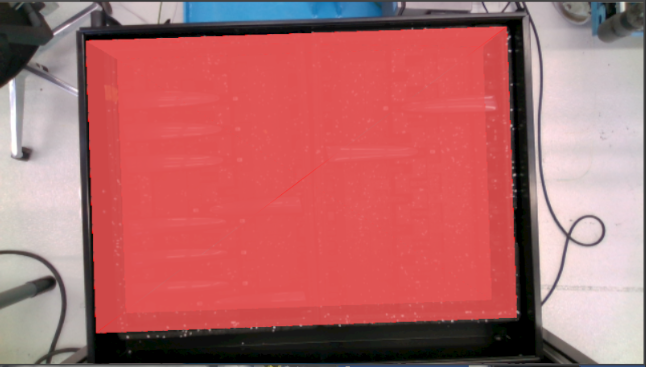}
			\caption{Blocking Mesh}
		\end{subfigure}
		\centering
		\caption{Blocking Mesh Example}
		\label{fig:blockingmesh}
	\end{figure}

	To allow the camera to render selectively the image frame object and the 3D labeling object (3D bounding box) on the mesh, we defined a culling mask\footnote{In Unity, the culling mask is a camera property used to include or omit layers of objects to be rendered by the Camera after assigning the corresponding layer to the object.} for the camera and assigned to it the culling layers of both the image frame and the labeling element. As a result, the camera shows only both of these layers without including another 3D object. In addition, the user is given the option to block the mesh, this allows the labeling elements that are behind the mesh to be seen. This is done using Unity's shader programs that are tailored to calculate post processing effects and also used to manipulate the camera's rendering information. Thus, to be able to block the mesh, we had to assign a blocking shader to the mesh. Briefly, we assign the shader a rendering queue number that is responsible for ordering how the camera should render objects. In our case we assigned a lower value to the rendering queue of the mesh comparing to the one of the labeling objects. As a result, when the user clicks on "Block" the mesh shader is switched to "Blocking" and the labeling objects are rendered before the mesh (see Figure~\ref{fig:blockingmesh}).
	
	\subsection{Minimum Bounding Rectangle}
	In the previous section, we described how the labeling object is rendered on the images. However as we discussed earlier, the input to any object detection machine learning model is an axis-aligned bounding rectangle. Therefore, we need to extract this box from the rendering we got from the back-propagation before we feed it to the machine learning model.
	\subsubsection{Getting the Image Ready}
\begin{figure}[!hbt]
		\centering
		\includegraphics[scale=1]{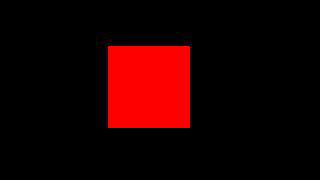}
		\caption{Sent Image of the Bounding Box}
		\label{fig:bb_image_sent}
	\end{figure}

	The images are sent through websockets by preparing a message as an object that encapsulates an array of bytes, a dictionary that holds the color of each labeling object and the shot ID. A "render texture" is created and assigned to the shot view camera as a target texture where everything that is rendered by this camera is projected onto this texture. It is not necessary to send all of the images to the server. To calculate the bounding rectangle, the server only needs to know the position of the labeling object in the image. Henceforth, we should only send the labeling object(s) by coloring the remaining image's pixels in black. For this case, we modify the camera's culling mask to only render the labeling element(s) and we changed the clear flag of the shot view camera to be able to clear with a solid background black color (see Figure~\ref{fig:bb_image_sent}).\\
	Since we are designing a real-time application and our whole goal is to make labeling faster, we also decided to down-sample the image to a smaller resolution. After thorough testing, we found that $320\times180$ is the best compromise between speed and accuracy. The texture is then encoded into an array of bytes.\\
	Moreover, the labeling object colors are sent so the server can differentiate between them to calculate the corresponding bounding box.\\
	Finally, the message is serialized into a JSON object and sent to the Python server.
	
	\subsubsection{Decoding and Reading Image} \label{imagetonumpy}
	Once the image is received on the Python side, we use the \textit{Pillow} Library to decode and read the image. We also access the additional color information to identify the labeling object(s) on the decoded image. To perform manipulations on the Image, we transform it to a $320\times180\times4$ \textit{Numpy} matrix in which each entry represents the color(RGBA) of a pixel at a given position.\\
	As we previously mentioned, the $320\times180\times4$ matrix will contain the color black and a different color for each additional labeling object. The next step is to identify the coordinates of each labeling object on the image. To do this, for each color of a labeling object we loop through the matrix and each time we encounter a pixel that has this color we save its coordinates(row and column in the matrix).\\
	Once we have the coordinates each labeling object. We can now calculate the minimum bounding rectangle.
	\subsubsection{Simple Method} \label{Simple Method}
	To define a rectangle, we only need 3 points. Given the indices of the labeling object, we simply need to find three of the following four values:(1) The pixel with the smallest $x$ value, (2) The pixel with the smallest $y$ value, (3) The pixel with the largest $x$ value and (4) the pixel with the largest $y$ value.\\
	Once this is achieved, all we need to do is solve the trivial geometry problem of calculating the fourth variable.
	\subsubsection{Small Bounding Rectangle} \label{smallbounding}
	In certain cases, when the labeling object is back-propagated on to the images. On certain images it might be very small because the object is exiting the view or it is hidden by another object. Calculating a bounding rectangle around this labeling object would lead to a rectangle with a very small area and would confuse the machine learning model. Therefore, it is better if we don't return any rectangles to Unity in these cases.
	This problem can be solved in two ways: (1) Make sure that the rectangle's area is larger than a certain threshold or (2) make sure that its length and width are each larger than a certain threshold.
	We decided to implement both options, after performing the steps described in section \ref{Simple Method} we only send the result if all of these conditions are met: 
	\begin{equation}
	\frac{rectangle area}{image area}*100 > Threshold
	\end{equation}
	\begin{equation}
	\frac{rectangle height}{image height}*100 > \sqrt{Threshold}
	\end{equation}
	\begin{equation}
	\frac{rectangle width}{image width}*100 > \sqrt{Threshold}
	\end{equation}
	\paragraph{Threshold: }Machine learning models usually struggle to train on objects with sizes smaller than $25\times25$ pixels. Thereupon we get the threshold by the following equation:
	\begin{equation}
	Threshold = \frac{25*25}{Resolution} = \frac{625}{320*180} = \frac{625}{57600} \simeq 0.0108 \simeq 1.1\%
	\end{equation}
	\subsection{Outlier Detection}
\begin{figure}[h]
		\caption{Example of an object blocking the annotated object.}
		\label{Figure:blockingbox}
		\centering
		\includegraphics[width=1\textwidth]{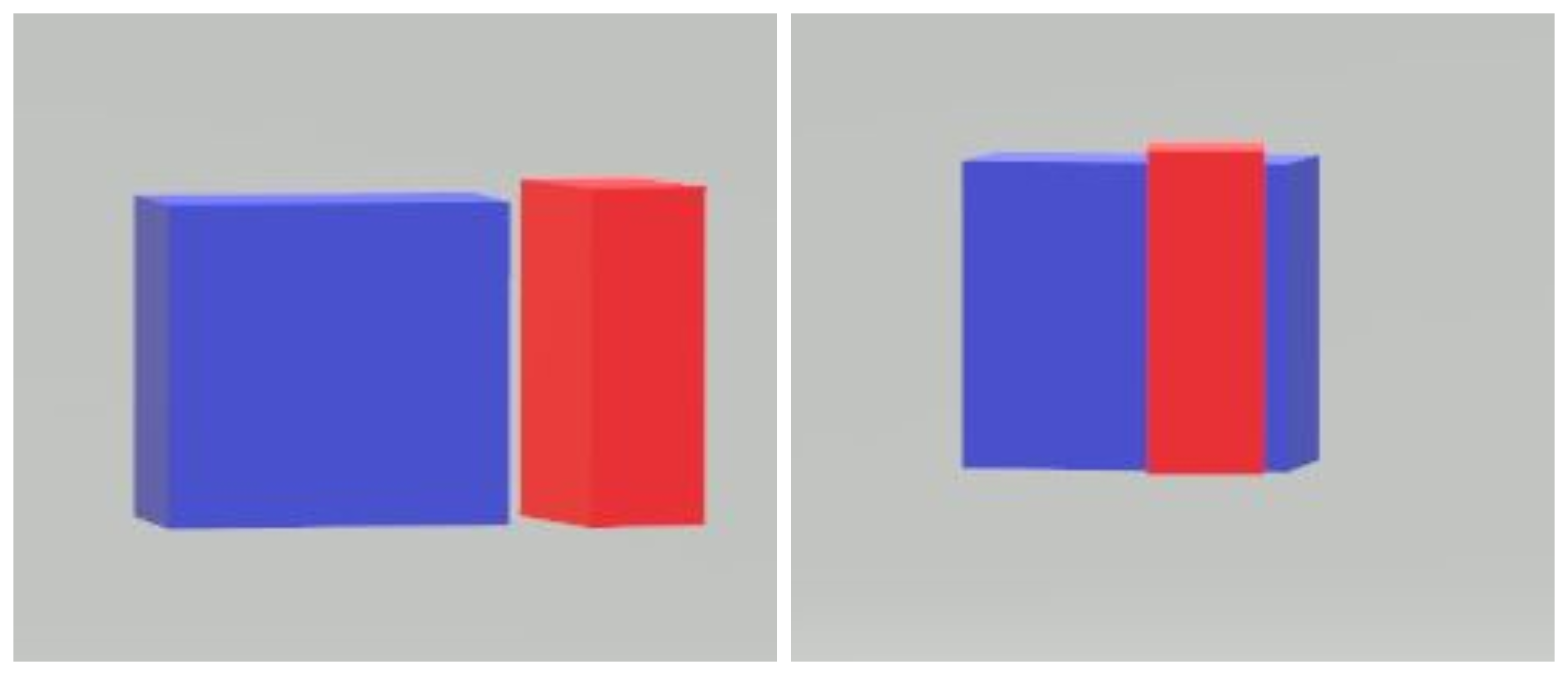}
	\end{figure}

\begin{figure}[h]
		\caption{Bounding box calculated using the simple method.}
		\label{Figure:simplebox}
		\centering
		\includegraphics[width=0.5\textwidth]{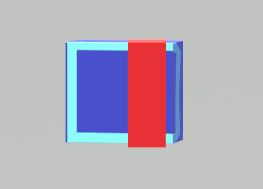}
	\end{figure}

\begin{figure}[h]
		\caption{Bounding box calculated using the simple method with outlier detection enabled.}
		\label{Figure:outlierbox}
		\centering
		\includegraphics[width=0.5\textwidth]{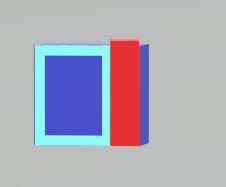}
	\end{figure}

	In some cases, a second object may be blocking the labeled object in some image but not blocking it at all in another. For example in Figure \ref{Figure:blockingbox}, in both images the	two objects are fixed in the same position, only the camera moves. The blue object is the annotated one and the red object is the blocking one. The left image shows an angle where the labeled object is not blocked while the image on the right shows another angle where it is. In these cases the simple method that we described earlier would calculate the light blue axis-aligned bounding rectangle illustrated in Figure \ref{Figure:simplebox}. This result is not the desired one, as including the red object in the bounding rectangle will confuse the machine learning model. The preferred calculated bounding rectangle is the one shown in Figure \ref{Figure:outlierbox}.
	This way the machine learning model will only see the actual object. Even though it is not the full object it is still better than including a foreign object in the bounding rectangle. Theoretically, this is equivalent to saying the right most part of the labeled object is very small compared to the left part and therefore can be considered as an outlier. Therefore, if we use outlier detection \footnote{Outlier detection is the method of identifying unusual objects in datasets or occurrences that vary from the norm.} on the labeling object and removing the outliers before calling the simple method described in \ref{Simple Method}, then the resulting bounding box would be the one illustrated in Figure \ref{Figure:outlierbox}.
	\subsubsection{Categorization of Anomaly Detection}
	Unlike the well-known classification system, where training data is used afterwards to train a classifier and test information measure efficiency, various setups are feasible when speaking about anomaly detection. Basically, the setting for anomaly detection relies on the labels in the dataset and we can differentiate between three primary kinds:
	\paragraph{Supervised Anomaly Detection:}Describes the configuration where the information includes fully labeled training and test data sets. First a normal classifier can be trained and subsequently implemented.
	With the exception that classes are typically heavily unbalanced, this situation is very comparable to traditional pattern recognition. Therefore, not all classification algorithms are perfectly suited for this assignment. For example decision trees such as in \cite {Quinlan:1993:CPM:152181}, cannot handle unbalanced data well, while Support Vector Machines (SVM) \cite{Scholkopf:2001:LKS:559923}, Artificial Neural Networks (ANN) \cite{Mehrotra:1996:EAN:241682} generally perform better. However, because of the assumption that anomalies are recognized and properly labeled, this configuration is practically not very important for numerous applications.
	\paragraph{Semi-supervised Anomaly Detection: }Once again training and test datasets are used, but training data consists only of ordinary data without any outliers thus requiring no labels. The fundamental concept is that a standard class model is learned and anomalies can be identified afterwards by deviating from that model. This concept is also known as the "one-class" classification \cite{Moya:1996:NCM:226986.226999}. Well-known algorithms are one-class SVMs \cite{Scholkopf:2001:ESH:1119748.1119749} as well as autoencoders \cite{Hawkins:2002:ODU:646111.679466}. Of course, any density estimation technique can generally be used to model the ordinary classes' likelihood density function, such as Gaussian Mixture Models \cite{10.2307/4153184} or Kernel Density Estimation \cite{Davis2011}.
	\paragraph{Unsupervised Anomaly Detection:}It is the most versatile configuration because it doesn't require any labels and it does not differentiate between a test and a training dataset. An unsupervised algorithm for anomaly detection scores the information based exclusively on the information set's inherent characteristics. Typically, distances or densities are used to offer what is normal and what is an outlier to an estimation. For these reasons, we will explore unsupervised methods more thoroughly.\\
	There are two possible outputs for an anomaly detection algorithm. First, a label that indicates whether an instance is an anomaly or not. Second, a score or confidence may be a more informative outcome that indicates the degree of abnormality. Due to available classification algorithms, a label is often used for supervised anomaly detection. On the other hand, scores are more prevalent for semi-supervised and unsupervised algorithms of anomaly detection. This is primarily owing to practical purposes, where applications frequently rank anomalies and only the top anomalies report to the user. We also use scores in this report as output and rank outcomes so that the ranking can be used to evaluate the performance. Moreover, any score can be converted to a label assuming a certain threshold.
	\subsubsection{Type of Anomalies}
\begin{figure}[h]
		\caption{Example of global anomalies(x1,x2), local anomaly x3 and micro-cluster c3 \cite{10.1371/journal.pone.0152173}}
		\label{Figure:anomalies}
		\centering
		\includegraphics[width=0.5\textwidth]{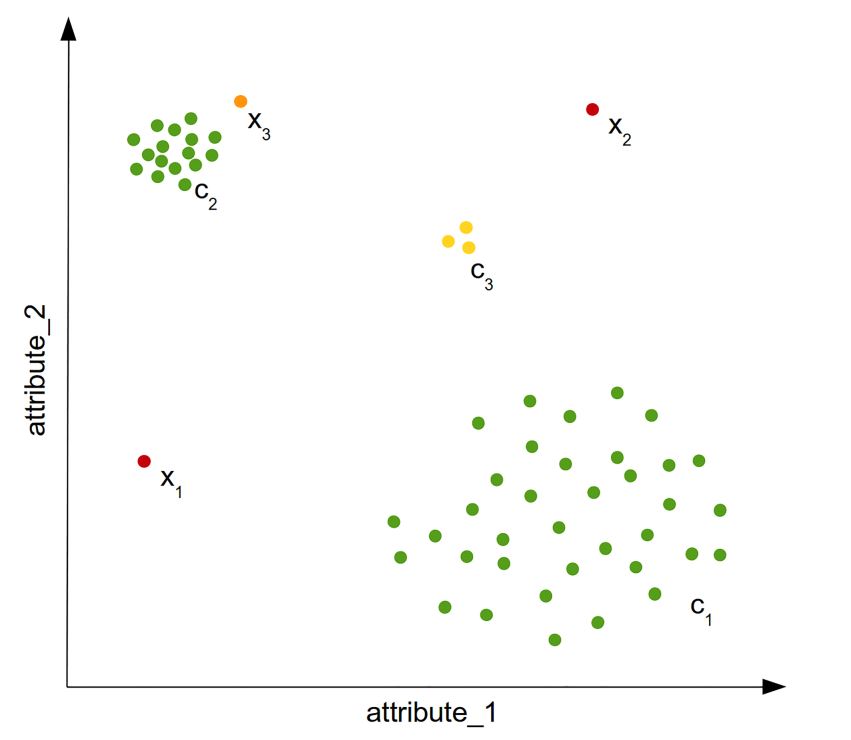}
	\end{figure}

	In practice, the idea of an anomaly is ambiguous. It heavily depends on the application it is used upon. The work in \cite{10.1371/journal.pone.0152173} proposes a very informative representation of this ambiguity in Figure \ref{Figure:anomalies}.
	By looking at this figure we can clearly discern that $x_1$ and $x_3$ are outliers. They  are called global outliers and can be extracted by looking at the entire dataset. On the other hand, $x_3$ may seem an inlier by performing the aforementioned observation. But if we only focus on $c_2$ and $x_3$, we can conclude that x3 is an outlier since it is the only one that is far away from the cluster. In this case, $x_3$ is known as a local outlier. Finally $c_3$ presents itself as another odd case, it is hard to say whether it should be identified as 3 anomalies or one miniature cluster. As we can observe, defining what an anomaly consists of is not set in stone. On the contrary, it entirely depends on the application and requires a very clear understanding of the data to be analyzed. To this extent, it is more obvious now that using scores (fuzzy) in favor of labels(binary) is the right way to go.
	\paragraph{Anomaly in our problem} The purpose of using anomaly detection in our application is to minimize as much as possible the possibility of creating training data that can confuse the object detection model. Since the user is labeling data in 3D, in certain pictures the object being labeled could be blocked by another object. In some cases this blocking may not be severe, but in others such as in the right side of Figure \ref{Figure:blockingbox}, it can cause the creation of a bounding box that includes the blocking object and accordingly confuse the model about the objects form, see Figure \ref{Figure:simplebox}. Therefore, an anomaly in our case is a point or small cluster in the data set that is far from the biggest cluster (biggest visible part of the object we are labeling).\\
	Now that we know in which cases we need to detect anomalies, we can make informative decision on the algorithm we will choose.
		\begin{table}
		\centering
		\small
		\resizebox{\textwidth}{!}{
		\begin{tabular}{ |c|c|c|c| } 
			\hline
			Formulations & Type & References  \\
			\hline
			\multirow{2}{6em}{Nearest-neighbor} & Global & K-NN\cite{Ramaswamy:2000:EAM:342009.335437} and $K^{th}$-NN\cite{Angiulli:2002:FOD:645806.670167} \\ 
			& Local & LOF\cite{Breunig:2000:LID:342009.335388}, COF\cite{Tang:2002:EEO:646420.693665}, LoOP\cite{Kriegel:2009:LLO:1645953.1646195}, LOCI\cite{1260802} and aLOCI\cite{1260802} \\ 
			\hline
			\multirow{2}{6em}{Clustering} & 
			Global & CBLOF\cite{1260802} and uCBLOF\cite{1260802}\\ 
			&	Local & LDCOF\cite{inproceedings} and CMGOS\cite{10.1007/978-3-642-41822-8_16} \\ 
			\hline
			\multirow{1.25}{6em}{Statistical} & Multi  &  HBOS\cite{hbos} \\
			\hline
			\multirow{1.5}{6em}{Subspace} & Multi  &  rPCA \cite{4137092} \\
			\hline
			\multirow{1.5}{8em}{Semi-Supervised} & Multi  &  One-Class SVM \cite{Chandola:2009:ADS:1541880.1541882} \\
			\hline
			\multirow{1.5}{10em}{Unsupervised GANs} & Multi  &  SO-GAAL\cite{DBLP:journals/corr/abs-1809-10816} and MO-GAAL\cite{DBLP:journals/corr/abs-1809-10816} \\
			\hline
		\end{tabular}
	}
		\caption{Outlier detection methods}
		\label{table:outlierdetection}
	\end{table}
	\subsubsection{Outlier detection methods}
\begin{figure}[h]
		\caption{Covariance matrix for 3-dimensional data}
		\label{Figure:covariance}
		\centering
		\includegraphics[width=0.5\textwidth]{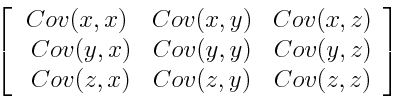}
	\end{figure}

	Over the years, many solutions have been proposed to solve the outlier detection problem. Table 5.1 shows a list of the most relevant methodes for outlier detection, but we will briefly go through some of them in this section.
	\paragraph{K-NN and Kth-NN: }The global unsupervised anomaly detection algorithm k-nearest-neighbor is a simple way to detect anomalies and not to be confused with the classification k-nearest neighbor. It focuses on global anomalies, as the name already suggests, and is unable to identify local anomalies. First, the k-nearest-neighbors must be discovered for each record in the dataset. Then, using these neighbors, one of two variants for an anomaly score is calculated: (1) the distance to the kth-nearest-neighbor \cite{Angiulli:2002:FOD:645806.670167} or (2) the median distance to all k-nearest-neighbors \cite{Ramaswamy:2000:EAM:342009.335437}. The first technique is referred to as kth-NN and the latter K-NN. In practice, K-NN is yields more favorable results \cite{6628667,66234324201}.\\
	Naturally, selecting the parameter $k$ is essential for the outcomes. If its value is selected too low, the record density estimate may not be accurate. On the other hand, if it is too big, it may be too coarse to estimate the density. As a rule of thumb, $k$ should be in the range $]10, 50[$.
	An appropriate $k$ can be determined in classification, e.g. by using cross-validation.
	Unfortunately, in unsupervised anomaly detection, there is no such method due to missing labels.
	\paragraph{LOF: }The local outlier factor \cite{Breunig:2000:LID:342009.335388} is the most well-known local anomaly detection algorithm and the pioneer of local anomalies. Today, in many nearest-neighbor algorithms, such as those outlined below, its concept is carried out. To calculate the LOF score, it is necessary to calculate three measures:\\
	1. For each record $x$, the $k$-nearest-neighbors must be found. More than $k$ neighbors are used in the event of a $k$\textsuperscript{th} neighbor's distance tie.\\
	2. The local density for a record is estimated by computing the local reachability density (LRD) using the computed $k$-nearest-neighbors $N_k$.
	\begin{equation}
	LRD_k(x)=\frac{1}{(\frac{\sum_{a\in N_k(x)}^{}d_k(x,o)}{|N_k(x)|})},
	\end{equation}
	where $d_k(.)$ is the Euclidean distance.\\
	3. Finally, the score is calculated by comparing the LRD of a certain instance's neighbors and its own LRD:
	\begin{equation}
	LOF(x)=\frac{\sum_{a\in N_k(x)}^{}\frac{LRD_k(o)}{LRD_k(x)}}{|N_k(x)|}
	\end{equation}
	Therefore, the LOF score is basically a local density ratio. This results in LOF's good property to get a score of about 1.0 in ordinary cases, where densities are as large as their neighbors' densities. Anomalies that have a small local density will lead to higher results.
	It is also evident at this stage why this algorithm is local: it depends only on its immediate neighborhood hood and the score is a proportion based primarily only on the neighbors $k$. Global anomalies can also be identified, of course, as they also have a small LRD compared to their neighbours. It is essential to note that this algorithm will generate false positives when local anomalies are not of interest. Similarly, for this algorithm, choosing the correct $k$ is essential.
	\paragraph{LoOP: } As we have already discussed, anomaly scores are a better metric compared to labels. Unfortunately, in LOF it is not yet evident after which score threshold we can obviously believe of an istance as an outlier.\\Local outlier probability (LoOP) \cite{Kriegel:2009:LLO:1645953.1646195} attempts to tackle this problem by outsetting a probability of anomaly rather than a score, which could also lead to a better comparison of anomalous data between distinct datasets. One thing to consider however is that this probability is relative to the outlier with the highest anomaly score, i.e. if record $x$ has the highest score out of all records, LoOP will assign a probability of 100\% to it. This could lead to a problem if for instance after assigning the probabilities, we add to the dataset another record that is more anomalous than the highest one set. As we can see in this easy case, the probabilities are still relative to the records and may not be too different from a regular score.
	\paragraph{LOCI: }Choosing $k$ is a key choice for detection efficiency for all the above algorithms. There is no way, as already stated, to estimate a good $k$ based on the given information. Nevertheless, using a maximization strategy, the Local Correlation Integral (LOCI) \cite{1260802} algorithm addresses this problem. The fundamental concept is that for each record, all feasible $k$ values are used and the highest score is taken in the end. LOCI describes the $r$-neighborhood by using a $r$ radius, which is extended over time, to accomplish this objective. Like LoOP, local density is also estimated using a half-Gussian distribution, but instead of distances, the quantity of records in the neighborhood is used. Local density estimation is also distinct in LOCI: it compares two neighbourhoods of distinct sizes instead of the local density ratio. A parameter $\alpha$ regulates the various neighborhoods' ratio. However, this modification comes at cost since typically, anomaly detection algorithms based on nearest-neighbor have a $O(n^2)$ computational complexity to find the closest neighbors. As the radius $r$ also needs to be expanded from one instance to the furthest in LOCI, the complexity increases to $O(n^3)$, making LOCI too slow and impractical for larger datasets.\\
	An improvement, aLOCI \cite{1260802} was also proposed by the authors to alleviate the complexity problem. They use quad trees to make the counting of the two neighborhoods faster by using a constraint on the parameter $\alpha$. The authors state that $O (NLdg + NL(dg + 2d))$ is the computational complexity of their algorithm, consisting of tree formation and outlier detection, where $d$ is the amount of dimensions. As typical of tree approaches, the amount of dimensions can be seen to have a very adverse effect on runtime. Moreover, it is shown in \cite{10.1371/journal.pone.0152173} that aLOCI is highly unstable and often produces bad results.
	\paragraph{CBLOF: }All previous algorithms for anomaly detection are based on estimation of density using nearest neighbours. Alternatively, the cluster-based local outlier factor (CBLOF) \cite{He:2003:DCL:770340.770389}, uses clustering to identify dense areas in the data and then performs a density estimate for each cluster. In theory, each clustering algorithm can be used in a first step to cluster the data. In practice however, $k$-means are commonly used to take advantage of the low computational complexity that is linear compared to the nearest-neighbor search's quadratic complexity. CBLOF uses a heuristic after clustering to classify the clusters resulting in large and small clusters. Finally, an anomaly score is calculated by multiplying the instances belonging to its cluster by the distance of each instance to its cluster center.
	\paragraph{HBOS: }The histogram-based outlier score \cite{hbos} is a simple algorithm for statistical anomaly detection assuming the characteristics(dimensions) are independent. The fundamental concept is to create a histogram for each function of the dataset. Then, for each instance in the data set, the inverse of each bin (each feature) it resides in are multiplied. The concept is very similar to the Naive Bayes algorithm in classification, where all autonomous probabilities of features are multiplied. At first, neglecting the dependencies between characteristics may seem a bit counterproductive, but comes with a large benefit in terms of computational complexity. HBOS can process a dataset within a minute, while computations based nearest-neighbor take more than 23 hours \cite{book}.
	\paragraph{OCSVM: }One-class support vector machines[24] are often used for semi-supervised detection of anomalies[15]. A one-class SVM is trained on anomaly-free data in this setting, and later it classifies the test set with either anomalies or normal data. One-class SVMs intend to separate the origin from the kernel space data instances, resulting in some type of complex hulls describing the normal data in the feature space. While one-class SVMs are strongly used as a semi-supervised technique of detection of anomalies, the use of a soft margin converts it into an unsupervised algorithm. In the unsupervised anomaly detection scenario, the one-class SVM is trained using the dataset and then each instance in the dataset is scored by a normalized distance to the determined decision boundary[40]. The parameter regularization $\gamma$ must be set to a value larger than zero so that a soft-margin properly handles the contained anomalies. In addition, one-class SVMs have been altered to include additional robust methods to deal explicitly with outliers during training. For example in [40], an improvement named $\eta$-OCSVM is introduced. In this enhancement, the parameter $\eta$ is integrated in the training process to estimate the \textit{normality} of an instance. Basically, this allows for outliers to have less of an impact on the final decision boundary.
	\paragraph{rPCA: }A frequently used method for identifying subspaces in data sets is the principal component analysis. It may also serve as an anomaly detection method, where deviations from usual subspaces is an indication of anomalous data. First, the data is standardization of the input variables(dimensions) in order for all of them to contribute equally in the analysis. This is done by applying the following equation:
	\begin{equation}
	z = \frac{t - \mu}{\tau}
	\end{equation}
	on each variable in the data set where $\tau$ is the standard deviation, $\mu$ is the mean and $t$ is the value.\\
	This is followed by the calculation of the covariance matrix between the variables. Let's consider the number of variables in the data set to be $p$. Then the covariance matrix would be a $p\times p$ matrix. In Figure \ref{Figure:covariance}, an example of a covariance matrix is shown representing a 3 dimensional data set. The diagonal is just the variances of each variable separately ($Cov(x,x) = Var(x)$) and we can clearly see that since $Cov(x,y) = Cov(y,x)$, this matrix is also symmetric. The entries in the covariance matrix provide us with one of two conclusions: (1) Correlation, in case of a positive value, means the two variables increase and decrease together or (2) Inverse Correlation, in the opposite case, means when one variable increases the other decreases and vice versa.\\
	After that, we can calculate the principal components. There exists as many principal components as there are variables. To compute them we have to compute the eigenvectors of the covariance matrix. Each eigenvector will also have its own eigenvalue. If we sort these eigenvectors (principal components) in decreasing order of eigenvalue, we will end up with the effective rank of the principal components in terms of variation captured. In other words, the first ranked principal component captures the largest possible variance in the data set.\\
	Finally, we have the option to choose which principal components to keep. For example, if we have 10 dimensions and decide to keep half of the calculated principal components then we would have successfully reduced a 10 dimensional problem into a 5 dimensional one. This of course comes at the cost of accuracy. Moreover, choosing the highest ranked principal components is not always the best choice. To recall what an outlier is, it is an instance that deviates from the norm. In other words, looking at the principal component that shows the most variation would actually help in detecting global (clustered) outliers and looking at the lower ranked ones would help detect more local outliers.
	\subsubsection{Computational Complexity}
	All nearest-neighbor algorithms require a computational time of O($n^2$) to calculate the nearest neighbors. The rest of the calculations are neglected since they don't account for a lot of the computation. Similarly in cluster based methods, the clustering algorithm is quadratic as well. HBOS, is a better candidate compared to clustering methods since it assumes that features are not related. Thus achieving near linear time complexity. Concerning One-Class SVMs, it is hard to gage the complexity since changing the gamma (with quadric complexity) and the number of support vectors has a significant impact on the runtime. Finally, rPCA has a complexity of $O(nd^2+d^3)$ and therefore depends on the number of dimensions. It is hard to make an informed decision on speed of computation given only theoretical runtimes, it would be more beneficial to compare these algorithms in a practical use case. This will be done in the following section. 
	\subsubsection{Benchmark \& Comparison}
	To evaluate the most prominent models, we used two measures: (1) AUC-ROC Performance and (3) Time Complexity. They are executed on 17 benchmark datasets with each one being split into 60\% training and 40\% testing. This evaluation is made possible by the Python library pyod \cite{zhao2019pyod} which provides implementations to many state of the art models.
	\paragraph{AUC-ROC:}It is not as straightforward as in the classical supervised classification situation to compare the anomaly detection performance of unsupervised anomaly detection algorithms. As opposed to simply comparing an accuracy value or $precision / recall$, consideration should be given to the order of the anomalies. An incorrectly classified example is definitely an error in classification. This is distinct in unsupervised anomaly detection. For example, if there are ten anomalies in a large dataset and they are ranked among the top-15 outliers, this is still a good result even though it is not perfect. To this end, a popular evaluation strategy for unsupervised anomaly detection algorithms is to rank outcomes by anomaly score and then iteratively apply a limit from the first to the last rank. This comes in $N$-tuple values (true positive rate and false positive rate)  which form a single receiver operator characteristic (ROC). Then, as a detection performance measure, the area under the curve (AUC) calculated by the integral of the ROC, can be used. A good interpretation of the AUC-ROC is also provided in [70] where it is proven to be a good measure in the anomaly detection domain. To clarify, the AUC-ROC is the probability that, given a random normal instance the algorithm would assign this instance a lower score than a random anomaly instance. Therefore, even though AUC-ROC favors ranking between instances over comparing relative differences in scores, we think it is the most adequate evaluation measure for these unsupervised models.
	\begin{table}
		\caption{Comparison of AUC between selected models on different data sets}
		\label{Figure:roc}
		\centering
		\includegraphics[width=1\textwidth]{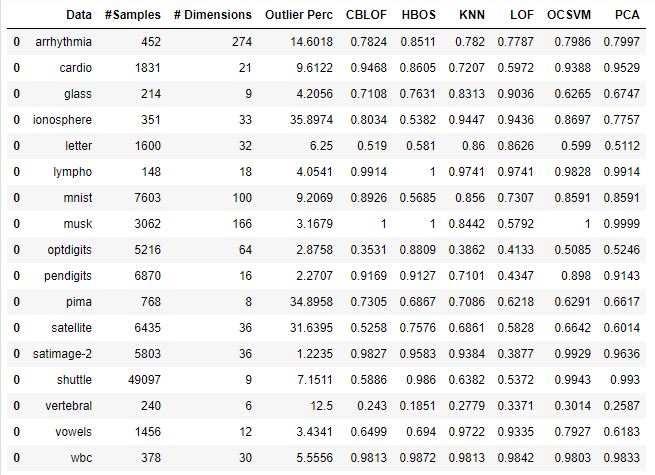}
	\end{table}
	By looking at Table \ref{Figure:roc}, we can see that the accuracies are quite similar. LOF has the worst results which is most probably due to its focus on locality and fails in instances where outliers are all global. CBLOF is also close behind and moreover in our practical tests CBLOF failed on instances where there were no anomalies in the data.\\
	\paragraph{Time Complexity:}Accuracy is not the only measure we decide upon. Since we are dealing with a real time application, computational complexity is extremely important.
	\begin{table}
		\caption{Comparison of time in seconds between selected models on different data sets}
		\label{Figure:time}
		\centering
		\includegraphics[width=1\textwidth]{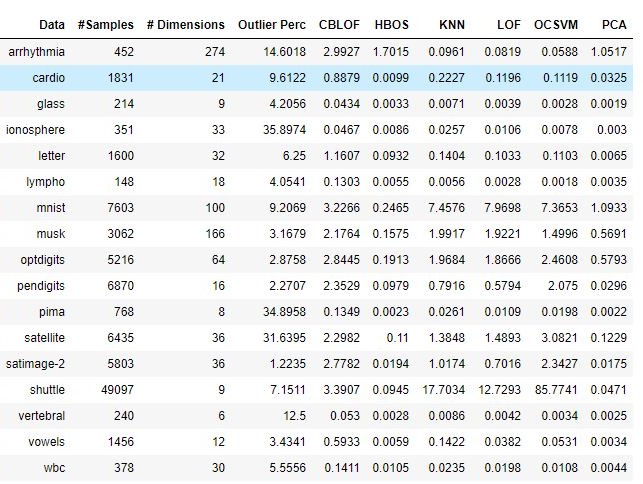}
	\end{table}
	In Table \ref{Figure:time} we show the time it took to run each model on the same datasets. The results are not surprising with rPCA and HBOS clearly the best in terms of speed. As we can see, the algorithm we should select is between the latter two. Making the final decision comes down to testing both in the tool on more practical cases. We realized that not only did rPCA outperform HBOS in runtime by a large margin -- probably due to the low number of dimensions rendering the complexity to $O(4n+8)$ which is practically linear -- but also in accuracy because of the better performance of rPCA against global anomalies.
	\subsection{Implementation}
	To implement rPCA we use the aformentionned python library pyod \cite{zhao2019pyod}. As for the number of principal points we evaluate on we can choose between 3 options: (1) use the most prominent one, (2) use the less prominent one or (3) use both. We tried all three options with different practical examples and found that the first option focused more on global anomalies, hence yielding the results we wanted with even better runtime\footnote{reduction to 1 dimension}.\\
	The outlier detection step is performed directly after reading the image into a NumPy\footnote{NumPy is the fundamental package for scientific computing with Python.} array \cite{oliphant2006guide} as described in Section \ref{imagetonumpy}. After we get the prediction from rPCA we remove all the anomaly points from the numpy array and then we perform the steps in sections \ref{Simple Method} \& \ref{smallbounding}.\\
	After all of the steps are done, we have the coordinates of four points in 2D. These are the 4 points that constitute the rectangle on the image, as previously described in Section \ref{Simple Method}. Of course, each image has its own calculated rectangle since the object moves in each camera shot. Lastly, this process can be easily scaled in the presence of multiple labeling objects, as for each image we repeat the same process for each labeling object. The color array which is sent to us along with the image contains the colors of each labeling object in that image.
	\subsection{Drawing the bounding rectangle on image}
\begin{figure}[!hbt]
		\centering
		\includegraphics[scale=0.5]{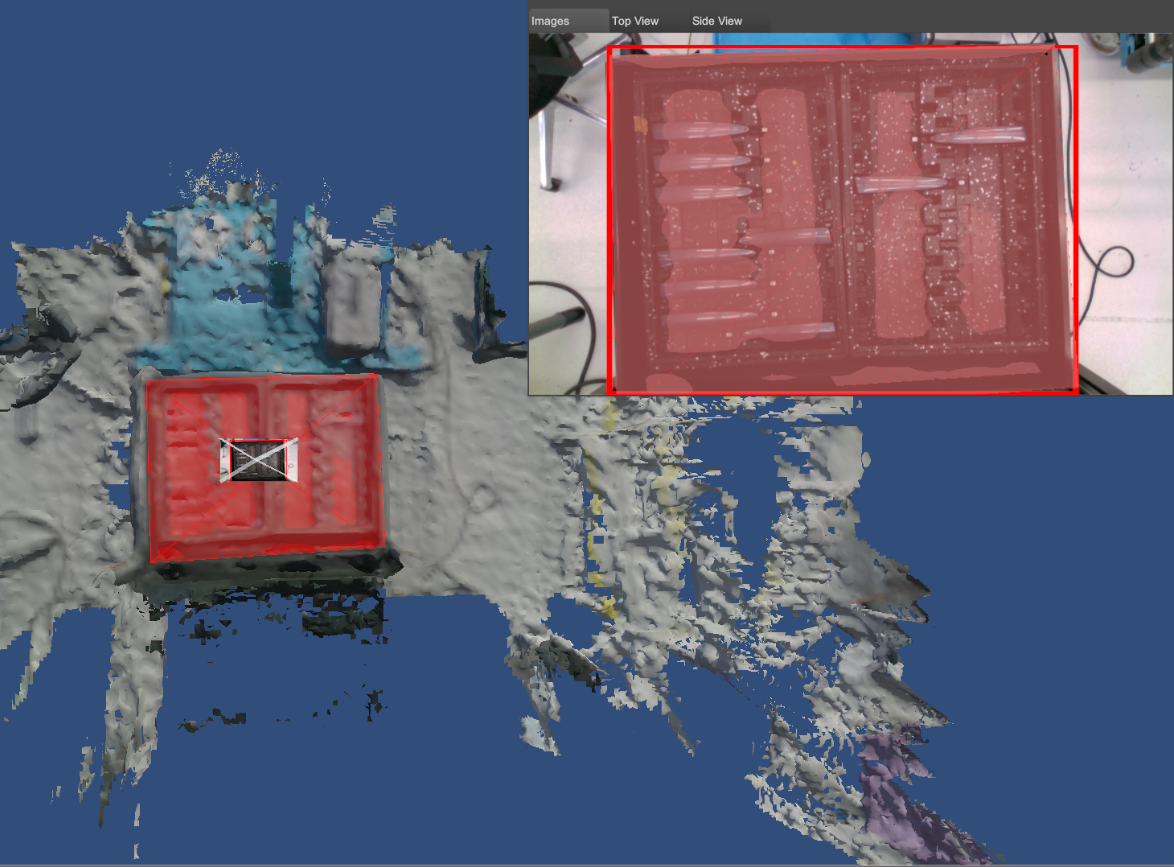}
		\caption{Bounding Box Example}
		\label{fig:boundingbox}
	\end{figure}

	 On the Unity side, the client receives the message from the server that holds a dictionary containing the minimum and maximum boundaries of each labeling object along with the shot ID. However, a class for the 2D objects is created that encapsulates the object's ID, color, 2D bounding box and class name. On each annotation an object of this class is created and stored in a dictionary.
	 \\ To draw the received bounding box on the texture of the image frame (see Figure~\ref{fig:boundingbox}), we used the "Shader" calculations in Unity, knowing that shaders calculate how the pixels are colored based on a given input and the material of the object. The calculations were based on the minimum and maximum boundaries that are given to the shader, this latter is computing the pixel to be colored by the 3D labeling object's color based on the line width of the 2D bounding box that was set to a static value; if the distance from the texture's pixel to the min or max boundaries is less than or equal to the line width then the pixel is colored with the labeling element corresponding color. 
	 \\Nevertheless, if two objects have the same color, the 2D bounding box will encapsulate both objects in the images. To avoid this error, the user is given the opportunity to change the color of the labeling object.
	\begin{figure}[!hbt]
		\centering
		\includegraphics[scale=0.4]{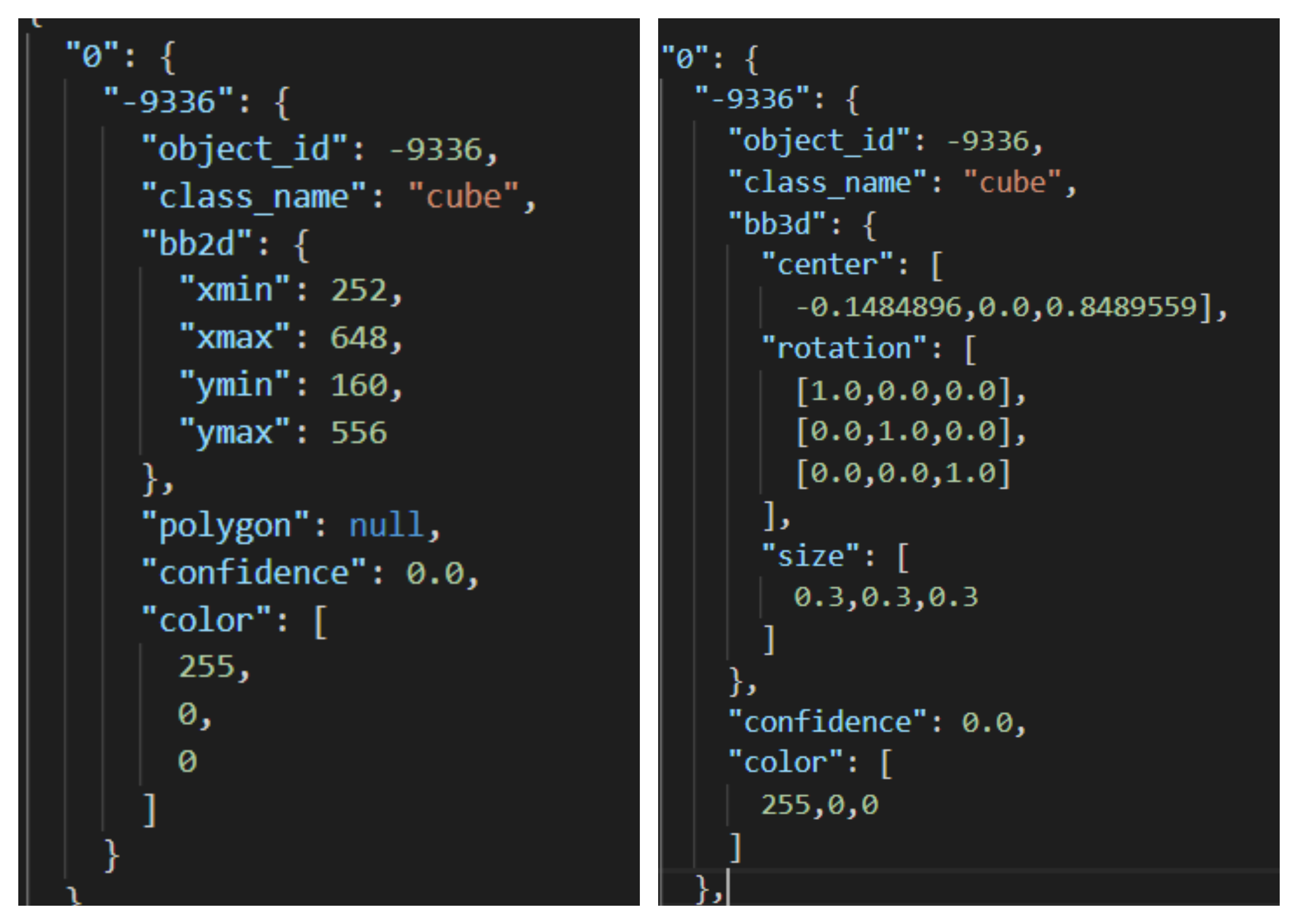}
		\caption{2D \& 3D Annotations Example}
		\label{fig:annotations}
	\end{figure}
	\subsection{Exporting Annotations}
	After the user is done with the labeling process, they can export the annotations in the 2D and 3D formats. The 2D annotations hold information about the bounding boxes that are labeled on the images. This information is established as follows: the shot ID, the object ID, class name, bounding box min and max boundaries and the box's color. All of this information is stored in a dictionary that is finally serialized into a JSON Format (see Figure~\ref{fig:annotations} on the left).\\
	Alternatively, the 3D annotations describe the labeling object's center position, rotation and color (see Figure~\ref{fig:annotations} on the right). Finally, both of these files can be fed to a machine learning model for training purposes along with the images and the depth as soon as this information is re-parsed into the accepted format by the machine learning algorithm.

	\chapter{Phantom Mode}
	\section{Introduction}
	Unlike defining a bounding rectangle on a 2D image\footnote{We only need to define the position and the scale of the rectangle on axes $X$ and $Y$ (4 DoF)}, placing a cube in its correct position in space is non trivial\footnote{We need to define the position, rotation and scale of the cube on the axes $X$, $Y$ and $Z$ (9 DoF)}. We can greatly simplify the process by defining corresponding points on the labeling object and the reconstructed mesh. Finding the optimal transformation parameters that transform the chosen points on the labeling object to the chosen points on the mesh is a known problem in Computer Vision, the registration problem.
	\paragraph{Problem Statement}\label{probstate}
	Given a reconstructed mesh $M$ and labeling object $L$; we want to give the user the possibility to select three or more points $V\lbrace v_a \vert a=1, 2, \ldots, N \rbrace$  and $X\lbrace x_i \vert i=1, 2, \ldots, K \rbrace$ on $L$ and $M$ respectively. We then have to find the transformation parameters (Rotation Matrix $R$, Translation vector $T$ and Scale vector $S$) that transform the selected point set $V$ to the selected point set $X$. The transformation parameters will satisfy the following equation:
	\begin{equation}
	\label{function:probstate}
	X= (S\odot R)V+T
	\end{equation}
	Where $R$ is a $3\times3$ matrix, $T$ is a $1\times3$ vector, $S$ is a $1\times3$ vector, $V$ is a $3\times N$ matrix and $\odot$ is a special vector-matrix multiplication defined as such:
	\begin{equation}
		\begin{pmatrix}
		s_1&s_2&s_3
		\end{pmatrix}
		\odot
		\begin{pmatrix}
		a&b&c\\
		d&e&f\\
		g&h&i\\
		\end{pmatrix}
		=
		\begin{pmatrix}
		s_1*a&s_2*b&s_3*c\\
		s_1*d&s_2*e&s_3*f\\
		s_1*g&s_2*h&s_3*i\\
		\end{pmatrix}
	\end{equation}
	The point sets do not suffer from outliers, noise or missing data. Our three biggest challenges are: (1) Registration accuracy, (2) Correspondence between the points and (3) Limiting the registration algorithm to only 9 degrees of freedom (Skew is not possible in unity)\\
	Registration of two given point sets is a common problem in computer vision. Given two point sets $X_i$ and $Y_i$, the problems we require to solve are two-fold. First, we need to find correspondence between the points that lie in the aforementioned point sets. Second, we need to find the value of the transformation parameters that map one point set to the other.\\
	Good registration algorithms satisfy the following requirements: 1) the ability to handle high dimensionalities of the point sets, 2) the ability to solve the problem with tractable computational complexity and 3) robustness to practical errors such as noise, missing points and outliers, all of which can occur due to errors in scanning. However, as we have already mentioned, we do not suffer from the latter.\\
	The transformations can be placed in two different categories: rigid or nonrigid. Truly Rigid transformations only allow for translation and rotation (6 DoF). However, conventional methods that solve for similarity transformations which also include isotropic (uniform) scaling (7 DoF) are also considered rigid registration methods. Both of these transformations preserve the shape of the point set but the former also preserves the Euclidian distance, hence its common name: Euclidian transformation. The simplest nonrigid transformation is the affine transformation. It allows for anisotropic scaling and skews (12 DoF). In its infancy, the problem was usually simplified to piece-wise affine and polynomial models, both of which are not truly nonrigid and are neither adequate for robust correspondence discovery nor correct alignment. Because of the large number of degrees of freedom, it is easy to deduce that registration methods that solve for nonrigid transformations tend to be very sensitive to outliers and noise, they also usually have high computational complexity. We seek a simpler version of the affine transformation, one that allows only anisotropic scaling (9 DoF). However, since registration algorithms don't target this specific case, we will find the best nonrigid registration algorithm and limit its degrees of freedom from 12 to 9.\\
	Many algorithms exist for rigid and nonrigid registration. We will provide a brief overview of the former and then give a comprehensive review of the latter and later advocate for the one we chose to solve our problem, as well as the modifications we performed on it.
	\section{Rigid Registration Methods}
	The Iterative Closest Point (ICP) algorithm \cite{121791,Zhang1994} is a staple of rigid body registration algorithms due to its simplicity and respectable computational complexity. As its name suggests, ICP is an iterative algorithm: it assigns correspondences based on the closest distances between the points then finds the least squares transformation that maps one to the other. It does this until it reaches a local minima. Many variants of ICP have been proposed over the years, each variant attempting to improve a specific part of the ICP algorithm \cite{article,924423}. ICP requires the point sets to have good initial positions so that the local minima achieved will be in fact the global minima.\\
	ICP uses binary correspondences. A logical improvement would be to use fuzzy correspondences, thus the emergence of probabilistic methods that use soft assignment for correspondences. The most successful of these methods is the Robust Point Matching (RPM) algorithm \cite{Gold97newalgorithms} and some of its variants \cite{RANGARAJAN1997379,854733}. In \cite{CHUI2003114}, it was proven that RPM behaves similarly to the Expectation Maximization (EM) algorithm for the Gaussian Mixture Model, where one point set is treated as data points and the other is treated as GMM centroids with equal isotropic covariances. In fact, the point set registration problem can be formulated as a Maximum Likelihood (ML) estimation problem to fit the centroids to the point sets. This formulation has been explored in \cite{954602,852377} where the centroids are parametrized by translation and rotation. The EM algorithm used to optimize the likelihood function consists of two steps: E-step to calculate the correspondence probabilities and M-step to calculate the transformation given these probabilities. Some methods add parameters to control the EM algorithm for the purpose of making it more robust to noise and outliers. These methods generally perform better than ICP. The work in \cite{DBLP:journals/corr/abs-1811-08139} uses Generative Adversarial Neural Networks (GANs) to effectively find correspondence even if the initial positions are not good, it achieves impressive results; however, neural networks are still hindered by their large running time which prevents this method from being used in real-time applications.
	\section{Nonrigid Registration}
	In this section, we will categorize nonrigid registration algorithms in terms of optimization methods. Then, we will provide a comprehensible walk-through of our chosen algorithm. After that, a modification which limits the algorithm's output results to only 9 degrees of freedom (rotation, translation and scaling) is proposed. Finally, a pseudo-code of the implemented algorithm is provided, analyzed and its implementation described. 
	\subsection{Optimization methods}
	There are four main types of optimization methods: 1) Local Deterministic methods that locally minimize an objective function but good initialization is critical or they will get stuck in local minima, 2) Global Deterministic methods avoid local minima and try to always converge to the global optimum, they either perform a full search with bounds and finds the exact global solution or relaxes the problem and finds a near global solution, 3) Stochastic methods that model correspondence and registration with probabilistic and statistic approaches to handle missing data, noise and outliers and (4) Machine learning methods that rely deep neural networks to learn to extract good correspondences and registration from training data.
\subsubsection{Global Deterministic}
\paragraph{Branch and Bound \& Tree Search }
Correspondence between the two point sets can be found using a decision tree, where each node represents a correspondence between two points $\{x_i,y_j\}$ and the root is the correspondence between two empty sets. Each path from the root to a leaf is the set of correspondences between two point sets. The search technique utilized by these methods is called branch-and-bound, it is based on a lower bound on the cost function. Naturally, the tighter the bound is, the more efficient the search will be.
A sophisticated implementation of this method is proposed in \cite{Zhu:2017:DSC:3072959.3073613}. They use the previously defined self-deformation distortion measure on the represented node correspondences of the tree to prune entire branches and drastically minimize the search space, hence increasing efficiency.
\paragraph{Graduated Assignment}
The deterministic annealing technique approximates a non-linear objective function by adding a regularization term. Another parameter is also added (chaos parameter) that controls convergence similar to how temperature controls simulated annealing in energy functions.
In \cite{854733}, thin plate splines are used to define the transformation model and graduated assignment for registration and optimization. They also add an additional parameter to the ones previously mentioned, one that controls the “rigidity” of the registration allowing the algorithm to favor and explore rigid transformations before gradually increasing the degrees of freedom.
In \cite{ZHANG2018183}, a finite mixture model able to deal with two features is introduced. The authors smoothly combine the original coordinates with the local structure descriptor through an annealing scheme to obtain the mixture structure descriptor (MSD). Then they obtain a fuzzy matrix by substituting the MSD into the constructed model. Additionally, they use the kernel Hilbert space to model the transformation space by an energy function which contains three main terms; the first is the $L_2$ estimation ($L_2E$) and the other two complementarily improve accuracy and robustness for transformation estimation at both a local and global scale.
\subsubsection{Local Deterministic methods}
\paragraph{Expectation Maximization}Some techniques alternate between solving the correspondence and the registration problems. For instance, first they fix transformation parameters and solve for correspondence, then they fix correspondences and solve for the transformation parameters. 
In \cite{5432191} the centroids of one point set is represented using a Gaussian Mixture Model (GMM), then they are aligned with the other point set. Next, the EM algorithm substitutes between the two following steps: (1) it uses the Bayes theorem to compute the posterior probability distribution of the aforementioned GMM centroids, and it improves the parameters (Covariances of GMM and transformations), (2) it calculates the rigid transformation parameters that maximize the likelihood. Adding to that, they extend this method to include nonrigid registration parameters by regularizing the displacement fields using coherence.
\subsubsection{Machine learning models}
\paragraph{Supervised}
Recently, PR-Net \cite{DBLP:journals/corr/abs-1904-01428} introduced a novel machine learning model to solve the registration problem. In contrast to non-learning based methods, this type of optimization is rarely studied. Similar to \cite{CHUI2003114} PR-Net uses TPS to model the geometric transformation, it also uses correlation tensor and shape descriptor tensor to solve the feature learning problem. 
\paragraph{Semi-Supervised}
In RPM-MR \cite{8510909}, registration is superiorly solved by casting it into a semisupervised learning problem, where a set of indicator variables are selected to differentiate outliers in a mixture model. They constrain the transformation with manifold regularization exploiting the intrinsic structure of the point sets which also plays a role of prior knowledge. Unlike PR-Net the transformation is modeled in the kernel Hilbert Space, and a sparsity-induced approximation is utilized to boost efficiency. Although this algorithm has proven to be superior to state of the art methods on public data sets, it still suffers from the downsides of local optimization techniques.
\subsubsection{Stochastic Methods}
\paragraph{VBPSM}
VBPSM \cite{7439870} is a Probabilistic Model for Robust Affine and Non-Rigid Point Set Matching. They propose a combination strategy based on regression and clustering to solve point-set matching issues within a Bayesian framework where the regression estimates the transformation from model to scene and the clustering establishes the correspondence between two point-sets. A hierarchically directed graph illustrates the point-set matching model, and a coarse-to-fine variational inference algorithm approximates the matching uncertainties. In addition, two Gaussian mixtures are proposed to estimate heteroscedastic\footnote{In statistics, a collection of random variables is heteroscedastic if there are sub-populations that have different variance from others.} noise and spurious outliers.
\paragraph{IPDA}
Point Clouds Registration with Probabilistic Data Association \cite{7759602} is a novel algorithm that, instead of solving two given dense point clouds, tackles the problem of aligning one dense point cloud to a sparse one and vice versa. It can be very useful when the two point sets are taken using different sensors, such as a vision-based sensor and laser scanner or two different laser-based sensors. Each point in the source point cloud is associated with a set of points in the target point cloud in this method; each association is then weighted to form a probability distribution. The result is an ICP-like algorithm, but more robust against noise and outliers.
\paragraph{CPPSR}
CPPSR \cite{7780570} is a Probabilistic Framework for Color-Based Point Set Registration. In almost all registration algorithms, only the position of the points is used from the point clouds, although state of the art scanners also acquire color information that can be useful information in registration. This algorithm exploits available color information by creating a mixture model of the point-color space. It uses ellipses to represent spatial mixture components where each ellipse being associated with a mixture model from the color space.
\paragraph{SVR}
SVR \cite{7410845} uses support vector machines and Gaussian mixture models to calculate registrations robust to occlusions\footnote{Missing Data}, outliers and noise. The main idea is that the robustness of a registration algorithm depends mostly on how the data is represented. The authors use a support vector-parametrized Gaussian Mixture (SVGM) to represent the data. The way it works is, each point set provided to the algorithm is mapped to the continuous domain by training a Support Vector Machine and mapping it to a Gaussian Mixture Model. Since SVMs are parameterized by a sparse intelligently-selected subset of data points, SVGM is compact and robust to fragmentation, occlusions and noise \cite{6291722}. The motivation for a continuous representation is that a typical scene consists of a single, rarely disjointed continuous surface, which cannot be fully modelled by a discrete sampled point-set from the scene. SVR calculates the optimal transformation parameters that minimise an objective function based on the $L_2$ distance between SVGMs.\\
An SVM classifies data by building a hyperplane separating data into two different classes, maximizing the margin between classes while allowing some mislabeling. Since point-set data contains only positive examples, one-class SVM \cite{Scholkopf:2001:ESH:1119748.1119749} can be used to find the hyperplane that separates data points in feature space from the origin or point of view. Training data is mapped to a higher-dimensional space with a non-linear kernel function (Gaussian Radial Basis Function (RBF) kernel) where it can be linearly separated from the origin. The optimization formulation in \cite{Scholkopf:2001:ESH:1119748.1119749} has a parameter $\upnu$ which controls the trade-off between the training error and the complexity of the model. It is a lower bound on the support vector fraction and an upper bound on the misclassification rate \cite{Scholkopf:2001:ESH:1119748.1119749}. The other parameter is the kernel width $\gamma$, which the authors estimate by noting that it is inversely proportional to the square of the scale $\sigma$.\\
To make use of the trained SVM for point-set registration, it must first be approximated as a GMM . Without altering the decision boundary, the authors use the transformation identified by Deselaers et al. \cite{Deselaers:2010:OCF:1752253.1752401} to represent the SVM within a GMM framework. Unlike standard generative GMMs, a GMM converted from an SVM will necessarily optimize classification performance instead of data representation, as SVMs are discriminative models. This enables redundant data to be discarded and reduces its susceptibility to varying point densities that are prevalent in real datasets.\\
Once the point-sets are in the form of a mixture model, the problem of registration can be posed as minimizing the distance between the two mixtures. The authors use the $L_2$ distance which can be expressed in a closed form. Conversely, the $L_2E$ estimator minimizes the distance between densities between $L_2$ and is known to be inherently robust to outliers.
\subsubsection{Special Methods}
In this section we will describe methods that we could not clearly classify with the rest of the registration algorithms.
\paragraph{Robust Point Matching via Vector Field Consensus}
In RPM-VFC \cite{6746218}, the authors start by creating a set of putative correspondences which will contain a lot of wrong correspondences and a limited amount of correct correspondences. Following that, they estimate a consensus of inlier points whose matching follows a nonparametric geometrical constraint. As a result they can then interpolate a vector field between the two point sets and solve for correspondence. They formulate a maximum a posteriori (MAP) estimate of a Bayesian model with hidden / latent variables indicating whether matches are outliers or inliers in the putative set. In addition, they use Tikhonov regularizers in a reproducing kernel Hilbert space to impose non-parametric geometric constraints on correspondence as a prior distribution. MAP estimation is carried out by the EM algorithm which is able to obtain good estimates very quickly (e.g. avoiding many of the local minima inherent in this formulation) by also estimating the variance of the previous model (initialized to a large value). This method proves extremely robust to outliers' frequencies even as high as 90\%.
\paragraph{Dependent landmark drift}
DLD \cite{DBLP:journals/corr/abs-1711-06588} uses prior geometric feature information as a way of improving registration accuracy. It encodes the provided shape information as a statistical shape model and defines a transformation model based on the combination of: (1) Motion coherence, (2) Statistical shape model and (3) Similarity transformation. Therefore, unlike previous methods, this method works extremely well if point sets have missing regions (due to knowledge of shape). Moreover, its computational cost is linear which makes it scalable to huge data sets. 
\paragraph{GLTP}
Global-Local Topology Preservation \cite{6909990} formulates the registration as a Maximum Likelihood (ML) estimation problem with two topologically complementary constraints. First, the established Coherent Point Drift [CPD] that encodes a global topology constraint by moving one point set to coherently align with the other. Second, Local Linear Embedding (LLE) is introduced to handle highly articulated deformations between the two point sets while sustaining local structure.
\section{Choice of Algorithm}
Choosing the best algorithm is equivalent to choosing the algorithm that best solves the four challenges described in section \ref{probstate}.
\subsection{Registration Accuracy}
A very good comparison of registration accuracy between state of the art methods with respect to the degree of deformation is presented in Figure 3 of \cite{Ma2015}. We can clearly see that even for the highest degrees of deformation, the average error between all of the compared algorithms varies roughly between 0.01 to 0.03. At this point, we have to select the algorithm that best fits our needs. First, since the user is choosing the corresponding points there will never be noise, outliers or missing data. Their always has to be 4 points on the labeling object and 4 corresponding points on the triangular mesh. Henceforth, we can rule out stochastic methods. Moreover, we have no exact idea what the object on the mesh will look like, nor the initial positions of either the labeling object or the object on the mesh. Adding to that, the labeling object is extremely far from the mesh when it is instantiated. These reasons rule out the use of local deterministic methods. We are left with global deterministic models and global machine learning models. We will not go for the latter since it require training and usually have high runtime complexities.
\subsection{Correspondence}
Now that we know we need a global deterministic registration algorithm, we need to narrow down the choices even more by choosing one that calculates correspondence as well as registration. The only global deterministic methods that contains both correspondence and registration calculation are the graduated assignment based optimization methods.
\subsection{Limiting to 9 DoF}
TPS-RPM contains a parameter $\lambda$ that controls the rigidity of the transformation. It allows the algorithm to first explore rigid transformations(6 DoF) and then if it doesn't minimize the objective function under a certain threshold, it starts incrementally increasing nonrigidity and allowing scaling and eventually sheering. Thanks to this parameter and given the correct correspondences, the algorithm will always choose our desired constrained (9 DoF) transformation before even exploring the fully nonrigid one. Therefore, even though the methods in \cite{ZHANG2018183,6619123} very slightly surpass TPS-RPM in accuracy for full non-rigid registrations(12 DoF), TPS-RPM guarantees constrained(9 DoF) transformations if they exist, whereas other methods would sometimes give affine ones.

\section{TPS-RPM}
\subsection{Thin Plate Splines}
\label{section:tps}
\subsubsection{Overview}
\begin{figure}[h]
	\caption{A Thin Plate Splines that passes through a set of control points}
	\label{Figure:tps}
	\centering
	\includegraphics[width=0.5\textwidth]{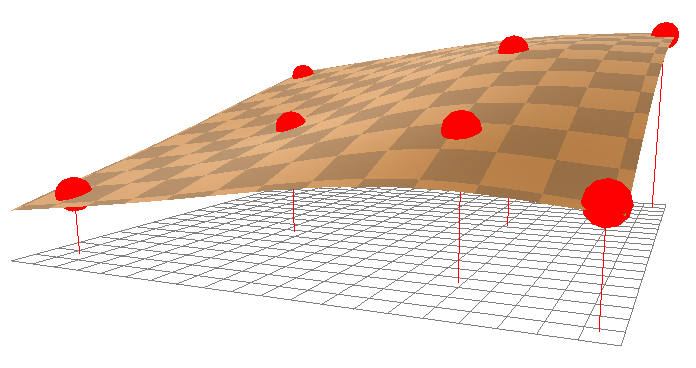}
\end{figure}

There exists quite a few techniques to ensure a smooth interpolation between a set of control points. One such technique is Thin Plate Splines, we will describe a quick overview based on the pioneering work done in \cite{24792}. A surface that passes through each control point is interpolated. Thus, a set of 3 points creates a flat plane. Control points are easy to think of as constraints of position on a bending surface. The ideal surface bends the least. Figure \ref{Figure:tps} shows a 7-point example of such a surface. All these 7 control points are forced to belong to the surface.
This least bent surface is given by the following equation:
\begin{equation}
f(x,y) = a_1 + a_2x + a_3y + \sum_{i=1}^{n}w_iU(|P_i - (x,y)|)
\label{equation:tps}
\end{equation}
The first three terms correlate to the linear part characterizing a flat plane that best fits all control points (this can be interpreted as a minimum square fit). The last term corresponds to a weighted sum over the $n$ control points bending forces. For each control point, there is a $w_i$ coefficient. $|P_i - (x,y)|$ is the distance between a given point $(x,y)$ and each control point $P_i$. The U function defines this distance as $U(r) = r^2 logr$ .
So far, for each control point, the coefficients $a_1, a_2, a_3$ and $w_i$ are unknown. All $w_i$ form the $W$ vector. The definition of these unknowns is:
\begin{equation}
L^{-1} Y = (W|a_1a_2a_3)^T
\end{equation}
We do know however the set of points $x_i$ and $y_i$ (chosen by the user) and their heights. We can therefore simply write: 
\begin{equation}
P = \begin{pmatrix} 
1 & x1 & y1\\
1 & x2 & y2\\
..    \\
1 & xn & yn
\end{pmatrix}, \text{the positions of the control points}
\end{equation}
\begin{equation}
Y = \begin{pmatrix} 
v_1 \\
v_2 \\
..    \\
v_n \\
0\\
0 \\
0 \\
\end{pmatrix}, \text{0 padded control point heights}
\end{equation}
We define a matrix K that evaluates $U(r_{ij})$ such that $r_{ij}$ is the distance between two given control points, $r_{ij}$ = $|Pi-Pj|$:
\begin{equation}
K = \begin{pmatrix} 
U(r_{11})& U(r_{12}) & ..\\
U(r_{21}) & U(r_{22}) & ..\\
.. & .. & U(r_{nn})
\end{pmatrix}
\end{equation}
The above-mentioned matrix $L$ is composed of matrix $K$ at its top-left corner, matrix $P$ at its right side, matrix $P^T$ at its bottom, and zeros at its bottom-right corner:
\begin{equation}
L = \begin{pmatrix} 
K& P\\
P^T & 0\\
\end{pmatrix}
\end{equation}
To find the matrix $(W | a_1a_2a_3)$, we either find the inverse $L^{-1}$, or solve the linear system $L (W | a_1 a_2 a_3) = Y$. The latter can be exactly solved using LU decomposition by building on the knowledge that $L$ is symmetric. Once we calculate $(W a_1 a_2 a_3)$, we can go back to equation \ref{equation:tps} and find the height $v = f(x, y)$ for any point $(x, y)$.\\
By using this reasoning, we can find the thin plate spline for points in 2D. The math can be easily adapted to 3D space and even to $N-D$ space by using a larger $P$ vector and adding more $a_i$ terms.
We can observe that the number of unknown terms is proportional to the number of control points, i.e. computational complexity increases as the number of control points increase. That being said, since our user will more often than not choose either three or four points computation of the thin plate spline is instant and no delay will be perceived by the user.
\subsubsection{Deformation}
\begin{figure}[h]
	\caption{Thin plates can be used to deform image (a) to (b)}
	\label{Figure:smile}
	\centering
	\includegraphics[width=0.5\textwidth]{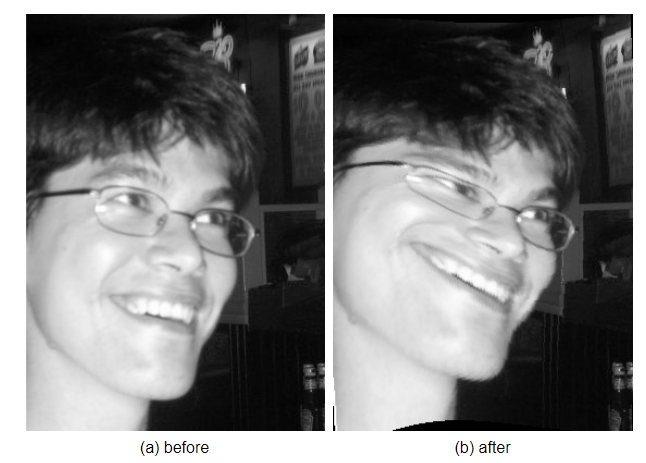}
\end{figure}

\begin{figure}[h]
	\caption{Interpolation of displacement in the (a) x direction and in the (b) y direction}
	\label{Figure:surface}
	\centering
	\includegraphics[width=0.5\textwidth]{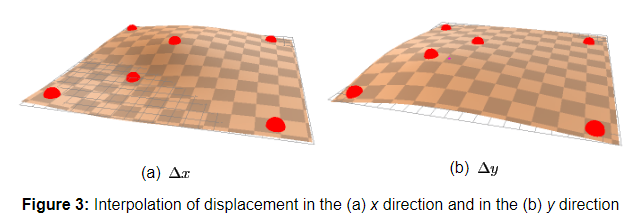}
\end{figure}

Normally, to deform an image, you require the position $(x,y)$ of each pixel on the initial image to then calculate the position $(x+d_x, y+d_y)$ on the deformed image. Conversely, by using thin plate splines we only require the height information. If we consider $x$ and $y$ separately, then two separate surfaces can represent their displacements $d_x$ and $d_y$ respectively. Logically by moving to 3D space, another hypersurface for $d_z$ will be added and so on with each increasing dimension.
Considering the deformation depicted in Figure \ref{Figure:smile}, we define 6 points: 1 in an eye, 4 in the image corners and 1 in the smile corner. 
There is no displacement at the corners, the $d_x$ and $d_y$ surfaces have no height. In this use case, the smile corner moved upwards, so the surface $d_y$ should have a considerable height, whereas the surface $d_x$ should have nearly no height. The eye position was moved to the left, which means that the surface $d_x$ should have a negative height while the surface $d_y$ height should be close to 0. The thin plate spline deformation is depicted in Figure \ref{Figure:surface}. 
\\
We can clearly conclude that $d_x$ and $d_y$ are then only parameters we need to perform deformation and finding the transformation parameters of two point sets means finding these two surfaces if we are working in 2D. In our problem we would need to find the three surfaces $d_x$, $d_y$ and $d_z$ to solve the registration problem.
\subsection{Robust Point Matching Algorithm Description}
\subsubsection{A binary linear assignment-least squares energy function}
\begin{figure}[h]
	\caption{An example of the binary correspondence matrix.}
	\label{Figure:corrmatrix}
	\centering
	\includegraphics[scale=0.8]{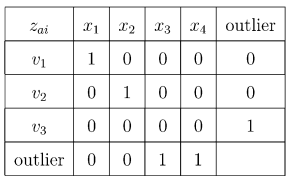}
\end{figure}

First, let’s reconsider $V$ and $X$ defined in section \ref{probstate}, two point sets in 2 dimensional space (For simplicity of explanation). They consist of the points $\lbrace v_a \vert a = 1, 2, \ldots, N \rbrace$ and $\lbrace x_i \vert i = 1, 2, \ldots, K \rbrace$, respectively. We will represent the nonrigid registration by the function $f$. Given any point $v_a$ in 2D space it can be mapped to a new point $u_a = f(v_a)$. Of course, $f$ in this case englobes rotation, translation and scaling and this function is a simplification of function \ref{probstate}. The authors then introduce an operator $L$ to define the smoothness measure $||Lf||^2$ which will help place appropriate constraints on the mapping.
The correspondence problem is cast as a linear assignment problem. Therefore, the goal is to minimize the following binary linear assignment-least squares energy function :
\begin{equation}
\label{equation:energy1}
 \min_{Zf}E(Z,f) = \min_{Zf}\sum_{i=1}^{K}\sum_{a=1}^{N}z_{ai}||x_i-f(v_a)||^2 + \lambda||Lf||^2-\zeta\sum_{i=1}^{K}\sum_{a=1}^{K}z_{ai}
\end{equation}
With $Z$ or ${z_{ai}}$ being the binary correspondence matrix consisting of two parts: (1) The last column and last row handle outliers and (2) the inner part is either 1 or 0, the former in the case of correspondence between $v_a$ and $x_i$ and the latter if no correspondence exists. Naturally, the correspondence is always one-to-one so the matrix $Z$ has to satisfy the following summation constraints,
$\sum_{i=1}^{K=1}z_{ai}=1$ for $a \in {1,2,..,K}$, $\sum_{a=1}^{N+1}z_{ai} =1$ for $i \in {1,2,..,N}$, and $z\{a_i\} \in$ \textbraceleft$0,1$\textbraceright. An example of $Z$ is given in Figure \ref{Figure:corrmatrix}, where points $v_1$ and $v_2$ correspond to $x_1$ and $x_2$, respectively, and the rest of the points are outliers. Note that the existence of an extra outlier row and outlier column makes it possible for the row and column constraints to always be satisfied.
The second term of the equation is the constraint on the transformation. The third and final term is the robustness control term preventing rejection of too many points as outliers. The parameters $\lambda$ and $\zeta$ are the weights that balance these terms. In this way, the point matching objective function (Equation \ref{equation:energy1}) consists of two interlocking optimization problems: a discrete linear assignment problem on the correspondence and a least-squares continuous one on the transformation. When considered separately, both problems have unique solutions. It is their combination that makes it difficult to match the non-rigid point problem. To solve these two problems, the authors propose an alternating algorithm in which the first step estimates correspondence and the second one calculates the transformation. Solving in this manner while using binary correspondences in $Z$ is not meaningful, therefore they use fuzzy correspondence until the algorithm starts to converge. Once it nears a reasonable solution correspondences switch to binary values. To achieve this, the authors utilize two techniques we will describe in the following paragraph.
\paragraph{Deterministic annealing and Softassign}
Softassign relaxes the binary correspondence matrix $Z$ to a fuzzy(continuous) matrix $M$ in the interval [0,1], yet keeps the column and row constraints using iterative normalization on both of them. From an optimization point of view, fuzzy correspondences will allow the defined energy function to behave better because correspondences will improve gradually instead of jumping around in binary space.
Correspondingly, deterministic annealing [18,45] is used to control this fuzziness. It adds the following entropy term $T\sum_{i=1}^{K=1}\sum_{a=1}^{N+1}m_{ai}logm_{ai}$ to the energy function in Equation \ref{equation:energy1}. T is called the temperature parameter because similar to physical annealing as we minimize the value of T the energy function \ref{equation:energy1} will be minimized. The higher the entropy term, the fuzzier the correspondences are. Each minimum obtained at a certain temperature will serve as a starting point for the next stage as the temperature is gradually lowered.
\paragraph{A fuzzy linear assignment least squares energy function}
These two techniques will transform the binary equation described in Equation \ref{equation:energy1} to the following fuzzy-assignment-least squares energy function:
\begin{multline}\label{equation:energyfuzzy}
	E(M,f)=\sum_{i=1}^{K}\sum_{a=1}^{N}m_{ai}||x_i-f(v_a)||^2+\lambda||Lf||^2\\+T\sum_{i=1}^{K}\sum_{a=1}^{N}m_{ai}logm_{ai}-\zeta\sum_{i=1}^{K}\sum_{a=1}^{N}m_{ai}
\end{multline}
where $m_{ai}$ satisfies the same sum constraints that were present on $z_{ai}$.
\\We can deduce that when $T$ becomes 0 the equation reverts to Equation \ref{equation:energy1} and accordingly $M$ reverts back to the binary matrix $Z$. An algorithm that alternatively solves for correspondence then for transformation reducing the temperature at each step has proven successful in rigid registration but two problems arise in the nonrigid case. First, there is no clear way to estimate the outlier control parameter $\zeta$. Since we do not suffer from the outlier problem, we have decided to set this parameter to zero thus reducing Equations \ref{equation:energy1} and \ref{equation:energyfuzzy} to the following :
\begin{equation}
\label{equation:energy2}
\min_{Zf}E(Z,f) = \min_{Zf}\sum_{i=1}^{K}\sum_{a=1}^{N}z_{ai}||x_i-f(v_a)||^2 + \lambda||Lf||^2
\end{equation}
\begin{equation}
\label{equation:energyfuzzy2}
E(M,f) = \sum_{i=1}^{K}\sum_{a=1}^{N}m_{ai}||x_i-f(v_a)||^2+\lambda||Lf||^2+T\sum_{i=1}^{K}\sum_{a=1}^{N}m_{ai}logm_{ai}
\end{equation}
Second, setting $\lambda$ for the prior smoothness term can be difficult as on one hand, the transformation can turn out to be too flexible at small values of $\lambda$. On the other hand, large values of $\lambda$ greatly limit the range of nonrigidity of the transformation. The authors decided to gradually reduce $\lambda$ via an annealing schedual by setting \\$\lambda = \lambda_{initial} \times T$. The larger $\lambda$ is the more global and rigid registrations are favored, and as it decreases with the temperature, more local and nonrigid transformation are calculated.
\paragraph{The robust point matching (RPM) algorithm}
It is essentially a dual update process, we will now describe these two steps after applying our modifications to the algorithm.\\
\textbf{Step 1} Update correspondance for points $i= 1,2,..,K$ and $a= 1,2,..,N$ in the fuzzy correspondance matrix $M$ using the following equation:
\begin{equation}
m_{ai}=\frac{1}{T}exp(-\frac{(x_i-f(v_a))^T(xi-f(v_a))}{2T})
\end{equation}
Then we run the normalization algorithm on both the row and columns to satisfy the constraints until convergence is reached:
\begin{equation}
m_{ai}=\frac{m_{ai}}{\sum_{b=1}^{N+1}m_{bi}},\;i=1,2,\ldots,K
\end{equation}
\begin{equation}
m_{ai}=\frac{m_{ai}}{\sum_{j=1}^{K+1}m_{aj}},\;a=1,2,\ldots,N
\end{equation}
\textbf{Step 2} Update Transformation: After dropping the terms independent of $f$, the following least square problem needs to be solved, 
\begin{equation}\label{eq:pseudocode}
	\min _{f} E(f)=\min _{f} \sum_{i=1}^{K} \sum_{a=1}^{N} m_{a i}\left\|x_{i}-f\left(v_{a}\right)\right\|^{2}+\lambda T\|L f\|^{2}
\end{equation}
The solution to this least squares problem depends on the particular form of the non-rigid transformation. We're going to discuss the solution for one form in the next sectionm, the thin-plate spline.\\
\textbf{Annealing} : We previously described an annealing scheme to control the energy function. $T$ is initiated to a temperature $T_0$ and is gradually reduced via an linear annealing schedual, $T_{new} = T_{old} \times r$ ($r$ is the annealing rate). This is performed until a previously set temperature $T_{final}$ is reached.\\
The parameters are chosen as follows: (1) $T_0$ is set to the largest square distance of all point pairs, (2) $r$ is set to 0.93 so the algorithm is slow enough to be robust, yet not too slow and (3) $T_{final}$ is chosen as $\frac{T_0}{100}$.
\subsection{Integrating TPS in RPM}
In the previous section, we defined the nonrigid transformation parameter in our energy function as $f$. We will now discuss the authors’ chosen form of nonrigid registration; Thin Plate Splines \cite{24792}. In section \ref{section:tps} we described how thin plate splines can be used to describe deformation. In this section, we will describe its integration in the robust matching algorithm.\\
Since nonrigidity allows for multiple mappings between the two point sets. The second term of the energy function (smoothness measure) is added to constrain the mapping between two point sets, it avoids mappings that are too arbitrary. One of the simplest measures is the space integral of the square of the second order derivatives of the mapping function. This leads us to the thin plate spline. Therefore, finding the TPS that fits the two point sets is equivalent to solving the following least squares energy function:
\begin{equation}\label{equation:intermediate}
E_{TPS}(f)=\sum_{a=1}^{N}||y_a-f(v_a)||^2+\lambda\int         \int[(\frac{\delta^2f}{\delta x^2})+2(\frac{\delta^2f}{\delta x\delta y})+(\frac{\delta^2f}{\delta y^2})]dxdy
\end{equation}
There exists a unique minimizer $f$ which comprises two matrices $w$ and $d$:
\begin{equation}\label{eq:lastone}
f(v_a,d,w)=v_a.d+\phi(v_a).w
\end{equation}
where $d$ is a $(D+1)\times(D+1)$ matrix representing the affine transformation and $w$ a $K\times (D+1)$ warping coefficient matrix representing the non-affine deformation.\\
Replacing \ref{eq:lastone} in \ref{equation:intermediate} yields the following energy function:
\begin{equation}\label{eq:becomes}
E_{TPS}(d,w)=||Y -Vd - \phi w||^2 + \lambda trace(w^t\phi w)
\end{equation}
where $Y$ and $V$ are just concatenated versions of the point coordinates $y_a$ and $v_a$, and $\phi$ is a $K\times K$ matrix formed from $\phi (v_a)$\\
Finding the least squares solution for equation x can be cumbersome. Instead, the authors perform a QR decomposition to separate affine and non affine warping space:
\begin{equation}
V = |Q_1Q_2|
\begin{pmatrix} 
R& \\
 & 0\\
\end{pmatrix}
\end{equation}
Equation \ref{eq:becomes} becomes:
\begin{equation}
	E_{TPS}(d,w)= ||Y-Vd-\phi w||^2 + \lambda _1 trace(w^T\phi w)+\lambda _2 trace[d-I]^T[d-I]
\end{equation}
A detailed mathematical explanation of the final equation can be seen in section 4 of \cite{CHUI2003114}.
Minimizing this equation solves the registration problem. For the same reasons explained before, $\lambda_1$ and $\lambda_2$ follow an annealing schedule. We set $\lambda_1$ to 1 and $\lambda_2$ to 0.01 so as to always favor affine transformations. Each temperature is calculated during five iterations before it is decreased. This is done upon reaching a global solution.
\subsection{Problem with correspondence}
After implementing the TPS-RPM algorithm described in the previous sections, we ran into some problems regarding the correspondence calculation step. Even though we took measures to limit the registration algorithm to only 9 degrees of freedom (choosing optimal temperature values), on some rare occasions, it calculated the correspondences in some way that requires skewing to be possible. As we have already discussed, this is not possible in Unity. Even if the transformation is technically 100\% accurate, in Unity it will seem wrong since the skew and the rotation will be applied as if they were just a rotation. Since the algorithm would have searched for an optimal restricted transformation (9 DoF) before as a last resort, searching for a fully nonrigid one. The calculated correspondences did not permit for a restricted transformation to be possible. We know that there exists another correspondence between the points that only needs rotation, translation and scaling to be optimal. Therefore in this section, we will find a solution that always calculates the correct correspondences in those rare instances when the algorithm fails.
\subsubsection{Proposed Solution}
There are three invariants in our problem: (1) there are no outliers between the two point sets, (2) the user will choose 4 points to define the three dimensions (height, width, depth) and (3) there is always a certain one-to-one correspondence between the the points that allows for a skew-less transformation. In our test cases, the probability of the algorithm choosing a correspondence that contains skewing is almost 5\%. That is because the parameters $T$ and $\lambda$ have been carefully chosen. However, even a 95\% success rate is unacceptable in an application whose main objective is to reduce time and effort in labeling data sets. Henceforth, we need to find a solution that not only provides an almost perfect accuracy but also does it without losing the tool's ``real-time" property.\\
Since the user always choses four points, then at most there are $4! = 24$ possible one-to-one correspondences between the points. Therefore, in the case of the algorithm choosing wrong correspondences, we can start by exploring the 23 other possible correspondences to find the one that minimizes the skew parameter. Hence, after the execution of the normal algorithm, we extract the skew parameter from the calculated $4\times4$ transformation matrix and test if it is significant or not. In our testing we found that testing if it is larger than 0.1 is the best option. If this condition is met then we have to calculate the transformation at each of the 23 correspondences and choose which one minimizes the standardized euclidean distance, $c_dist$, between the result and the target chosen by the user to calculate the result we simply apply Equation \ref{probstate}. Therefore, we need to keep track of the smallest $c_dist$ achieved; let's call the variable that does this bookkeeping $min_d$. For each correspondence, we test the new found $c_dist$ value to the smallest one calculated to date and replace it in the case it is smaller, otherwise we keep it the same. When the algorithm is completed, the transformation resulting from the correspondence that had the smallest $c_dist$ is the one returned. Keep in mind that for any transformation's $c_dist$ to be eligible of being compared to $min_d$, its skew has to be lower than 0.1.
\subsubsection{Improvements}
The first improvement we can add is setting a breaking condition where, we no longer try the rest of the possible correspondences. After checking that the skew is lower than 0.1, if the $c_dist$ is lower than a certain threshold. Testing cases showed that a decent value of the error threshold is 0.15.\\
The second improvement is parallelizing the full process. By parallelizing we mean launch a separate thread that evaluates each correspondence separately. One thing we have to keep in mind is for the variable $min_d$ that keeps track of the smallest $c_dist$, concurrency should be handled as it is checked and possibly changed by each thread. We protect this concurrency by applying a fundamental concept in parallel computing, the \textit{Semaphore}. We use a simple semaphore that we initialize to the value 1; it is acquired by each thread to check the condition on $min_d$: if the condition is true $min_d$ is updated and then the semaphore is released, otherwise the semaphore is directly released. In the next section, we will provide the pseudo-code explaining the proposed solution and its improvements.
\subsection{Algorithm Pseudo-Code}

\begin{algorithm}[H]\label{alg:corr}
	\SetAlgoLined
	\KwResult{ $4\times 4$ Transformation Matrix }
	Initialize parameters $T$ , $\lambda_1$, $\lambda_2$ and $M$\;
	Initialize parameters $d$ and $w$\;
	Begin A: Deterministic Annealing:\\
	\Indp Begin B: Alternating Update:\\
	Update transformation parameters $(d,w)$ using \ref{eq:pseudocode}.\;
	End $B$\;
	\Indm Decrease $T$ , $k_1$; and $k_2$.\;
	End $A$\;
	Acquire mutex\;
	\If{skew$\textless$ 0.1}{\If{$c_dist$ $\textless$  $mean_d$}{$mean_d$ = $c_dist$\\optimal\_trans = calculated\_trans\;}}
	Release mutex\;
	\caption{TPS-RPM assuming given correspondences (launched in threads)}
\end{algorithm}

\begin{algorithm}[H]\label{alg:tpsrpm}
		\SetAlgoLined
	\KwResult{ $4\times 4$ Transformation Matrix }
	Initialize parameters $T$ , $\lambda_1$ and $\lambda_2$\;
	Initialize parameters $M$, $d$, and $w$\;
	Begin A: Deterministic Annealing:\\
	\Indp Begin B: Alternating Update:\\
	Step I: Update correspondence matrix $M$ using (3)–(5).\;
	Step II: Update transformation parameters $(d,w)$ using Equation \ref{eq:pseudocode}.\;
	End B\;
	\Indm Decrease $T$ , $k_1$ and $k_2$.\;
	End A\;
	\If{skew \textgreater 0.1}{
		Initialize global parameters $mean_d$, mutex, optimal\_trans\;
		 \For{$i\gets0$ \KwTo $23$}{
			LaunchThread(Algorithm\ref{alg:corr}(correspondance($i$)))\;
			\If{$mean_d$ \textless 0.15}{\textbf{return} optimal\_trans}
		}
	}
	\textbf{return} optimal\_trans
	\caption{Original TPS-RPM algorithm without outlier calculation in M}
\end{algorithm}
\subsubsection{Analysis}
The authors claim that TPS-RPM has a worst case runtime complexity of O($n^3$). After our modifications, in the worst case, we will call a simplified (given correspondence) TPS-RPM at most 23 more times. Thus, on a single threaded machine or if the loop is not broken by any thread, the worst case time complexity is O($24*n^3$), which is rendered asymptotically to O($n^3$). As a result, even in the most unlikely of cases, our modifications preserve the asymptotic run time complexity of the original algorithm is preserved. Consequently, we have preserved the speed of TPS-RPM while increasing our registration accuracy to 100\%.
\subsection{Implementation}
Everything regarding the registration calculation was done on the Python server side. The point sets are provided by Unity through websockets. Upon retrieval the Python server executes Algorithm \ref{alg:tpsrpm} and returns its output to Unity. The latter applies it on the labeling object to transform it to the correct position. We implemented TPS-RPM from scratch and modified it with the aforementioned improvements.\\
The returned transformation matrix has the following form:
\begin{equation}
	\begin{pmatrix}
	s_x*r_{11}&s_y*r_{12}&s_z*r_{13}&t_x\\
	s_x*r_{21}&s_y*r_{22}&s_z*r_{23}&t_y\\
	s_x*r_{31}&s_y*r_{32}&s_z*r_{33}&t_z\\
	0&0&0&1\
	\end{pmatrix}
\end{equation}
where $s_x,s_y$ and $s_z$ are the values of $S$ in the $x,y$ and $z$ axes respectively, $t_x,t_y$ and $t_z$ are the values of $T$ in the $x,y$ and $z$ axes respectively and $r_{ij}$/i,j$\in$$[1,3]$ are the values of $R$.

\section{Colliders in Unity3D}
The points that are being selected are primitive objects of type spheres. To be able to add the point on the desired object, the concept of colliders should be introduced so that the point (sphere object) can collide with the corresponding object and be attached to it.\\
In Unity3D, there’s a possibility to add a “Collider” component on an object, that defines the shape of the object for the purpose of physical collisions, the collider is invisible and should be the exact same shape as the object to produce accurate collisions with other objects. There are multiple types of colliders already implemented whether for 3D or 2D use. In this application, three types were used:
\begin{enumerate}[noitemsep]
	\item \textbf{Box Colliders:} This component was added to the labeling objects.
	\item \textbf{Sphere Colliders:} It was attached to the spheres that are eventually added as a set of points on the labeling object.
	\item \textbf{Mesh Colliders:} This type was only attached to complex meshes, including labeling object imported as CAD objects.
\end{enumerate}

\begin{figure}[!hbt]
	\centering
	\includegraphics[scale=0.2]{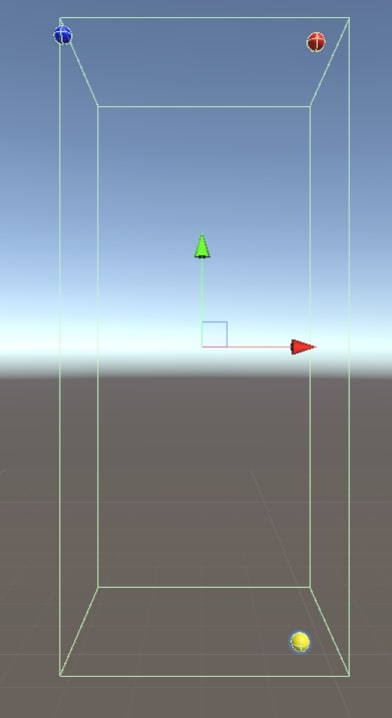}
	\caption{Box and Sphere Colliders}
	\label{fig:boxcollider}
\end{figure}

\begin{figure}[hbt!]
	\centering
	\includegraphics[scale=0.5]{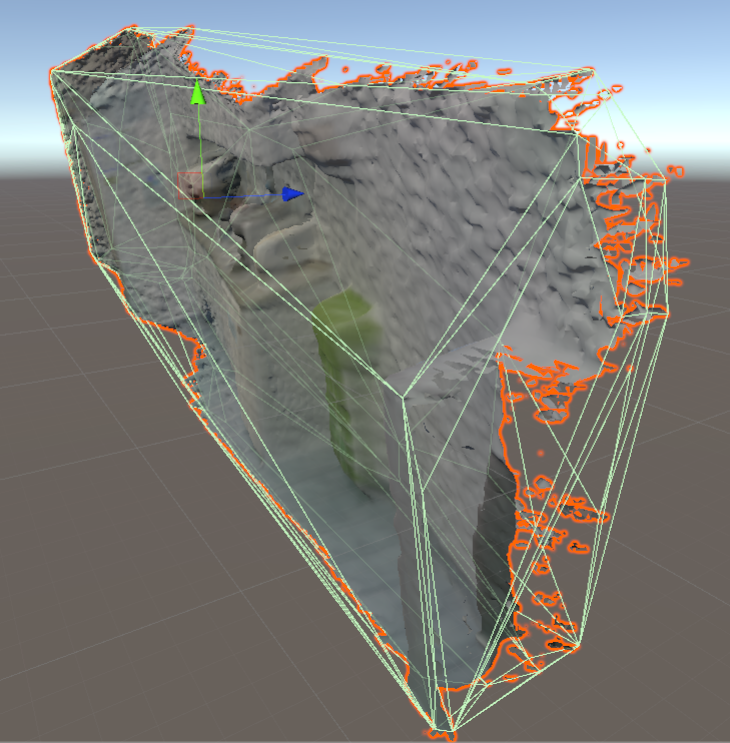}
	\caption{Convex Mesh Collider}
	\label{fig:convexmeshcollider}
\end{figure}

\noindent As a result, the points can now be attached to the labeling objects as shown in Figure~\ref{fig:boxcollider}. Note that, in this figure the objects are not rendered, only for the purpose of showing the sphere and box colliders (in green color). After the points are attached to the labeling object, they are assigned as children of this object; therefore, their coordinate system is now referenced to the object and not Unity’s world coordinate system. That being said, the coordinates of these points can now be sent to the server and the algorithm can work efficiently to return the transformation of this labeling object.\\
Furthermore, to select the points on the target object, a mesh collider should be attached to the imported mesh. In Unity3d, the mesh collider component takes a mesh as an input and builds the collider based around that mesh. However, to a add a collider on a specific mesh, that latter must contain a number of vertices which is less than or equal to $65536$.\\
Two solutions were proposed, either adding multiple primitive colliders to the mesh or simplifying the mesh so that each time it is loaded the number of vertices is simplified to the desired amount that Unity3D accepts. However, the first solution was somehow impossible, hence the complexity of the mesh and the fact that it is loaded as one object that is not segmented into sub-meshes. The second solution was feasible, but with the condition of keeping the imported rendered mesh with the same high number of vertices and only simplify the mesh that has to be taken as an input for the collider. In addition to the two solutions already mentioned, mesh colliders in Unity3D possess a convex property that could be enabled or disabled depending on whether the developer wants to add the option of colliding with other mesh colliders. However, one of the drawback of using this option is that it only supports up to 255 triangles and it is obviously not suitable for this mode's use case. Figure~\ref{fig:convexmeshcollider} displays a convex mesh collider attached to one of the test meshes that have been used. In fact, it is clear that this property limits the details of the collider and thus provides a poor experience for the user trying to set the points on a specific part of the mesh. For example: the green trash bin showing in Figure~\ref{fig:convexmeshcollider}.
Therefore, applying a simplification algorithm on the mesh and feeding it to the collider for maximum collision accuracy, was the chosen solution to tackle the given problem. 

\subsection{Mesh Simplification}
\subsubsection{Background}	
In recent years, the problem of surface simplification has received a lot of attention. Many algorithms have been formulated to solve this problem and they can be categorized into 3 classes:

\begin{enumerate}[]
	\item \textbf{Vertex Decimation:} It is based on iteratively selecting a vertex to be removed, removing the adjacent faces and re-triangulating the resulting hole.  
	\item \textbf{Vertex Clustering:} This algorithm is based on creating a bounding box around the mesh and dividing it into a grid containing different cell. Within each cell, the vertices are clustered together into one vertex and the faces are updated accordingly. This method could be very fast but it provides a quality that is often quite low. A solution for this problem was later on introduced, explaining that this method could be generalized to use an octree\footnote{An octree is a tree data structure in which each internal node has exactly eight children.}.
	\item \textbf{Iterative Edge Contraction:} This method consists of simplifying the model by choosing an edge and collapsing it (see Figure~\ref{fig:edgecollapse}). Many algorithms implemented this idea and the real difference between them is the way that they chose the edge that is going to be contracted. 
	
	\begin{figure}[!hbt]
		\centering
		\includegraphics[]{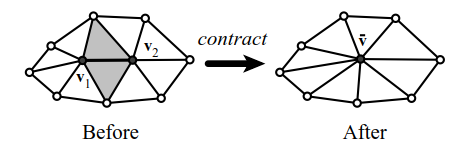}
		\caption{Edge Collapse}
		\label{fig:edgecollapse}
	\end{figure}
\end{enumerate}
\noindent
Micheal Garland and Paul S.Heckbert~\cite{garland1996} discuss the limitations of the algorithms already mentioned, by saying that these 3 classes of algorithms don't provide a high quality approximation (Vertex Clustering), nor general (Vertex Decimation) nor supports aggregation. However, they developed an algorithm that supports all these 3 properties at the same time. This method is based on decimation via pair contraction which is a general form of edge contraction. Nevertheless, since the algorithm uses edge collapse, a criterion should be added to select an edge. Given two vertices, $v1$ and$v2$, a pair $(v1,v2)$ is a valid pair for collapsing if:
\begin{itemize}[noitemsep]
	\item $(v1,v2)$ is an edge (see Figure~\ref{fig:edgecollapse}), or
	\item $|v1-v2| < \epsilon$, where $\epsilon$ is a user defined constant that should be carefully set. High threshold could enable vertices that are far from each other to collapse which will ruin the topology of the model. (see Figure~\ref{fig:nonedgecollapse})
	
	\begin{figure}[!hbt]
		\centering
		\includegraphics[width=0.5\linewidth]{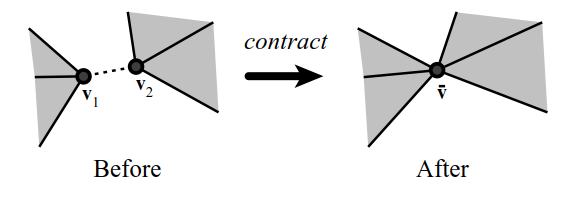}
		\caption{Non-Edge Collapse}
		\label{fig:nonedgecollapse}
	\end{figure}
\end{itemize}
In addition, to select the appropriate contraction at a given iteration, this algorithm introduces a notion of \textit{cost} of a contraction. This cost is defined as the error at each vertex $\Delta (v) = v^{T}Qv$, where $Q$ is a $4\times4$ symmetric matrix assigned to each vertex. Note that, since $Q_{v}$, is $4\times4$ matrix $\Delta V = \delta$, is the set of all points whose error with respect to $Q$ is $\delta$ which is a surface of second degree in $v$ (quadric surface). Hence, this error is referred to as the quadric error metric. In fact, $Q_{v} = \sum{M_{P}(v)} $ where $ P = <a,b,c,d>$ is a plane that contains an incident triangle of $v$ and $M_{P}$ is equal to the following matrix:

\begin{equation}
\centering
M_{P} = PP^{T} =
\quad
\begin{bmatrix}
a^{2} & ab & ac & ad \\
ab & b^{2} & bc & bd \\
ac & bc & c^{2} & cd \\
ad & bd & cd & d^{2}
\end{bmatrix}
\quad
\end{equation}

\noindent
Furthermore, after the pair of vertices is chosen, a simple way to collapse is to move $v1$ to $v2$ or vice versa, or to $(v1+v2)/2$. However, this algorithm introduces a better way to collapse the vertices by moving to a new point $v$ that minimizes the error $\Delta (v) = (v^{T}Q_{v1}v + v^{T}Q_{v2}v)/2 = (v^{T}(Q_{v1} + Q_{v2})v)/2 $. After $v$ is computed, (v1,v2) is collapsed into $v$ with the error value $\Delta (v)$ and an error metric matrix of $Q_{v1} + Q_{v2}$. Moreover, finding the minimum of a quadratic function is a linear problem and to find the new vertex $v$ we need to solve the following equation: 
\begin{equation}
\centering
\delta\Delta/\delta x_{v} =\delta\Delta/\delta y_{v} = \delta\Delta/\delta z_{v} = 0 .
\end{equation}

\subsubsection{Algorithm}
\begin{algorithm}
	\SetAlgoLined
	Compute the error value and error matrix for each vertex of the mesh\;
	Select all valid edges $(v1,v2)$ such that $|v1-v2| < \epsilon$ \;
	\While{an edge $(v1,v2)$ exists}{
		Minimize $\Delta (v) = (v^{T}(Q_{v1} + Q_{v2})v)/2 $ to find v\;
		Let $\Delta (v) = (\Delta (v1) + \Delta (v2))/2$ and $Q_{v} = Q_{v1} + Q_{v2}$ \;
		Place all selected edges in a heap using $\Delta (v)$ as a key\;
	}
	\While{Heap contains edges}
	{
		Remove the top edge $(v1,v2)$\;
		Collapse it to the computed v\;
		Update the mesh and the keys\;
	}
	\caption{Mesh Simplification Using Quadric Error Metrics}
\end{algorithm}

\subsubsection{Implementation in Unity}
This algorithm is already implemented by Mattias Edlund~\cite{meshsimp2018} and configured to be able to work in Unity projects. It introduces two functions that simplify the mesh, one with a quality parameter and the other without one. In our implementation, we used the function that gave the opportunity to change the target quality of the mesh, because it makes the target vertex count more controllable than the one with no parameter. Moreover, this parameter is a number between 0 and 1 that is multiplied by the number of triangles to provide a threshold for the main iterations. The function returns the simplified mesh when the original number of triangles subtracted by the number of deleted triangles is less than or equal to the threshold.\\
\begin{figure}[!hbt]
	\centering
	\includegraphics[scale=0.5]{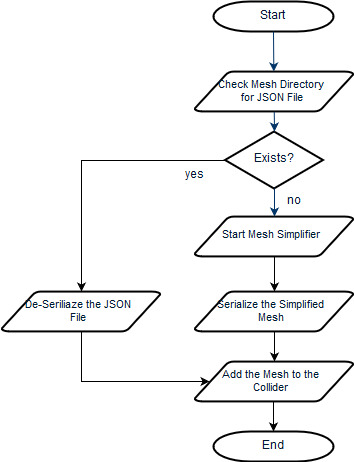}
	\caption{Loading Collider Flowchart}
	\label{fig:flowchart_collider}
\end{figure}
However, after integrating this method into the tool and testing it, we started changing the quality by trial and error and deduced the resulting number of vertices by making sure that the vertex count is always less than or equal to $65536$. After the testing phase, we concluded that the mesh vertices are always varying and could lead to a simplified mesh that could not be loaded as a collider in Unity. As a pre-processing method before loading the data into the tool, the mesh is simplified so the vertex count is $\sim$ 1 million.\\
\begin{figure}[!hbt]
	\centering
	\includegraphics[scale=0.6]{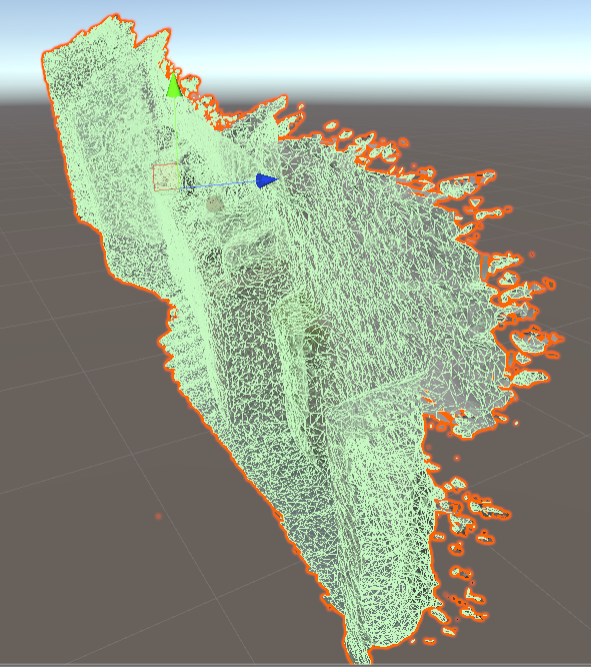}
	\caption{Concave Mesh Collider}
	\label{fig:concavemeshcollider}
\end{figure}
Furthermore, to optimize the workflow of loading the collider when entering Phantom Mode, we serialize the simplified mesh into a JSON file saved in the same folder of the input mesh. Hence, the mesh will be simplified once, and then it will be de-serialized and loaded much faster. The example in Figure~\ref{fig:concavemeshcollider} shows the same mesh of Figure~\ref{fig:convexmeshcollider} but after using the mesh simplifier and adding the concave mesh collider. It shows the high level of details that could be reached in the mesh in case of trying to collide with it. In conclusion, this workflow (see Figure~\ref{fig:flowchart_collider}) added the possibility to apply Phantom Mode on any loaded mesh, leading to a faster labeling action.	
\section{Websockets}\label{websockets}
WebSocket is a standardized communication protocol which produces a communication over a TCP connection. It is a way to communicate in an asynchronous way between a client which is the Unity software in our case and the server which can be on the same machine or even on another machine. Communication can be done using a secure protocol and can be used by any client or server application.\\
The connection is bidirectional which means the data flows in both ways. The WebSocket API lets the server and the client to push messages to each other at any given time. This is be achieved without having the need to establish a new connection, which allows a real time data flow.

\subsection{Server Implementation}
Before implementing the server side of our application, many technologies were considered to enable a bi-directional communication between the Unity application and the server, that is responsible to compute the main algorithm and send the results back to the client. The first technology that we came across is Sofi, a framework that was developed to provide a “pythonic” GUI by packaging HTML5 libraries, in a way that all the processing is done using web sockets within Python. However, this framework does not suit the use case of our project, knowing that, it doesn’t include any interactions with the web, for the meantime. The same issue was encountered when using the SocketIO technology. SocketIO is a protocol that enables a real time connection between clients and a server, but the client in this case is a web application or mobile application, which leads to multiple complications when implementing a SocketIO server, using the SocketIO Python library, and issuing a connection to the Unity application.\\
Therefore, websockets Python library was used on top of asyncio~\cite{asyncio} to implement the server and ensure a real-time communication with the Unity client. Nevertheless, the asyncio library provides the ability to write asynchronous programs using the async/awaits syntax. With asyncio programming there is no parallelism, the program is running on a single thread. However, when the functions’ execution is pending, Python can run other functions and proceed when they all have what they need to continue their execution. These functions are called Coroutines, they are declared with async/await syntax placed before the function, for example: “async def main():”. When a coroutine runs, it generates a task that executes till the end, except if it encounters an await, then it gets suspended and runs the other coroutine. The asyncio.get\_event\_loop() will return the suspended task to continue its execution. The websockets library provides the program with the server’s send and receive functions, using the websockets.server module that defines the WebSocket server API. This module includes a serve function that takes as parameters: the coroutine to be executed, the host and the port on which the server is listening, and returns an awaitable object that is handled by the asyncio library using the asyncio.get\_event\_loop().run\_until\_complete (the awaitable object).\\
The server implemented in Python, listens on port 4444 and returns an awaitable that executes the coroutine. In this function, the server awaits data from the connected client using the coroutine websocket.recv(), handles the data and then sends it back to the client using the coroutine websocket.send(). To keep the server on and listening at all times, the asyncio.get\_event\_loop().run\_forever() function that runs the main coroutine indefinitely was used.\\
The client sends two $Nx3$ matrixes, one for the source points and the other for the target points, where $N$ signifies the number of points selected. At first, the client sends the source and target matrixes separately as a chain of characters where the points are separated by a “/”, and the coordinates by a comma. On the server side, each sent matrix was handled by splitting the points and the coordinates to be able to create the arrays and pass them to the main algorithm. The main registration algorithm returns a $4\times4$ matrix, this latter is also transformed into a chain of characters and sent to the client.\\
Moreover, to ensure that the computations done by the server are fast and reliable, the server code was implemented on another machine. The client is connected to the same network on which the server is on, and for the websockets.server.serve function, the argument “None” that maps to the host was passed. By that, the server was capable of receiving and sending data to the client on the port 4444.

\subsection{Message Structure}
\begin{figure}[!hbt]
	\centering
	\includegraphics{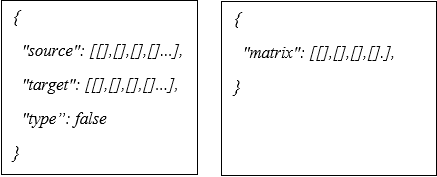}
	\caption{Message Structure}
	\label{fig:messagestructure}
\end{figure}

\begin{figure}[!hbt]
	\centering
	\includegraphics{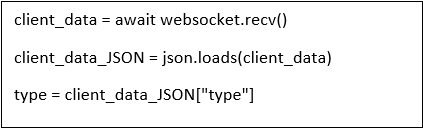}
	\caption{Python JSON handling}
	\label{fig:pythonjson}
\end{figure}

JSON is a lightweight data-structure that is easy to read and to parse by the machines, this data structure is written in key/value pairs. The format used to represent the data is seen in Figure~\ref{fig:messagestructure}.
On the server side, the json Python library provided functions to parse the JSON sent by the client. The json.loads() function is used to convert the chain of characters taken as an argument, into a JSON object that could be easily manipulated by calling the keys of the JSON format (see Figure~\ref{fig:pythonjson}).

\subsection{Client Implementation}
In our project, multiple data must be sent to the server, so it can do the calculations on the backend. Then, the client will receive back the $4\times4$ matrix containing necessary information for the translation and rotation. There are two cases, either the client wishes to perform a rotation and translation normally, or they wish to only perform a rotation using the normal of the points.\\
In the first case, the user can choose as much points as they wisheon the source and target objects. Then, these points will be sent to the server in the corresponding order using the WebSocket in a JSON format, specifying the type of transformation requested by the user, along with the source and target points, so it can easily be interpreted by the server.
In the second case, the user should choose three points on two objects, the source and target. The program will calculate the normal of each object using the chosen points and send these two vectors with the type of task to the server.\\
Moreover, the client should establish a connection using the server's IP and port number. Then, after clicking on an execution button to activate one of the two transformation methods, the source and target points can each be sent separately using only one connection. Nevertheless, JSON englobes the data in one block, thus, the content of the data is easily extracted using JSON’s functions. For futures tasks, JSON increases the scalability: if more types of transformations or variables are to be added, it could be simply included as a JSON key/value.\\
For the client’s implementation, WebSocketSharp library~\cite{websocketsharp} was used, which is a subset of the Socket.IO library. This latter can be found in the Unity asset store for free which also contains a library to handle JSON objects. It also works on most platform focusing equally on reliability and speed. For the WebSocket implementation, some basic event handlers should be defined: “On Open”, “On Close”, and “On Message”.
\begin{itemize}
	\item “On Open” event will be activated when the connection between the client and the server is established.
	\item “On Message” event will be called when receiving data to the client socket. After receiving a message, the connection will automatically close, but this can be modified in the library.
	\item “On Close” event will run when the connection to the server ends.
\end{itemize}

\noindent
To send a message, the send() function already implemented in the imported WebSocketSharp library has to be called, followed by the message of type String which will be the JSON message to be sent.\\
Moreover, a JSON object from the received message is created to access its data, but a problem occurred when using the JSON library’s constructor: the accuracy of the data went down from numbers in 62 bits to 32 bits. A solution had to be found since the accuracy plays a major role in moving objects. Since only double types were needed, some unnecessary functions using or converting to float type were removed. After creating the object, the program will access each data in the received message, add them to a $4\times4$ matrix and finally lunch a block of code to run, which will be the start of the transformation.

\section{The Labeling Element Transformation}
After selecting the points on the source and target objects in the right spots with the help of the accurate colliders, the points are sent to the Python server through websockets in a custom format. Then, the server returns a transformation matrix, responsible for the change of position, rotation and scale of the source object onto the target mesh.\\
Knowing that the points were relative to the source object origin, the transformation should be done accordingly. To be able to do that in Unity, we dissected the returned transformation matrix to extract the position, rotation and scale, then we applied each one on the source object so it can be finally transformed to the desired location.

\section{User Interface}
\begin{figure}[h]
	\centering
	\includegraphics[width=\linewidth]{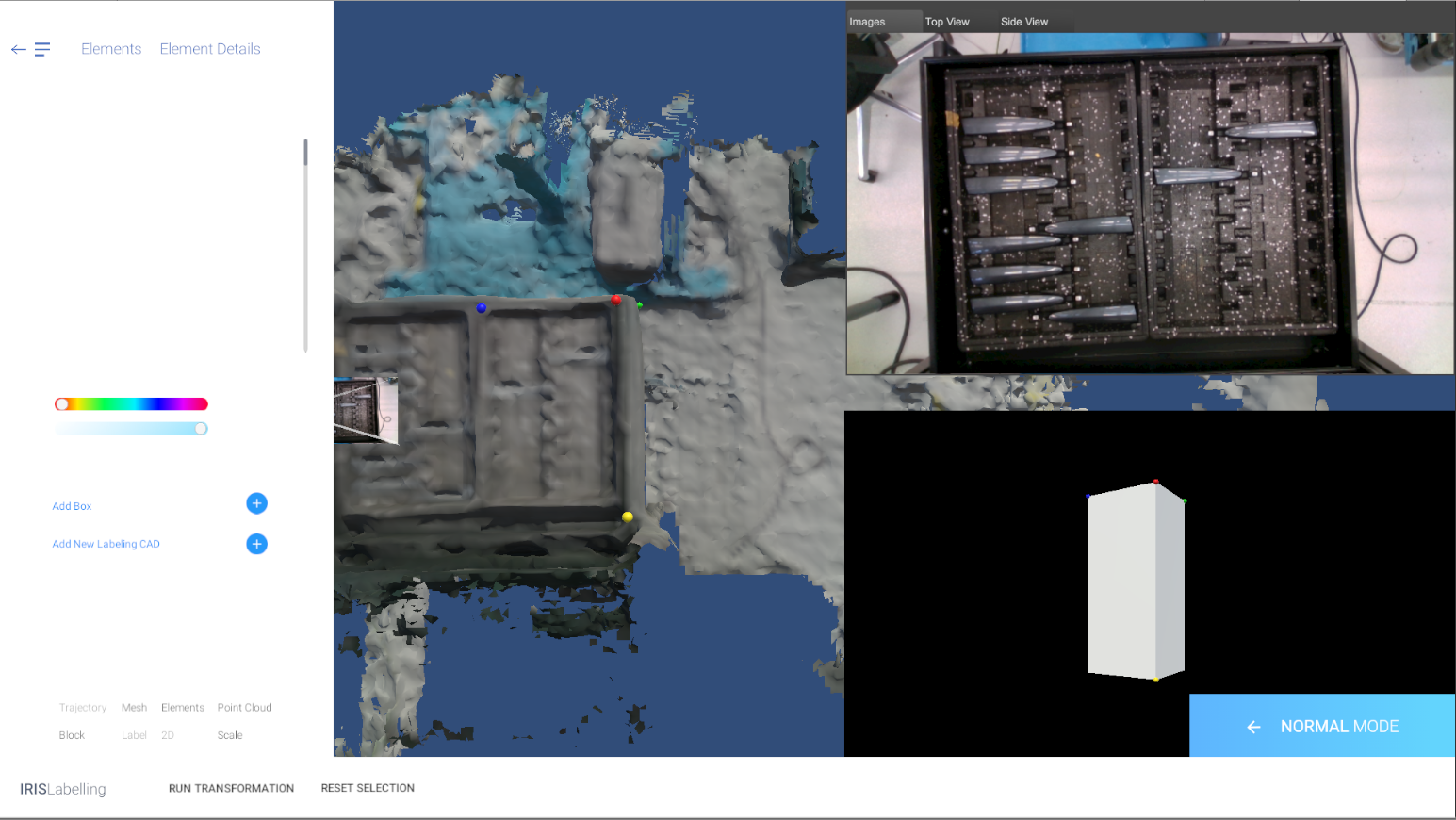}
	\caption{Phantom Mode User Interface}
	\label{fig:phantommodeui}
\end{figure}

For the user interface of this mode, we added a button labeled \textit{Phantom Mode}, takes the user into another panel (see Figure~\ref{fig:phantommodeui}). In this window, we notice that the source points are already been selected on the source object, the user needs only to add the points on the mesh and click on \textit{Run Transformation} in the footer to launch the process.\\
However, the user can still select the custom points on the source target if they click on \textit{Rest Selection} button. In addition, if a point is wrongly added the user can still click on the escape key to revert the last point added. As a result, the user experience became more intuitive. When the user is done using this mode, they can click on \textit{Normal Mode} to go back to the standard labeling process.

\chapter{Labeling Without a Mesh}
	\section{Problem Statement}\label{problem}
	We have already covered the process of using a 3D reconstructed mesh from RGBD images to label said images. However, in certain cases, where we only have access to the RGB images and the camera positions, we cannot reconstruct the mesh. In this chapter, we tackle the case where we label the images without the presence of the mesh.
	\section{Assumption}
	To render this problem feasible, we make a prior assumption that we already know the size of the object we want to label. For instance, if we have a CAD object that represents the object we want to label.
	\section{Theory}\label{theory}
	\subsection{Definition}\label{def}
\begin{figure}[h]
		\caption{Rectangular Cuboid $B$}
		\label{Figure:cuboid}
		\centering
		\includegraphics[width=0.25\textwidth]{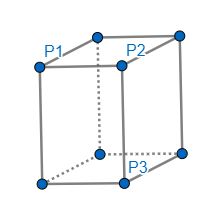}
	\end{figure}

	Let's define a rectangular cuboid $B$ (Represented in Figure \ref{Figure:cuboid}) with specific known dimensions (height, width, depth). Let's take three points on this object (${p_1,p_2,p_3}$). We also define $O$ as a point in space. We can solve the problem in Section \ref{problem} by reducing it to the following problem:\\
	Given three lines ($l_1, l_2,l_3$) created by connecting $O$ and ($c_1,c_2,c_3$) respectively, we need to find the position of three points ($f_1,f_2,f_3$) on ($l_1, l_2,l_3$) respectively while keeping the distances ($p_1p_2, p_2p_3, p_1p_3$) equal to ($f_1f_2, f_2f_3, f_1f_3$).\\
	\begin{figure}[h]
		\caption{Problem Representation}
		\label{Figure:problem}
		\centering
		\includegraphics[width=0.7\textwidth]{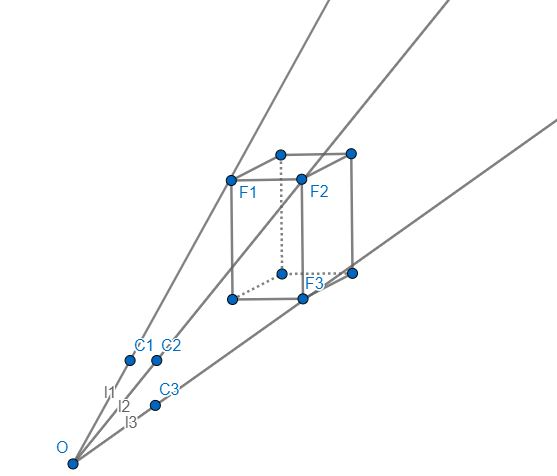}
	\end{figure}
	A geometrical representation of the problem can be found in figure \ref{Figure:problem}.
	\subsection{Problem Breakdown}
	The first thing to consider is that $f_1$, $f_2$ and $f_3$ should each be on $l_1$, $l_2$ and $l_3$ respectively, thus defining the following three constraints: (1) $f_1\in l_1$, (2) $f_2\in l_2$ and (3) $f_3\in l_3$.\\
	The equation (in vector form) of a line in 3D space passing through the point $(x_0,y_0,z_0)$ and traveling in the direction $(a,b,c)$ is:
	\begin{equation}
	(x,y,z)=(x_0,y_0,z_0)+t(a,b,c)
	\end{equation}
	where $t$ is a parameter describing a particular point on the line.
	We can calculate $t$ for the points ($c_1,c_2,c_3$) for each line:
	\begin{equation}
	\begin{cases} t_{c_1} = c_1 - O \\ t_{c_2} = c_2 - O \\ t_{c_3} = c_3 - O \end{cases}
	\end{equation}
	Thus the coordinates of $p_1,p_2$ and $p_3$ are :
	\begin{equation}\label{eq:pointonline}
	\begin{cases} f_1 = O + t_{c_1} * u_1 \\ f_2 = O + t_{c_2} * u_2 \\ f_3 = O + t_{c_3} * u_3 \end{cases}
	\end{equation}
	where $u_1,u_2$ and $u_3$ are the unknowns we need to determine.\\
	Second we define three equations that represent the distance constraint mentioned in section \ref{def}:
	\begin{equation} \label{eq:dist1}
	p_1p_2 = f_1f_2
	\end{equation}
	\begin{equation}\label{eq:dist2}
	p_2p_3 = f_2f_3
	\end{equation}
	\begin{equation}\label{eq:dist3}
	p_1p_3 = f_1f_3
	\end{equation} 
	The distance between two points in 3D space can be calculated using the following equation:
	\begin{equation}\label{eq:dist}
	dist = \sqrt{(x_{1} - x_{2})^2+(y_{1} - y_{2})^2+(z_{1} - z_{2})^2}
	\end{equation}
	If we replace Equation \ref{eq:dist} \& \ref{eq:pointonline} in Equations \ref{eq:dist1}, \ref{eq:dist2} \& \ref{eq:dist3} we get the following non-linear system of equations:
	\begin{equation}\label{eq:system}
	\begin{cases} dist(p_1,p_2) = dist(O + t_{c_1} * u_1,O + t_{c_2} * u_2)\\ dist(p_2,p_3) =  dist(O + t_{c_2} * u_2,O + t_{c_3} * u_3) \\ dist(p_1,p_3) =  dist(O + t_{c_1} * u_1,O + t_{c_3} * u_3) \end{cases}
	\end{equation}
	\subsection{Solving the Equation}
	We can clearly see that equation \ref{eq:system} is a system of three equations with three unknowns($u_1,u_2,u_3$). We can also deduce that it is a non linear system since if we square both sides of the equations we obtain a sum of squares.	We will solve for the three unknowns using a numerical method.
	\subsubsection{Newton-Raphson}
	The Newton-Raphson method is the first in the class of Householder's methods, it is defined by the following equation:
	\begin{equation}
	x_{n+1}=x_n-\frac{f(x_n)}{f'(x_n)}
	\end{equation}
	It can be generalized to solve systems of $k$ nonlinear equations by left multiplying with the inverse of the $k\times k$ Jacobian matrix $J_F(x_n)$ instead of dividing by $f'(x_n)$:
	\begin{equation}
	x_{n+1}=x_{n}-J_{F}(x_{n})^{-1}F(x_{n})
	\end{equation}
	Newton-Raphson is a method of quadratic convergence and hence rapid. However, global convergence is not guaranteed for all initial estimates and for example fails for the two following cases:
	\paragraph{Stationary Iteration point:} Consider the function $f(x) = 1-x^2$, if we start at $x_0 = 0$ then $x_1$ will be undefined since the tangent to a point on the $X$ axis (0,1) is parallel to the $X$ axis: $x_1 = x_0 - \frac{f(x_0}{f'(x_0)} = 0 - \frac{1}{0}$
	\paragraph{Starting point enters a cycle:} Consider the function $f(x)=x^{3}-2x+2$, for a starting point of 0, the algorithm will alternate between 1 and 0 and will never converge.\\
	As we can see we cannot rely on this method directly. We decided to use a method that supports constrained solution searching and use its result as an input to the Newton-Raphson method.
	\subsubsection{DOG\_BOX}
\begin{figure}[h]
		\caption{Bounded Problem $B$}
		\label{Figure:problem2}
		\centering
		\includegraphics[width=0.7\textwidth]{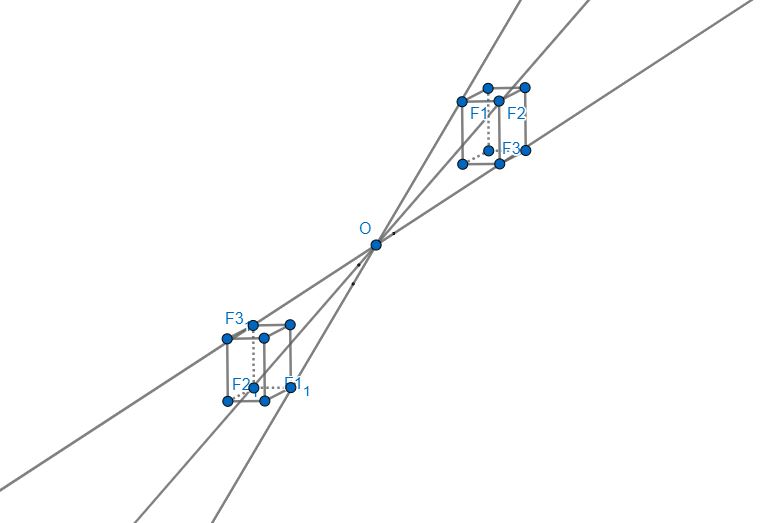}
	\end{figure}

	The DOG\_BOX method \cite{VOGLIS} is a trust region method for solving bound constrained nonlinear optimization problems. Unlike other popular methods that use ellipsoid trust regions, this procedure uses a rectangular shape which is much simpler due to the constraints being linear. It uses a modification of Powel's dogleg technique \cite{POWELL} to find the solutions.\\
	The advantage of this method is that it supports bounds on the unknown variables, which means we can constrain the depth value of the unknowns to be superior to the depth value of $O$. This is done because in our problem the box $B$ can fit either in front of $O$ or behind it, please refer to Figure \ref{Figure:problem2} for a visual representation.
	After we solve the system of non linear equations \ref{eq:system} using the DOX\_BOX method. We input the approximated ($u_1,u_2,u_3$) as a starting estimate to the generalized Newton-Raphson method and use its output as the final solution.
	\subsubsection{Calculating ($f_1,f_2,f_3$)}
	We replace the final values of ($u_1,u_2,u_3$) in equation \ref{eq:pointonline} and get the coordinates of the three points ($f_1,f_2,f_3$). 
	\subsection{Rigid Registration}
		Now that we have the three points ($f_1,f_2,f_3$) on the lines ($l_1,l_2,l_3$), and we already have the three points ($p_1,p_2,p_3$) on the box $B$, we calculate the optimal rigid transformation matrix (6 DoF) that transforms ($p_1,p_2,p_3$) to ($f_1,f_2,f_3$).\\
	For the same reasons discussed before, we need a global deterministic rigid registration method. We used \cite{88573} in the the first part of our final year project, it assumes correspondence are given and always finds the optimal rigid transformation matrix while avoiding any reflection.
	The returned transformation matrix has the following form:
	\begin{equation}
	\begin{pmatrix}
	r_{11}&r_{12}&r_{13}&t_x\\
	r_{21}&r_{22}&r_{23}&t_y\\
	r_{31}&r_{32}&r_{33}&t_z\\
	0&0&0&1\
	\end{pmatrix}
	\end{equation}
	\section{Problem Reduction}
	To solve the problem described in Section \ref{problem}, we have to reduce it the the problem in Section \ref{theory}. In consequence, solving the first problem is done by simply solving the second one. By performing the following reductions:
	\begin{itemize}
		\item $B$ is the labeling object.
		\item $O$ is the center of the camera that took a 2D image in which and object we want to label exists.
		\item ($p_1,p_2,p_3$) are the points selected by the user on $B$
		\item ($c_1,c_2,c_3$) are the points selected on the 2D image that together with $O$ will create the lines.
		\item ($f_1,f_2,f_3$) are the 3D points we wish to find, they represent where $B$ should be placed (in other words, where ($p_1,p_2,p_3$) should be transformed to)
	\end{itemize}
	We successfully reduce our original problem into one we can solve. In the next section we will present a detailed walk-through of how we implemented this solution.
	\section{Implementation in Unity}
In this section, we will talk about how this idea is translated into Unity along with the implementation of the user interface and all of the necessary functionalities. 

\subsection{CAD Objects as Labeling Elements}
\begin{figure}[!hbt]
	\centering
	\includegraphics[width=\linewidth]{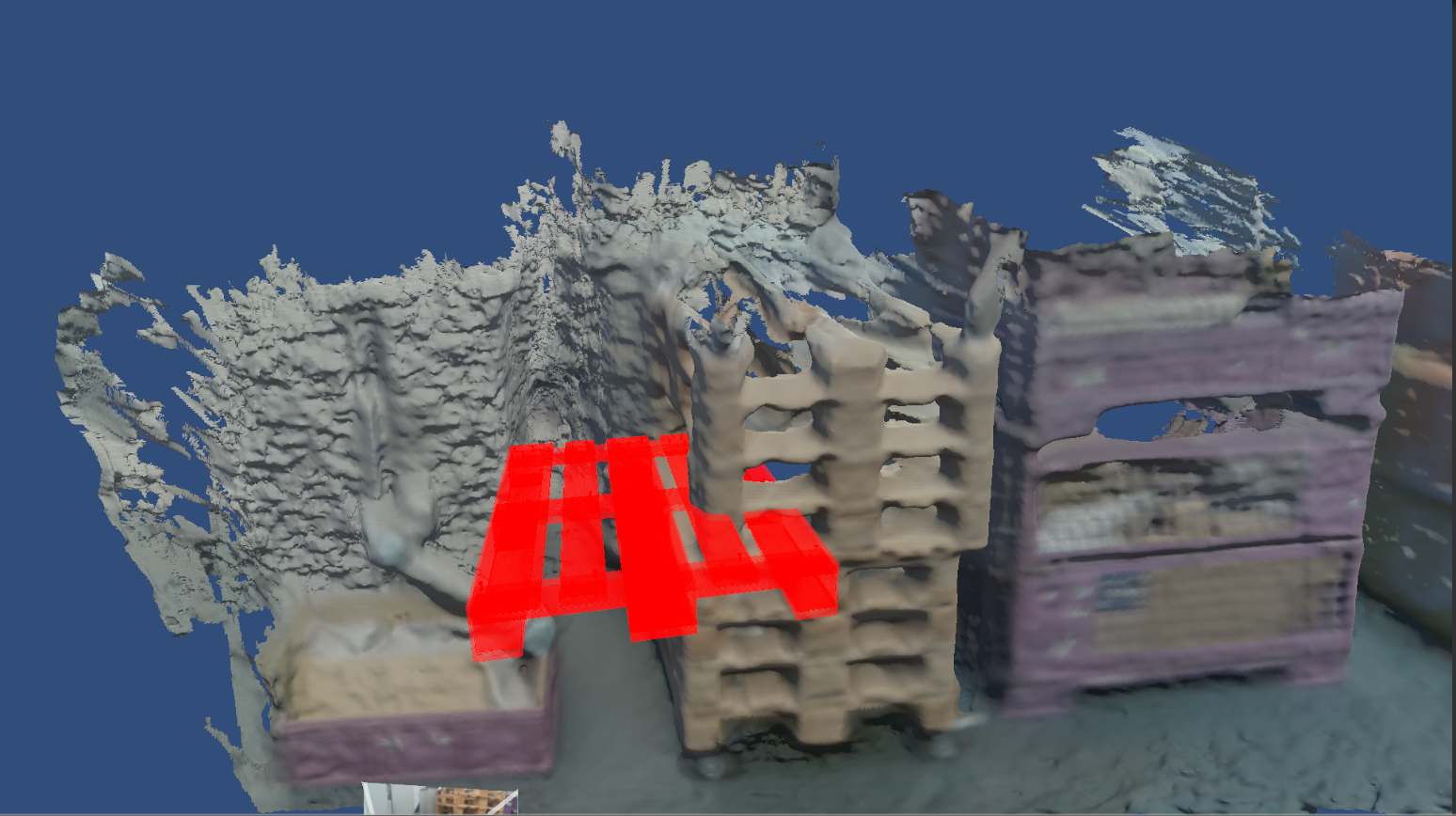}
	\caption{CAD Object Example in Scene}
	\label{fig:cadObjExample}
\end{figure}

To be able to apply this method in the tool, the user must acquire the CAD models of the objects that the user wants to label, knowing that this method is used as a rigid transformation and the source object must have the same size of the target object. For that matter a "Run Time Importer" was integrated to the tool from Unity's assets store. This package loads an ".obj" file and creates a game object in the scene that contains the mesh of this imported 3D model (see Figure~\ref{fig:cadObjExample}).

\subsubsection{Inventory Panel}
\begin{figure}[!hbt]
	\centering
	\includegraphics[width=\linewidth]{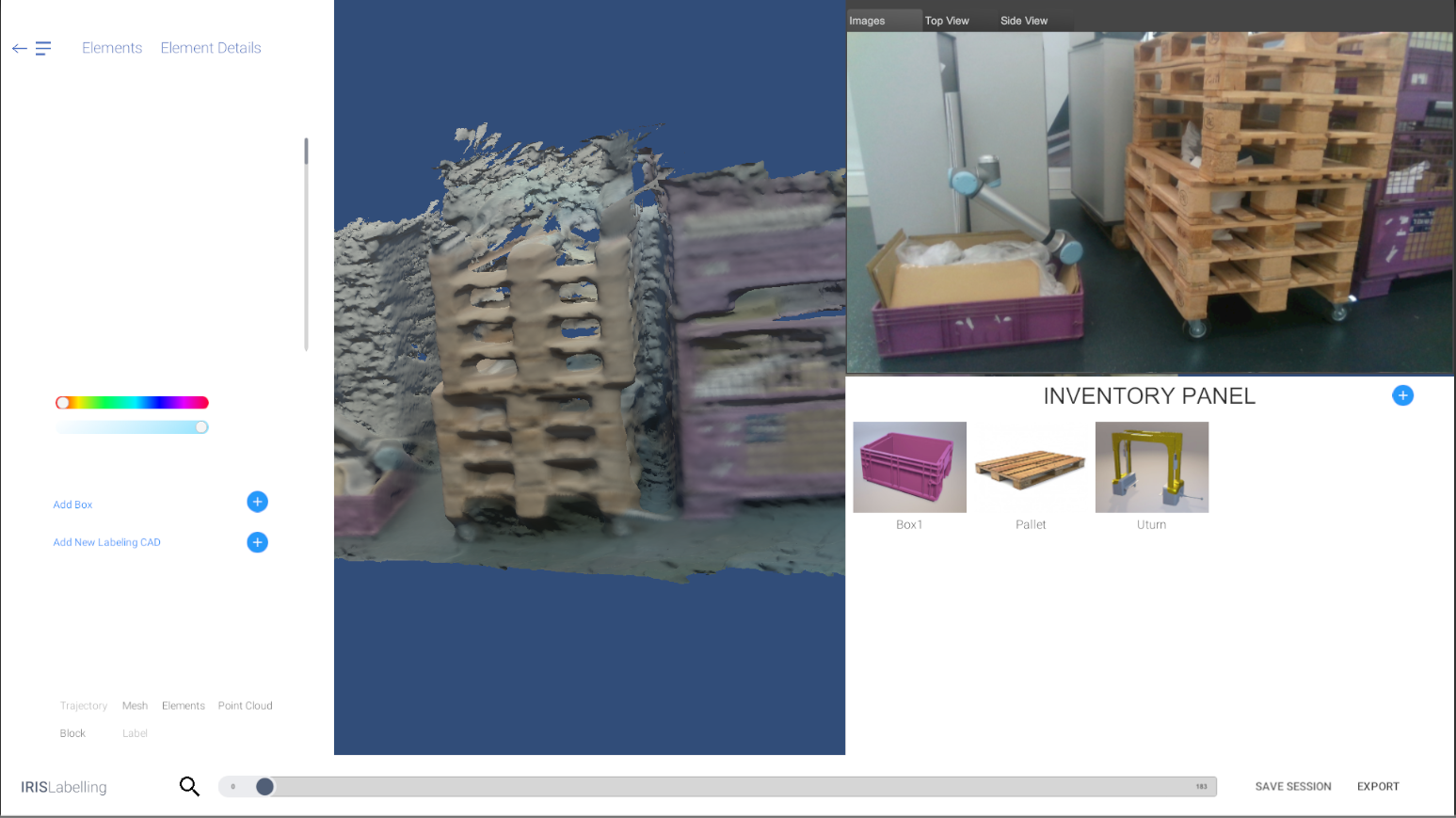}
	\caption{Inventory Panel}
	\label{fig:inventorypanel}
\end{figure}

To be able to get the most out of this asset to complement the user experience in the tool, an inventory panel was implemented to encapsulate all the CAD models that the user imports. In this case the user does not have to import the model each time he wants to add it to the scene, the user can check the inventory and load the CAD object from it. Nevertheless, we used Unity's "Streaming Assets" to store the objects in it after they gets imported. When a project is built in Unity, most of the assets are encapsulated into one package except the "Streaming Assets" folder which is a normal file system accessible via a path name: ``Application.dataPath + "/StreamingAssets"".\\
Furthermore, the inventory was created as a panel on the right hand side of the tool where the user can toggle it on or off. In this panel there's a "+" button to import the desired CAD objects into the inventory. Note that if there is an image of the CAD model in its path, it also get imported and shown in the inventory. Finally, the user clicks on the image to load the CAD object to the scene (see Figure~\ref{fig:inventorypanel}).

\subsubsection{CAD Objects in Phantom Mode} 
When the user enters ``Phantom Mode", all of the added labeling objects (CAD or primitive objects) are shown in a preview window in the bottom right corner, where the user clicks on the object to add the points that needs to be registered using the Python server. However, the preview window is a camera that renders only objects that the user wants to apply the transformations to. Moreover, because it is hard to know the exact size of the CAD objects, so there is a possibility that they wont fit into the preview window. We added a camera control script for the preview window camera, and as a result, the user could navigate in the preview window while adding the source points.\\
In contrast, when the CAD models are imported into the scene, we made sure that if they contained sub-meshes, they would be combined into one mesh, hence one object. This is done to make the transformation straightforward for one object, without thinking about how to transform the child objects in case the point is selected on that specific child object; in other terms, after the transformation is applied, the object will be decomposed.

\subsection{Point Selection}
\begin{figure}[!hbt]
	\centering
	\includegraphics[width=0.4\linewidth]{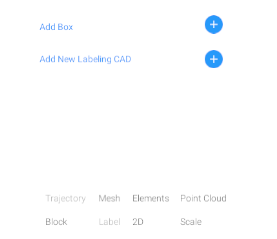}
	\caption{``2D" Toggle}
	\label{fig:2dtoggle}
\end{figure}

To activate this mode, the user must toggle on the ``2D" text on the down left panel (see Figure~\ref{fig:2dtoggle}). As a result, the user can now select points on the CAD object in the same fashion as the normal mode and the RGB images (top-right side of the main panel) are now interactive and the user can click on them to select the target points. In addition, if the mesh isn't loaded and the user clicks on ``Phantom Mode", the ``2D" mode is directly activated.\\
Note that in this mode, the points correspondence is mandatory, that is why the user should keep track of the color sequence on the source and target objects and ensure that the order and placement of the points is convenient.

\subsubsection{Source Point Selection}
\begin{figure}[!hbt]
	\centering
	\includegraphics[width=0.5\linewidth]{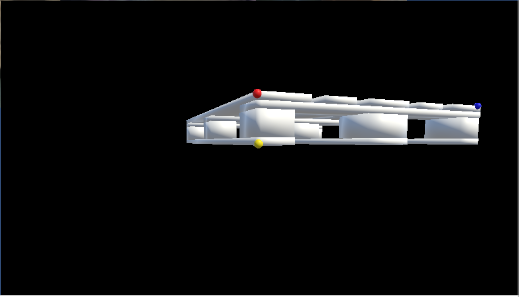}
	\caption{Source Point Selection Example}
	\label{fig:sourceobject2d}
\end{figure}

As already discussed in the ``Phantom Mode" chapter on colliders and specifically mesh colliders, in this mode we apply the mesh collider on the CAD model to be able to select the points (spheres) on it because of the collision between the two game objects. Nevertheless, to perform the problem reduction in our application, the object shown in Figure~\ref{fig:sourceobject2d} represents the labeling object $B$ and the points that are selected on it represent the set ($p_1,p_2,p_3$) that are finally added to an array to be sent to the Python server.

\subsubsection{Point Selection on the Image}
Furthermore, after selecting the points on the source labeling object, the user can now click on the image and add the three points on the target object with the same color correspondence of the source set of points.\\
First, the image that the user clicks on is the render texture of the shot view camera that we talked about in the bounding box chapter. In other terms, it is a scaled version of the camera's image frame. For that matter, in this application, the origin of the camera is the origin of the shot view camera.\\
Second, when the user clicks on the image, they are actually activating a ray to be generated from the shot view camera to the image frame on the same position of his cursor. This is done using the ``ScreenPointToRay" function in Unity that is responsible to return a ray object from the camera through a screen point given as an argument to the function. Nevertheless, the challenge was to calculate the cursor's position that should be given to this function. Knowing that the camera's raycast in Unity is equivalent to where the mouse position hits, helped us to form a logic of the workflow that should be done to produce the results that we needed. First of all, we calculated the $x$ and $y$ coordinates of the cursor's position relative to the texture of the raw image on which the user clicks on (top-right image of Figure~\ref{fig:inventorypanel}). To do that, we used Unity's ``ScreenPointToLocalPointInRectangle" function that transforms the cursor's position to the rectangle's (raw image) local space. Then, knowing that the resolution of the showing rectangle is different than the texture's resolution, we had to find the coordinates of the point in the texture's coordinate system by multiplying the cursor position by the ratio of the texture's and the rectangle's resolutions, then limiting the value between 0 and the texture's width or height. This can simply be translated into the following equation that calculates the $x$ coordinate of the point to be added on the image frame (same equation for the $Y$ axis using the height):
\begin{equation}
Imageframe.x = Localcursor.x \frac{Texture.width}{Rectangle.width}
\end{equation}

\subsection{Sending the Points}

\begin{figure}[!hbt]
	\centering
	\includegraphics[width=\linewidth]{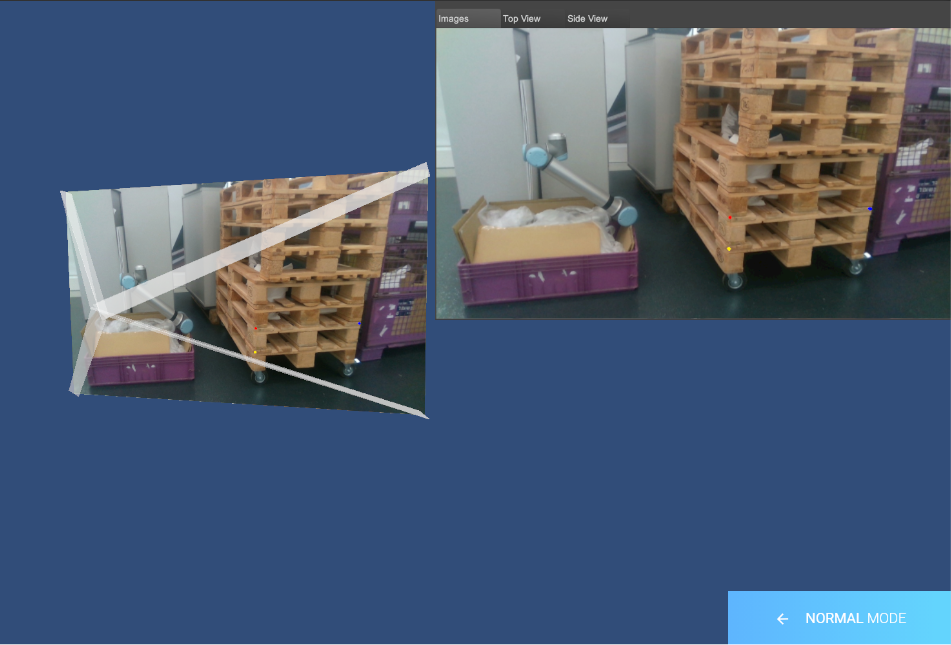}
	\caption{Selecting Points on Image Example}
	\label{fig:shotviewexample}
\end{figure}

As a result, on each ray hit, a sphere is created on the image frame (that contains a collider) and then rendered and shown directly onto the raw image on which the user clicked (see Figure~\ref{fig:shotviewexample}). These spheres corresponds to the set ($c_1,c_2,c_3$), they are added to an array along with the camera's position (origin).\\
Finally, the source and target points are sent to the server along with a key that says "2D" (to notify the server of the transformation to be done) through websockets. 

\section{Implementation in Python Server}
After receiving the two arrays of points from the Unity side, we unpack the target array and get the 4 points (origin and 3 points selected by the user on the image). We then basically follow the same steps described in Section \ref{theory} to build the system of non linear Equation \ref{eq:system}. Once that is done, we use the \textit{pyneqsys} \cite{Dahlgren} to apply the DOX\_BOX method with the only bound given on the depth value to be larger than the origin's depth. We then use the result of DOG\_BOX as an initial estimate to the Newton-Raphson method (also implemented in the \textit{pyneqsys} library). Then, we get the target points from the unknowns we calculated. Finally, call the Umeyama method\cite{88573} passing the source and the calculated target points. We will receive the optimal rigid transformation matrix that transforms the source points onto the target points. We send that matrix back to the Unity side through the same websockets connection.
\section{Results}
When the server finishes all the calculations, it sends back a transformation matrix describing how the labeling object should be translated and rotated. On the Unity side, this transformation is applied and the results are shown in the following Figures~\ref{fig:results1} and~\ref{fig:results2}. Note that, these results are produced from a data set where the mesh was not imported. Hence, this translates the use case of this functionality in the tool in case the data set does not include a mesh.\\
\noindent
In contrast, this method has a drawback concerning the user experience in terms of labeling, because the object can cover the whole image and that prevents the user to continue labeling other objects behind it, as showing in Figure~\ref{fig:results2}. As a solution, we recommend the user to change the transparency of the labeling object to be able to see behind it (see Figure~\ref{fig:results3}).

\begin{figure}[h]
	\centering
	\includegraphics[width=\linewidth]{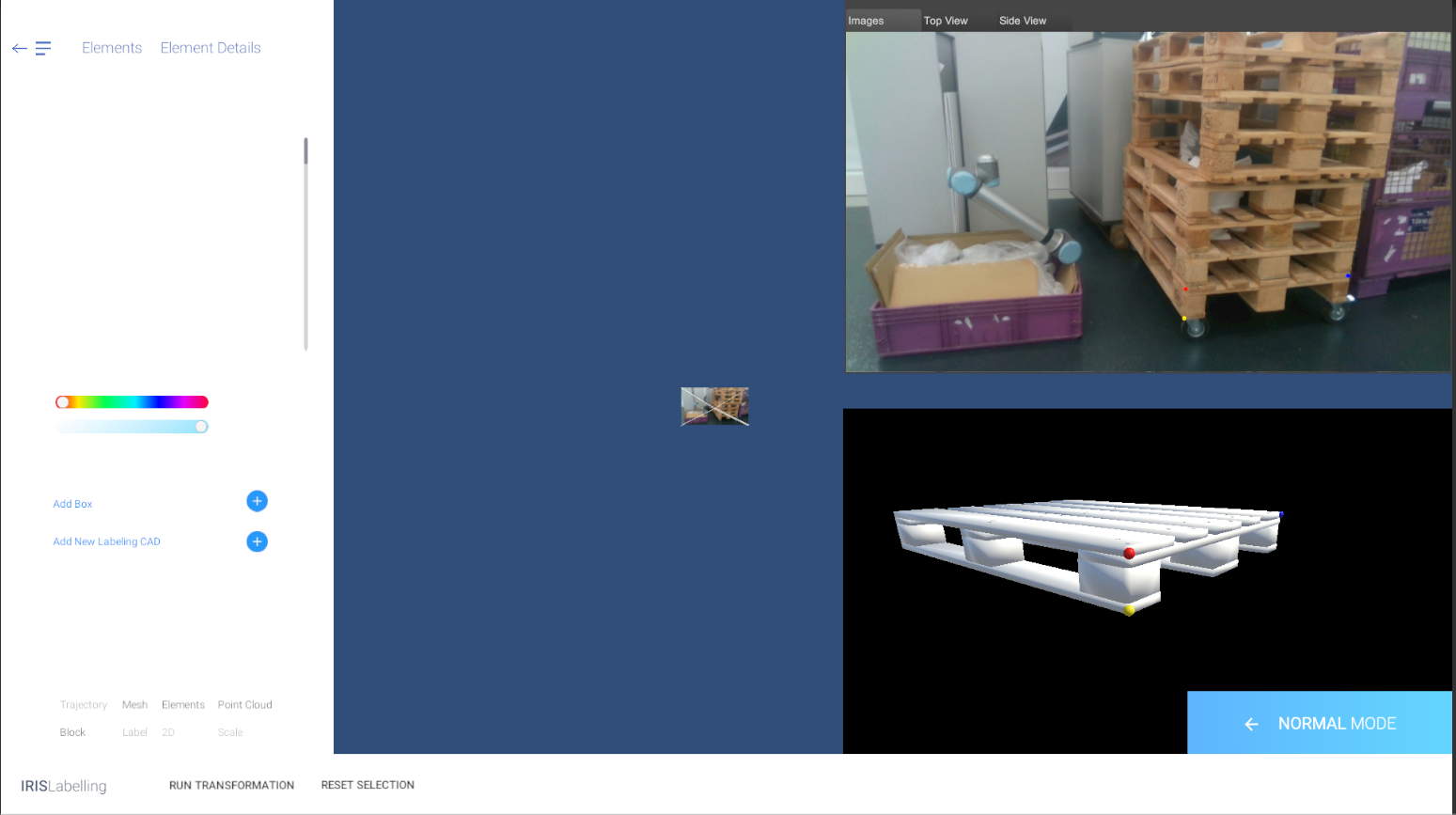}
	\caption{Results - 1}
	\label{fig:results1}
\end{figure}

\begin{figure}[h]
	\centering
	\includegraphics[width=\linewidth]{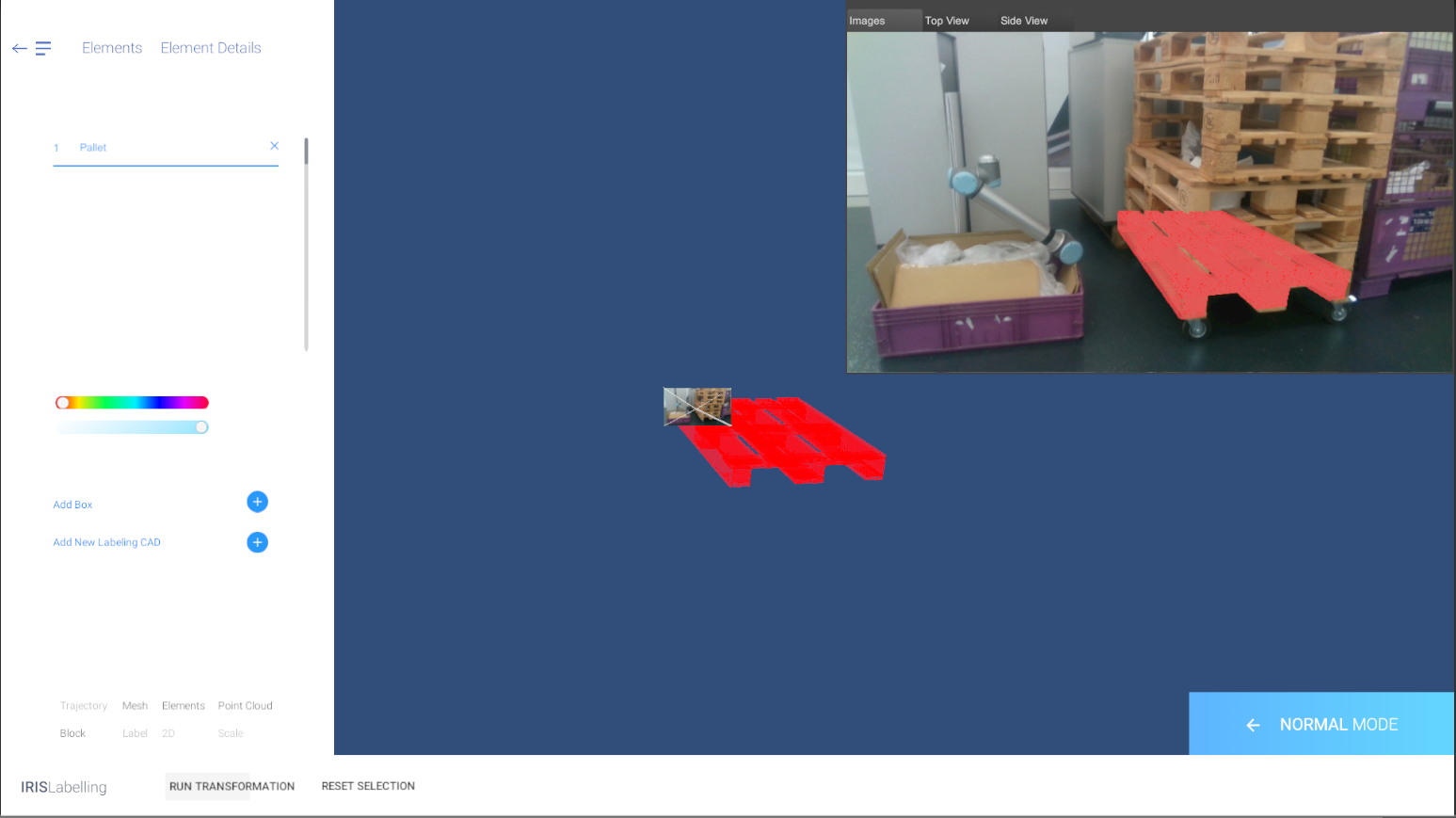}
	\caption{Results - 2}
	\label{fig:results2}
\end{figure}

\begin{figure}[h]
	\centering
	\includegraphics[width=\linewidth]{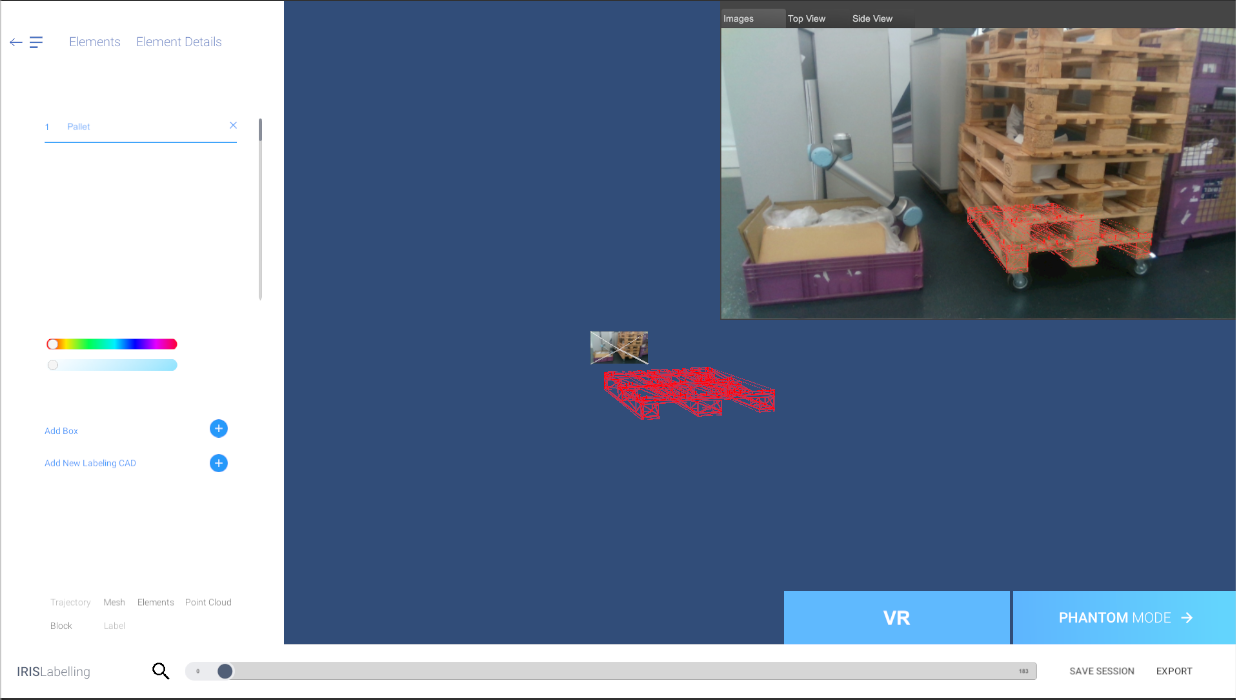}
	\caption{Results - 3}
	\label{fig:results3}
\end{figure}

	\chapter{Labeling Large Point Clouds}

	\section{Motivation}
	We have already described the usual way to create data for the tool. That is using a camera equipped with a depth sensor and then reconstructing the mesh from the RGBD images taken by it. However, at BMW Group, the majority of the plants possess laser scanned point clouds of their plants; see Figure \ref{Figure:brazilpc}. These scans are usually taken by a scanner named \textit{FARO}. They would like to label objects in these scans but labeling point clouds directly is still an unexplored field, moreover all of the object recognition models learn on labeled 2D images. Therefore, we need to create a workflow that can take these large pointclouds and transform them into data that can be labeled by our tool. In other words, (1) extract a triangular mesh (Section \ref{surfacereconstruction}) and (2) extract RGB images (taken at specific camera positions) of that mesh (Section \ref{datasampler}).
	\section{Surface Reconstruction}\label{surfacereconstruction}
	A blossoming subfield of computer graphics is perfect candidate to solve our first problem. In essence, it is the process of inferring (reconstructing) a 3D object from a set of unorganized discrete points that represent its shape. New advancements in data acquisition are the reason in the spark of interest in surface reconstruction. From optical laser-based range scanners, LiDAR scanners, structured light scanners, multi-view stereo to now inexpensive commodities like the Microsoft Kinect, point cloud extraction has become more accessible as time passes. Although the plethora of acquisition methods are great advancements in computer graphics, they also pose significant problems since each scanner produces point clouds with certain properties and certain imperfections. To cope with this variety, different classes of surface reconstruction algorithms were created each solving for a certain set of prior assumptions and robust against a certain type of imperfection. From methods that make little to no assumption on the quality of the point cloud and output non-mesh based reconstructions, to methods that assume a well sampled point cloud and produce a water tight triangular mesh.\\ In this section, we will present an overview on reconstruction methods comparing their different characteristics and explaining different prior assumptions they pose in the inputted point cloud. We will then delve deeper into the algorithms that we tried and compare their pros and cons. Finally, we will present a workflow that takes an unorganized point cloud and outputs a watertight mesh.
	\subsection{Point Cloud Imperfections}
\begin{figure}[h]
		\caption{2D representation of point cloud imperfections}
		\label{Figure:imperfections}
		\centering
		\includegraphics[width=0.75\textwidth]{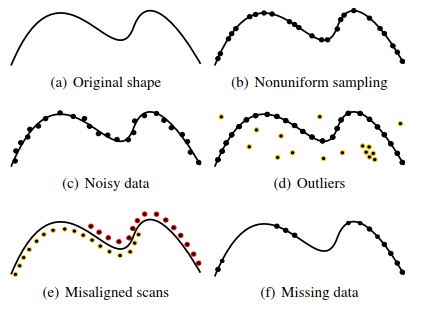}
	\end{figure}

	The five main imperfections that can plague a certain point clouds are: (1) noise, (2) sampling density, (3) misalignment, (4) outliers and (5) missing data. Refer to Figure \ref{Figure:imperfections} for a visual representation.
	\begin{figure}[h]
		\caption{Example of Point Cloud received from BMW plants}
		\label{Figure:brazilpc}
		\centering
		\includegraphics[width=0.75\textwidth]{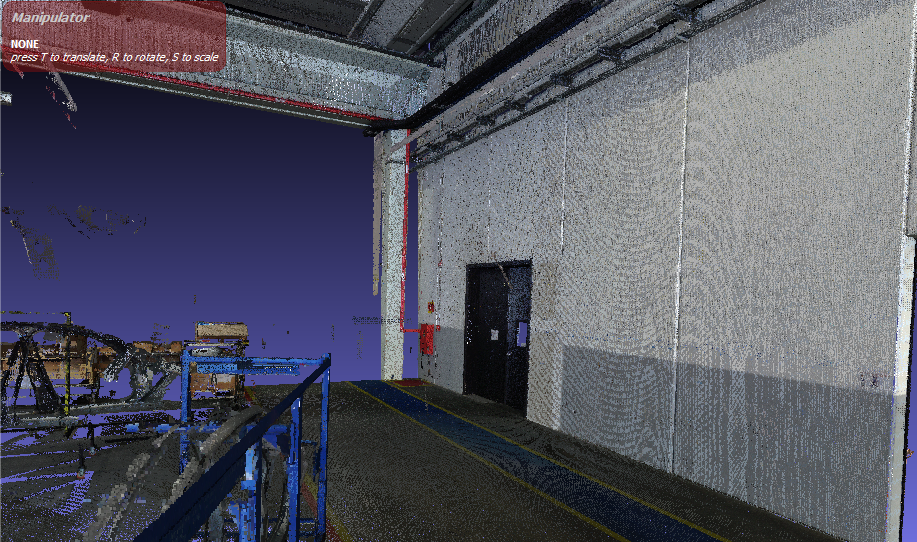}
	\end{figure}
	\subsubsection{Noise}
	Traditionally, points that are spread randomly close to the surface are regarded as noise. Noise can be introduced in different ways such as depth quantization, sensor noise and orientation or distance of the surface in relation to the scanner. The now conventional way of dealing with noise in surface reconstruction algorithms is to to produce a surface that passes near the points as much as possible but without over-fitting the noise. \cite{Kazhdan:2006:PSR:1281957.1281965} that imposes smoothness on the output and \cite{ztireli2009FeaturePP} which uses robust statistics are two algorithms very robust agains noise.
	\subsubsection{Sampling Density}\label{samplingdensity}
	Sampling density is one of the most import artifacts for point clouds. It describes the distribution of the points that sample the surface. It is critical because it defines a neighborhood which in turn defines the local geometry of the surface. The neighborhood should be small enough to capture small details in the mesh, yet large enough to describe the local geometry. The problem is that usually, scanners produce non-uniform sampling on the surface. A good practice we adopted is using poisson disk sampling \cite{Wei:2008:PPD:1360612.1360619} and keep point samples given a specific distance between them. Thus the point cloud becomes uniformly sampled.
	\subsubsection{Misalignment}
	The biggest cause of misalignment is the flawed registration of range scans. Contrarily to what one might think, handling misalignment is not the same as handling noise. For example, by assuming a prior of Manhattan world \cite{Vanegas:2012:AEM:2360744.2360844}, a scene consists of planar primitives aligned by three orthogonal axes, so planar primitives from mistakenly rotated scans can be robustly "snapped" onto one of these axes.
	\subsubsection{Outliers}
	Points that are very far from the surface are outliers. Unlike noise, outliers should not be taken into account when reconstructing the point cloud. In most cases, outliers can be handled implicitly using robust surface reconstruction methods \cite{Mullen}, but in some cases where the outliers are more structured (high density clusters of points exist far from the surface) like in multi-view stereo acquisition (where view-dependent specularities can results in false correspondences) it's better to handle them explicitly through outlier detection.
	\subsubsection{Missing Data}
	Missing data is a motivating factor behind many reconstruction techniques. Missing data is due to variables such as restricted sensor range, elevated light absorption, and occlusions in the scanning method that do not sample big parts of the form. Despite continuous improvement of the aforementioned artifacts, missing data tends to persist due to the device's physical limitations. We notice that missing data differs from non-uniform sampling, since the sampling density in these areas is zero. Most methods
	deal with missing data by assuming that the scanned point cloud is watertight. Other approaches seek to reconstruct higher-level information such as shape primitives, canonical regularities, skeleton and symmetry relationships for significant missing data.
	\subsection{Input}
	Different inputs are required by different reconstructions algorithms. The base input however is the position of each point in 3D space. That may suffice for some simple reconstructions but most of the time other inputs are added to make reconstruction simpler, especially for very challenging point clouds. We detail the different possible inputs in this section.
	\subsubsection{Surface Normals}\label{normals}
	Surface normals are defined at every point and represent the direction perpendicular to the points tangent space. In other words, its the direction perpendicular to the surface that surrounds a certain point. Normals can be oriented where a normal is consistently pointing inside or outside the surface, or unorientated in the opposite case.
	\paragraph{Unoriented Normals:} When a surface normal is either pointing on the outside or the inside then the normal is unoriented. According to our research they are used in one of three ways: (1) Constructing an unsigned distance field \cite{Amenta:2004:DPS:1186562.1015713}, (2) Projecting a point onto an
	approximation of the surface \cite{1175093} and (3) determining planar regions	in a point cloud \cite{Schnabel}.\\
	Unoriented Normals can be estimated from a raw point cloud. The best known way is using principal component analysis of the neighborhood of a point and use the smallest principal component from the calculated covariance matrix \cite{Hoppe:1992:SRU:142920.134011}. However, since it uses the neighborhoods of the points it can be sensitive to imperfections such as noise and sampling density. As a result, reconstruction methods used after estimation of these sometimes ``noisy" normals should be robust to them.
	\paragraph{Oriented Normals:}
	Oriented normals also either point to outside or the inside, but the difference is when an oriented normal is pointing on the inside we know the point is on the interior of the surface and when it is pointing to the outside then the point is on the exterior of the surface. At first, it was used to construct a signed distance field over the ambient space, where the field either takes a positve value indicating that the point is on the exterior and a negative values in the opposite case. We can then extract the surface by performing a zero crossing on the calculated signed distance field. More recently, this concept was generalized to implicit fields and indicator functions, but the basic idea still holds true \cite{Carr:2001:RRO:383259.383266,Kazhdan:2006:PSR:1281957.1281965,Ohtake:2003:MPU:1201775.882293}.\\
	Similarly to unoriented normals, oriented normals can also be estimated. As an example in \cite{Hoppe:1992:SRU:142920.134011}, a graph is constructed of the point cloud and weights each edge $w_{ij}$ for points $p_i$ and $p_j$ based on the similarity between the respective points' unoriented normals $n_i$ and $n_j$ as $w_{ij} = 1-|n_i ·n_j|$. A minimal spanning tree is then built, where upon fixing a normal orientation at a single point serving as the root, normal orientation is propagated over the tree. Comparatively to unoriented normals, these methods are sensitive to imperfections in the point clouds and thus can orient normals in the opposite direction.
	\subsubsection{Scanner information}\label{scannerinfo}
	The scanner that obtained the point cloud can provide helpful surface reconstruction data. Its 2D grid design makes it possible to estimate the sampling density that can be used to identify certain types of outliers–points whose lattice neighbors are at a much higher range than the sampling density are probable outliers. Caution must be taken to distinguish outliers from sharp features, however.\\
	Scanner data can also be used to define a point's confidence, which is helpful to manage noise. Some scanners (e.g. LiDAR) provide measurements of confidence in the form of reflectivity measured at each point. One can also derive trust through information on the line of sight. Line of sight is the collection of line segments between each point in the point cloud and the position of the scanner head from which the point was obtained. If the angle between the line of sight and the normal  is big in active scanning systems, i.e. laser-based scanners, this can lead to a noisy depth estimate, i.e. bad laser peak estimate \cite{Curless:1996:VMB:237170.237269}, which implies low confidence.\\
	Note that the line of sight also defines the regions that are lying outside of the shape. The combination of line of sight from multiple scans refines the boundary area in which the surface is located, this is called the visual hull.
	\subsubsection{RGB Images}
	Different methods of acquisition that complement the acquisition of depth can be of excellent help. RGB image acquisition is a very popular method that goes hand in hand with countless sensors, like the Microsoft Kinect. In the case of the Kinect, the RGB camera is co-located with the IR camera, so if we assume the two are calibrated, the identification of corresponding depth and RGB values at a pixel level is straightforward.\\
	RGB images are most helpful for reconstruction in complementing depth information that is either not as descriptive as their visual appearance, or simply not measured by the data. For example, if a color image and 3D scan are at a wide baseline with very different views, then segmented parts of the image can be used in the original scan to infer 3D geometry \cite{Nan20142DDLF}.
	\subsection{Priors}
	Priors' growth is mainly influenced by emerging technologies for data acquisition. Acquisition techniques set expectations for the class of objects that can be obtained and the type of artifacts connected with the information obtained, also informing the sort of output generated by reconstruction algorithms and the fidelity of reconstruction. In this section we will lay out an overview of available priors, discussing their produced outputs and expected inputs, additionally characterizing these scenarios by the typical shape classes and procurement techniques. We will end by analyzing and choosing the best prior for our case to narrow our search for the optimal reconstruction algorithm.
	\subsubsection{Surface Smoothness}
	There are three types of surface smoothness priors. First, \hspace{0pt}local smoothness seeks smoothness in close proximity to the data. The output of these methods are usually robust to noise and nonuniform sampling but they struggle to preserve fine detail in the reconstruction \cite{Hoppe:1992:SRU:142920.134011,1175093}. Second, global smoothness strives for large-scale smoothness, higher order smoothness or both. Large-scale herein relates to the spatial scale where smoothness is enforced not just near the input. It is common for these methods to focus on reconstructing individual objects, producing watertight surfaces \cite{Carr:2001:RRO:383259.383266,Kazhdan:2006:PSR:1281957.1281965}. High order smoothness relates to the variation of differential properties of the surface: area, tangent plane, curvature, etc. As a consequence, this confines the shape class to objects that can be captured as fully as possible from various perspectives. Desktop scans capable of scanning tiny objects (i.e. 1 inch) to medium-sized objects (i.e. several feet) are frequently used to create such point clouds. Ideal devices for such scenarios are laser-based optical triangulation scanners, time-of-flight (TOF) scanners, and IR-based structured lighting scanners.In addition, due to the close proximity of the sensor to an object and its high resolution capabilities, the reconstruction of very fine-grained detail is commonly emphasized.
	\subsubsection{Visibility}
	The previous visibility makes assumptions about the rebuilt scene's exterior space, and how this can provide clues to combat noise, non-uniform sampling, and missing information.
	Visibility of scanners is a powerful precedent as discussed in Section \ref{scannerinfo}, as it can provide for an acquisition-dependent noise model and be used to infer empty space areas \cite{Curless:1996:VMB:237170.237269}. This allows the filtering of powerful, organized noise for water-tight reconstruction of individual objects \cite{4408983}, a common feature of multi-view stereo inputs. More recently, some scanners expanded the visibility prior to scene reconstruction like for example The Microsoft Kinect and the Intel Real Sense(Already seen in Chapter 3).
	\subsubsection{Volume Smoothness}
	Similarly to surface smoothness, volume smoothness imposes smoothness on the shape's volume. This prior mainly solves for point clouds with significant missing data \cite{Livny:2010:ART:1882261.1866177}. In fact, they assume that the scanners capturing the point cloud have limited range and mobility and therefore cannot extract a well sampled point cloud. Other methods extract the skeleton structure of the shape from significant missing data.
	\subsubsection{Primitives}
	The geometric primitives prior assumes that a compact set of easy geometric forms, i.e. cylinders, boxes, spheres, planes, etc., can explain the scene's geometry. In instances where watertight reconstruction of individual objects is concerned, the detection of primitives \cite{Schnabel2009} may subsequently be used for primitive reconstruction extrapolation when confronted with big quantities of missing information \cite{Schnabel2009}. Some examples that benefit from this type of prior are CAD objects(usually modeled through simple geometry shapes) and indoor environments(can be summarized as a collection of boxes and planes)
	\subsubsection{GLobal Regularity}\label{globalregularity}
	Many shapes such as man-made shapes, CAD models and architectural shapes, contain a certain level of regularity in their higher level composition. The global regularity prior takes advantage of this fact combat missing data. Regularity in a shape can take many forms, such as a building made up of façade components, building interiors made up of periodic shape arrangements, or a mechanical component made up of recurrent orientation relationships between sub-parts \cite{Li:2011:FLD:2355573.2356274}.
	\subsubsection{Data Driven}
	The data driven prior takes advantage of the large-scale availability of 3D data acquired or modeled, mainly in scenarios where the input point cloud is highly incomplete. In these scenarios we are mainly concentrated on the reconstruction of individual objects, or a collection of objects, as 3D databases tend to be populated with well-defined classes of semantic objects. For example, we may be concerned with acquiring a watertight reconstruction of an individual object from just one or more depth scans. An object database may be used to best match the unfinished point cloud to a full model, or to compose components from various designs \cite{Shen:2012:SRP:2366145.2366199}. In other cases, the reconstruction of the scene and the use of the database to complement the objects detected in the scene \cite{Kim2013} may be implicated. Because of its generality, the data driven prior's reconstruction is only as good as the provided data.
	\subsubsection{User Driven}
	The user guided prior includes the user into the reconstruction phase of the surface, enabling them to provide intuitive and helpful signals for reconstruction. The specific form of interaction is mainly motivated by the type of shape being reconstructed and how it was obtained. Often the focus is on topology recovery owing to incomplete sampling from the sensor. Some methods therefore concentrate on the reconstruction of arbitrary forms, while others acknowledge interactions with the skeletal model of a shape, relying on volumetric smoothness before guiding reconstruction. Scanning in outdoor environments can cause large gaps in acquisition in the reconstruction of architectural buildings, similar to the case of façades mentioned in Section \ref{globalregularity}. User interaction can therefore assist discover global regularity, as well as how the identified regularity can be applied to the remainder of the point cloud \cite{Nan:2010:SIU:1778765.1778830}. If fine grained reconstruction control is required, the user may indicate primitive geometry to model the construction, guided by the interactions found in the input \cite{Arikan:2013:OOS:2421636.2421642}.
	\subsection{Analysis}
	\subsubsection{Imperfections}
	The point clouds we receive are captured using FARO laser scanner. After analyzing these point clouds we realized that Missing data, misalignment of scans and outliers were rare occurrences. On the other hand, noisy data and nonuniform sampling were very prominent. Therefore, we should first use poisson disk sampling to sample the points in a uniform manner as seen in Section \ref{samplingdensity}. Moreover we should choose a surface reconstruction algorithm that is very robust against noise.
	\subsubsection{Point Cloud inputs}
	As we have previously discussed, the base requirement of any surface reconstruction algorithm is the coordinates of the points. The more inputs we use the easier it is to deal with imperfections and the better accuracy we will achieve. In the scans we receive, we are provided with oriented normals which as we have discussed in Section \ref{normals}  provide extremely useful	cues for reconstruction algorithms – see \cite{Carr:2001:RRO:383259.383266,Kazhdan:2006:PSR:1281957.1281965}. Unfortunately, scanner information and RGB images are not provided.
	\subsubsection{Priors}
	To choose the best prior we should use we have to recapitulate the nature of our received point clouds. The point clouds we receive are very large scans of the indoors of a warehouse. It is full of pallets, boxes, robots and dolleys among other objects. Since we will use this mesh to extract rgb images that we label in the tool and train the machine learning model with, we require the reconstruction of this point cloud with the highest possible accuracy. The finest details should be preserved. Priors like visibility and primitives are not beneficial since they are mainly robust against reconstructing single objects like in the case of CAD Models. Data-driven priors rely on available databases of objects which we don't poses. User-driven priors requires input from the user that might be useful but for very large meshes it would be extremely time consuming and difficult for the user to provide good enough feedback that would justify the compromise of time. Global Regularity is more focused on reconstructing objects that contain pattern in them, however again, the size of the mesh and therefore its diversity will make it nearly impossible to extract good patterns, additionally it usually fails to preserve fine detail. Volume smoothness solves for the very specific problem of the abundance of missing data, a problem which we do not face thanks to the mobility of the FARO laser scanner. Finally, surface smoothness comes three-fold, local smoothness does not preserve sharp details so it's not the best candidate. Lastly, Global smoothness targets large scale reconstructions and are usually very robust against noise, they require point clouds without missing data which we possess and reconstruct with emphasis on preserving the smallest details. Forthwith, we will explore reconstruction algorithms based on the global smoothness prior and requiring only oriented normals as inputs.
	\subsection{Global Surface Smoothness Prior}
	\subsubsection{Radial Basis Functions}
	RBFs are a well-known interpolation technique for scattered information. Because of a collection of points with prescribed function values, RBFs reproduce features that maintain a large degree of smoothness through a linear combination of features that are radially symmetric. For surface reconstruction, the \cite{Carr:2001:RRO:383259.383266} method builds the surface by finding a signed scalar field defined by RBFs whose zero level set represents the surface. According to Table 1 from \cite{Berger:2017:SSR:3071788.3071810}, The aformentioned method is very weak when dealing with noise therefore we will not develop it further.
	\subsubsection{Indicator Functions}
\begin{figure}[h]
		\caption{Taking the gradient of the indicator function $\chi$
			reveals its connection to the point cloud normals. Poisson
			reconstruction \cite{Kazhdan:2006:PSR:1281957.1281965} optimizes for an indicator function
			whose gradient at $P$ is aligned to $N$.}
		\label{Figure:indicatorfunc}
		\centering
		\includegraphics[width=0.75\textwidth]{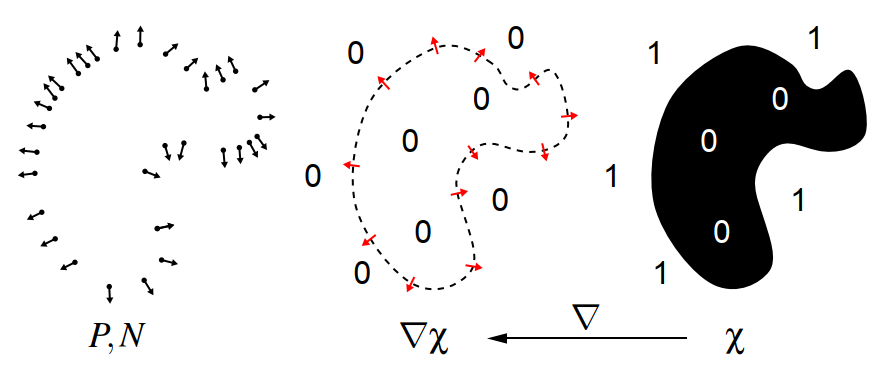}
	\end{figure}

	These methods approach the reconstruction of the surface by estimating a soft labeling that discriminates the inside of a solid shape from the outside. Indicator function methods are an instance of gradient-domain techniques \cite{Perez:2003:PIE:1201775.882269}. Such a gradient-domain formulation results in robustness to non-uniform sampling, noise, and outliers and missing information to some extent for surface reconstruction. This is achieved by finding an implicit function $\chi$ which best reflects the indicator function. In this class of techniques, the main observation is that, assuming a point cloud with oriented normals, $\chi$ can be calculated by making sure that the gradient of the indicator function measured at the point cloud $P$ is aligned with normals $N$; see Figure \ref{Figure:indicatorfunc}
	The indicator function thus minimizes the following quadratic energy:
	\begin{equation}
		\underset{\chi}{\operatorname{argmin}} \int\|\nabla \chi(\mathbf{x})-\mathcal{N}(\mathbf{x})\|_{2}^{2} d \mathbf{x}
	\end{equation}
	The differential equation that describes the solution to this
	problem is a Poisson problem; it can be derived by applying
	variational calculus as $\Updelta \chi =\nabla\mathcal{N}$. Once a solution of this equation is discovered, a suitable surface-specific iso-value is chosen as the average (or median) of the indicator function assessed at P. The implicit function's gradient being well-constrained at the data points enforces smoothness and a quality fit for the data, and since the gradient is assigned zero away from the point cloud, in such regions $\chi$ is smooth and well-behaved.\\
	\cite{Kazhdan:2005:RSM:1281920.1281931}'s strategy solves the Poisson problem by transforming it into the frequency domain where the Fourier transforms of $\Updelta \chi$ and $\nabla\mathcal{N}$ result in a straightforward algebraic shape to obtain the Fourier representation of $\chi$. However, by working in the frequency domain, a regular grid must be used to apply the FFT, thus restricting the output's spatial resolution. For the purpose of scaling to larger resolutions, the \cite{Kazhdan:2006:PSR:1281957.1281965} technique solves for $\chi$ directly in the spatial domain through a multi-grid strategy, hierarchically resolving $\chi$ in a coarse-to-fine resolution way. An extension of this method for streaming surface reconstruction, where the reconstruction is done on a subset of the data at a time, has also been proposed \cite{Manson:2008:SSR:1731309.1731324}. It represents the wavelet-based indicator function, and calculates the base coefficients efficiently using simple local sums over an adjusted octree.\\
	A known problem with \cite{Kazhdan:2006:PSR:1281957.1281965}'s strategy is that fitting directly to the $\chi$ gradient can result in data being over-smoothed[ \cite{Kazhdan:2013:SPS:2487228.2487237}, Fig. 4(a)]. To tackle this, the \cite{Kazhdan:2013:SPS:2487228.2487237} technique directly utilizes the point cloud's positional limitations in the optimization , leading to the screened Poisson problem:
	\begin{equation}
		\underset{\chi}{\operatorname{argmin}} \int\|\nabla \chi(\mathbf{x})-\mathcal{N}(\mathbf{x})\|_{2}^{2} d \mathbf{x}+\lambda \sum_{\mathbf{p}_{i} \in P} \chi^{2}\left(\mathbf{p}_{i}\right)
	\end{equation}
	Setting a big value for $\lambda$ ensures that the zero-crossing input of $\chi$ will be a closer fit to the input samples $P$. While this may reduce over-smoothing, it may also result, similarily to terpolatory methods, to over-fitting. The \cite{doi:10.1111/j.1467-8659.2011.02058.x} technique also incorporates limitations on position and gradient, but also involves a third term, an indicator function constraint on the Hessian:
	\begin{equation}
		\begin{aligned} \underset{\chi}{\operatorname{argmin}} & \sum_{\mathbf{p}_{i} \in P}\left\|\nabla \chi\left(\mathbf{p}_{i}\right)-\mathbf{n}_{i}\right\|_{2}^{2}+\lambda_{1} \sum_{\mathbf{p} \in P} \chi^{2}\left(\mathbf{p}_{i}\right) \\ &+\lambda_{2} \int\left\|H_{\chi}(\mathbf{x})\right\|_{F}^{2} d \mathbf{x} \end{aligned}
	\end{equation}
	Which may improve surface extrapolation in missing data regions[ KH13, Fig. 6(a)]. The primary distinction between these two methods is that \cite{Kazhdan:2013:SPS:2487228.2487237} solves the issue through a multi-grid finite-element formulation, whereas \cite{doi:10.1111/j.1467-8659.2011.02058.x} uses finite-differences, due to the complexity of discretizing the Hessian word; in specific, the Poisson formulation \cite{Kazhdan:2013:SPS:2487228.2487237} is up to two orders of magnitude quicker than \cite{doi:10.1111/j.1467-8659.2011.02058.x}, see[ \cite{Kazhdan:2013:SPS:2487228.2487237}, Table 1].
	\subsection{Comparison}
	In this section we compare the four most prominent global smoothness surface reconstruction algorithms together: (1) Screened Poisson Reconstruction, (2) Original Poisson Reconstruction, (3) Wavelet reconstruction and (4) Smooth Signed Distance reconstruction. For all of these methods we use the recommended parameter settings as defined by the authors. For both of the Poisson methods, we used the publicly available code made available by the authors on GitHub \cite{poissonrecon}. For the SSD reconstruction, we used an improvement that uses a hash octree data structure based on Morton codes which is observed to allow much more efficient access to leaf cells and their neighbors, also available on GitHub \cite{ssd}. Finally, for the wavelet reconstruction we modified the authors original implementation \cite{wavelet} to adapt to the new specifications of the OBJ file format.
	\subsubsection{Accuracy}
	To evaluate the accuracy of these methods, we used an assortment of real world data scanned from laser scanners. These include: (1) The Stanford Bunny (0.2M points), (2) the Lucy (1.0M points), (3) the David (11M points), (4) the Awakening (10M points) and (5) the Neptune (2.4M points). We randomly divided the points into two subsets of the same size for each dataset: input points for the reconstruction algorithms, and validation points for measuring point-to-reconstruction distances.
	\begin{figure}[h]
		\caption{Depth 10 reconstructions of Neptune with a close up on the beard using: original Poisson, Wavelet, SSD and Screened Poisson (left to right)}
		\label{Figure:neptune}
		\centering
		\includegraphics[width=1\textwidth]{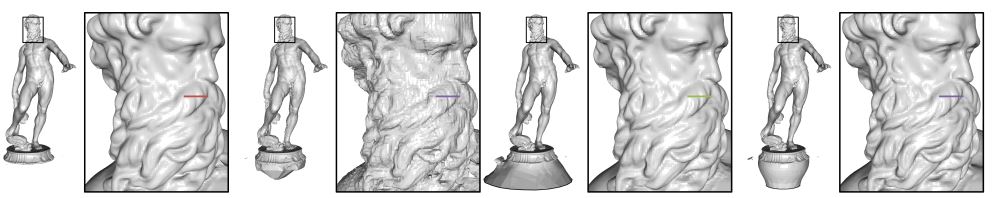}
	\end{figure}
	\begin{figure}[h]
		\caption{Depth 10 reconstructions of David with a close up on the eye using: original Poisson, Wavelet, SSD and Screened Poisson (left to right)}
		\label{Figure:david}
		\centering
		\includegraphics[width=1\textwidth]{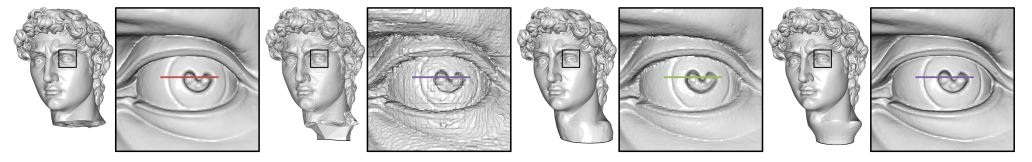}
	\end{figure}
	We visualize the reconstructions of Neptune and David in Figures \ref{Figure:neptune} \& \ref{Figure:david} respectively. We can observe that the Wavelet reconstruction has apparent derivative discontinuities. Additionally, the original Poisson and SSD reconstructions are to a certain extent over-smoothing the resulted mesh. Conversely, the screened Poisson method offers a reconstruction that fits the samples faithfully without causing any noise.\\
	Moreover, Figure 4b displays quantitative outcomes across all datasets, measured using distances from the validation points to the reconstructed surface in the form of RMS errors.
	\begin{figure}[h]
		\caption{For all models, plots of one-sided RMS errors, measured from the evaluation points to the reconstructed surface, as a function of the resolution depth (8,
			9, and 10)}
		\label{Figure:rms}
		\centering
		\includegraphics[width=1\textwidth]{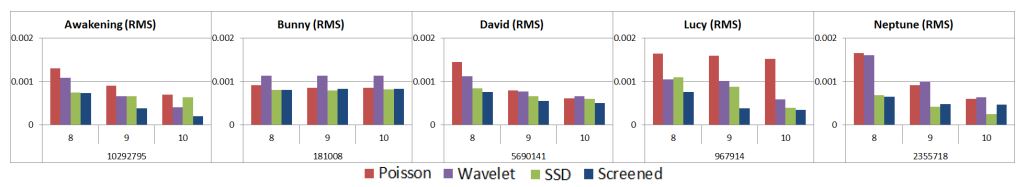}
	\end{figure}
	We can deduce that SSD and screened Poisson produce comparable results and greatly surpass the other two.
	\subsubsection{Computational Complexity}
	To evaluate the computational complexity, we compared the running time and memory usage of the four algorithms for depths 8, 9, 10 and 11 on the point clouds of Neptune and David. All of these experiments were run on the same machine provided in the Lab.	
	\begin{table}
		\centering
		\begin{tabular}{ |c|c|c|c|c|c| } 
			\hline
			Model & Depth & Poisson & Wavelet & SSD & Screened  \\
			\hline
			\multirow{2}{6em}{Neptune} & 8 & 10 & 3&275&14\\ 
			& 9 & 25&4&547&20\\ 
			& 10 & 89&6&3302&44\\ 
			& 11 & 320&9&15441&126\\ 
			\hline
			\multirow{2}{6em}{David} & 8 & 41 & 9 & 492 &48\\ 
			& 9 & 108 & 12 & 2355 & 73\\ 
			& 10 & 412 &20&19158& 182\\ 
			& 11 & 1710&43&119119&609\\ 
			\hline
		\end{tabular}
		\caption{Comparison of runtime in seconds between the four models for depths 8,9,10 and 11 on Neptune and David point clouds}
		\label{table:time}
	\end{table}
	\begin{table}
		\centering
		\begin{tabular}{ |c|c|c|c|c|c| } 
			\hline
			Model & Depth & Poisson & Wavelet & SSD & Screened  \\
			\hline
			\multirow{2}{6em}{Neptune} & 8 & 113 & 4 & 238 & 133\\ 
			& 9 & 149 & 11 & 455 & 269\\ 
			& 10 & 422 & 35 & 1247 & 604\\ 
			& 11 & 1387 & 118 & 3495 & 1622\\ 
			\hline
			\multirow{2}{6em}{David} & 8 & 427 & 11 & 863 &454\\ 
			& 9 & 510 & 38 & 1724 & 932\\ 
			& 10 & 1498 & 151 & 4895 & 2194\\ 
			& 11 & 5318 & 545 & \textgreater8192 & 6188\\ 
			\hline
		\end{tabular}
		\caption{Comparison of memory usage in MB between the four models for depths 8,9,10 and 11 on Neptune and David point clouds}
		\label{table:memory}
	\end{table}
	\paragraph{Time Complexity:}Table \ref{table:time} compares the four reconstruction algorithms in terms of time in seconds. We vary the octree depth from 8 to 11 and the results are shown for both David and Neptune models. Wavelet reconstruction is quick ; its use of compactly supported, orthogonal baseline functions allows the reconstruction algorithm to calculate the coefficients of the implicit function through integration; never needing an explicit linear system solution. On the other hand, the others use non-orthogonal basis functions, thus requiring a global system solution. For the Poisson algorithms, the multigrid solver performs a constant number of conjugate gradient iterations at each level, giving linear complexity in the number N of octree nodes. Increasing the depth by one therefore quadruples the time of computation. Whereas the SSD reconstruction utilizes conjugate gradients to solve simultaneously for all coefficients, which has an O(N1.5) complexity, resulting in significantly slower performance at greater resolutions.
	\paragraph{Space Complexity:}The memory usage of the various reconstruction algorithms further highlights the expense of formulating the reconstruction problem in terms of a linear system solution. Because the Poisson and SSD reconstructions use the two-ring neighbors to define a linear system, the scheme matrix can have as many as 125 entries per row, leading in a significant overhead to store the matrix. By contrast, a linear system does not need to be solved by the Wavelet reconstruction algorithm and therefore avoids the associated memory overhead.
	\subsubsection{Algorithm decision}
	It is evident to rule out SSD as a candidate since its time \& memory consuming and it's reconstruction results are similar to the screened Poisson reconstruction algorithm. Moreover, we can rule out the original poisson reconstruction since the screened version surpassed it in every aspect of our comparison. Finally, even though the wavelet reconstruction is fast and memory efficient, its reconstruction results are visually the worst when compared to the others. Therefore, we will solve this problem using the screened Poisson reconstruction algorithm.
	\subsection{Point Cloud Size}
	All point clouds we receive for reconstruction are usually huge in size. Containing around 300 Million points each. This led the screened Poisson reconstruction algorithm to fail due to memory overload. To deal with this, we modified the algorithms tree node indexes. In the authors' implementation, they use integers to keep track of the tree nodes' indexes. This limits the number of nodes to $2^{31} = 2,147,483,648$ possible nodes. After performing this improvement, the algorithm could finish successfully but the maximum possible octree depth we could reconstruct at was 13. According to the poisson paper, the number of nodes at the maximum depth d in an octree is $4^{d}$. Which means that for a depth of 13, the number of nodes is $4^{13} = 67,108,864$. It gave good results but if we compare that number to the number of input points (300 million), we can clearly see that there is room for improvement. We decided to cut up the point cloud into smaller pieces that we reconstruct independently and once we finish we stitch them all back together. We know that if we cut the mesh into 4 pieces then a reconstruction of depth 12 on each piece would yield the same overall resolution as reconstructing the whole point cloud using depth 13 ($4 * 4^{12} = 4^{13}$). Following the same logic, if we cut the point cloud into 8 pieces and reconstruct each one at depth 12 then it would be equaivalent to reconstructing the original at depth 14 which was impossible to use before. Of course, we could also reconstruct four pieces with depth 13 and that would lead to a depth 14 overall reconstruction. Thus, there are two key parameters to keep in mind: (1) the depth of the octree when upon reconstruction and (2) the number of pieces we cut the original point cloud to. We can take as an example one of the point clouds we received of the Brazil Plant which had $370,038,992$ points. We can deduce that this number of points fits between a depth 14 (268,435,456 nodes) and a depth 15 (1,073,741,824 nodes) octree. We need to choose the best compromise between time and accuracy, as achieving each incremental depth will quadruple the runtime complexity. If we choose to achieve a reconstruction of depth 14 then we can do it in multiple ways: (1) reconstructing 4 pieces at depth 13, (2) reconstructing 8 pieces at depth 12 or (3) reconstructing 12 pieces at depth 11 and so on.. In the next section we will provide our implemented workflow for reconstructing any point cloud and reaching any desired resolution no matter the machine's hardware capabilities, of course, at the expense of time.
	\subsection{Workflow}
	The point clouds we receive are multiple .ptx scans of the same plant, therefore we have to merge all of the these scans and reconstruct the merged point cloud. As we have already said, given a certain point cloud we will provide two ways of reconstructing it. 
	\subsubsection{Meshlab}
	Meshlab \cite{Cignoni2008} is an open source system for processing and editing 3D triangular meshes and point clouds.	It provides a set of tools for editing, merging, converting, cleaning, healing, inspecting, rendering, texturing and converting meshes. In this section, we will use its merging and conversion features. Meshlab also installs a local server which you can communicate with through the command prompt and execute Meshlab commands without opening the GUI version.
	\subsubsection{ChunkPLY}
	To perform the mesh cutting, we use an open source script written in C++ available on Github \cite{poissonrecon}. It is a simple script for breaking up geometry looking at which cell of a regular grid it falls into. For points, the cell is determined by the point's position.
	\subsubsection{Approach 1}
	\begin{enumerate}
		\item Merge all .ptx point clouds into one large .ptx point cloud using Meshlab.
		\item Convert the large .ptx point cloud into a .ply point cloud using Meshlab.
		\item Reconstruct the .ply point cloud using the modified screened Poisson reconstruction at an octree depth chosen by the user.
	\end{enumerate}
	As we have already mentioned, for the large point clouds we are reconstructing, the maximum octree depth our machine can handle is 13.
	\subsubsection{Approach 2}
	\begin{enumerate}
		\item Merge all .ptx point clouds into one large .ptx point cloud using Meshlab.
		\item Convert the large .ptx point cloud into a .ply point cloud using Meshlab.
		\item Cut .ply into chunks depending on the users input
		\item Reconstruct each chunk using the modified screened Poisson reconstruction at an octree depth chosen by the user.
		\item Merge the reconstructed chunks into one large triangular mesh
	\end{enumerate}
	In this approach, even though the individual octree depth limit is 13, when they are combined they can produce even higher resolutions.
	\subsection{Implementation}
	Each approach is implemented in a separate Python script. Because these approaches heavily rely on external application and scripts we decided to create a containerized environment that already has all these required applications installed.
	\subsubsection{Docker}
	\paragraph{Definition:}As per Docker’s own definition \cite{Merkel:2014:DLL:2600239.2600241}: Docker is “an open source project to pack, ship, and run any application as a lightweight container.” In layman terms a Docker “allows you to package your application along with all of its dependencies and configurations, making sure that the application can run on any infrastructure with almost no configuration changes on the customer premises”.
	\paragraph{Docker Container Vs Virtual-machine:}
	\begin{figure}[h]
		\caption{Difference between virtual machine and docker architectures}
		\label{Figure:docker}
		\centering
		\includegraphics[width=1\textwidth]{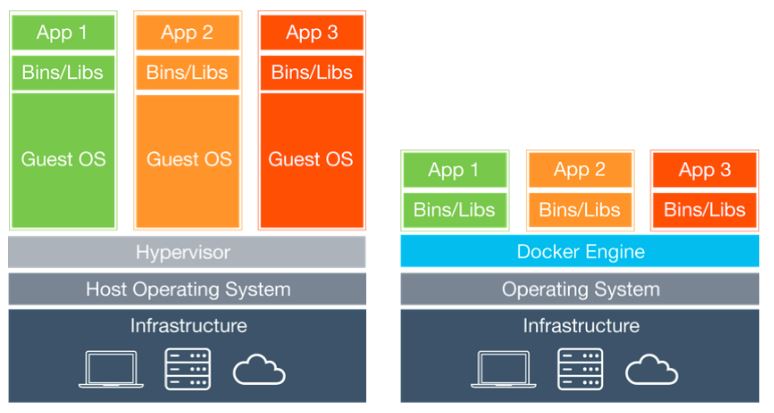}
	\end{figure}
	\begin{itemize}
		\item Footprint: VMs are inherently heavyweights. They need to run a complete OS to be able to run a packaged application. This is needed because the system calls made by the Apps are made to the underlying Guest OS. The Guest OS sends the system calls to the Host OS, via Hypervisor, and then relays the return value of the call back to the App. In the case of containers, the Docker engine does not need a Host OS. All System calls are intercepted by the Docker Engine and are relayed back to the Host OS. Hence, the Docker containers are extremely lightweight.
		\item Resources: VMs should be allocated a defined amount of resources, which cannot be shared between multiple VMs. If you share X amount of RAM to a particular VM, then this X amount of RAM would be dedicatedly allocated to the VM. In the case of containers though, they utilize the resources as per the need. If a container is running a very lightweight application, it will utilize just the right amount of RAM
		\item Automation: It is possible to create Docker containers on the fly by writing a couple of lines of configuration. 
		\item Instantiation: Instantiating a VM instance is a time taking process, sometimes taking tens of minutes. But Docker containers can be started within seconds.
		\item Collaboration: It is very easy to share your Docker images (and containers) with other users. A Docker provides you with Registries which can store and share your images, publicly or privately. VMs are not that easy to share.
	\end{itemize}
	\paragraph{Docker Architecture:}
	\begin{itemize}
		\item Docker Images: A Docker image is a read-only template with instructions for creating a Docker container. For example, an image might contain an Ubuntu operating system with an Apache web server and your web application installed. You can build or update images from scratch or download and use images created by others. An image may be based on or may extend, one or more of the other images. A Docker image is described in a text file known as a Docker file, which has a simple, well-defined syntax.
		\item Docker Container: A Docker container is a runnable instance of a Docker image. You can run, start, stop, move, or delete a container using the Docker API or CLI commands. When you run a container, you can provide configuration metadata such as networking information or environment variables. Each container is an isolated and secure application platform but can be given access to resources running on a different host or container, as well as persistent storages or databases.
	\end{itemize}
	\subsubsection{Creating Docker Image}
	First, we started with a pre-built docker image that runs Ubuntu and has Meshlab and its server installed. For ChunkPLY we clone the Github Repository where its code exists and compile it using the g++ compiler of Linux. For the modified screened Poisson reconstruction, we forked the authors' original repository, made the modifications and use our own forked repository in the docker. We also compile its code using g++. Finally, we install Python 3.7 and all of the required libraries to run our script.
	\subsubsection{Python Script}
	To run the approaches, we create two separate scripts. We will first talk about what is common between the two scripts. Both scripts take multiple arguments:
	\begin{itemize}
		\item data path: the path where the multiple ptx scans lie.
		\item output path: the path where the reconstructed mesh will be created.
		\item octree depth: the upper bound octree depth used for reconstruction (max is 13)
	\end{itemize}
	The difference is that the second script also takes the number of chunks required by the user. The scripts execute both approaches in the order described in the previous section. To execute Meshlab, ChunkPLY and the screened Poisson reconstruction we call Linux terminal commands from the python script using the library \textit{subprocess}.\\
	Until now the way we would run reconstructions is we launch a container from our image that has access to the folder from the Host OS that contains the ptx point clouds. We then run the python script we want providing our desired arguments and destroy the container upon reconstruction completion. In the next section, we will automate the step of creating and destroying the container using yet another Python script. 
	\subsubsection{Automation}
\begin{figure}[h]
		\caption{Top Level View of Surface Reconstruction Workflow}
		\label{Figure:surfacerecon}
		\centering
		\includegraphics[width=1\textwidth]{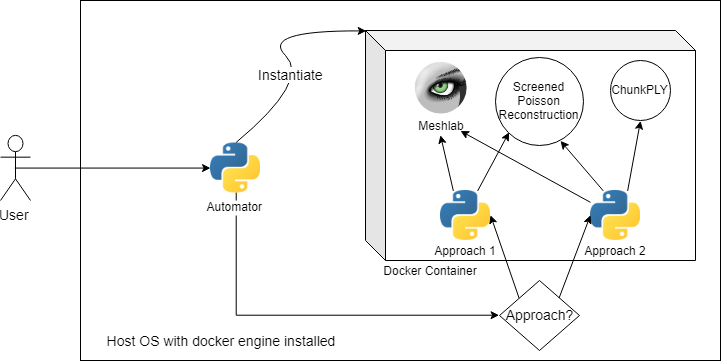}
	\end{figure}

\begin{figure}[h]
		\caption{Example of a Reconstructed Mesh of the Point Cloud in Figure \ref{Figure:brazilpc}}
		\label{Figure:brazilmesh}
		\centering
		\includegraphics[width=1\textwidth]{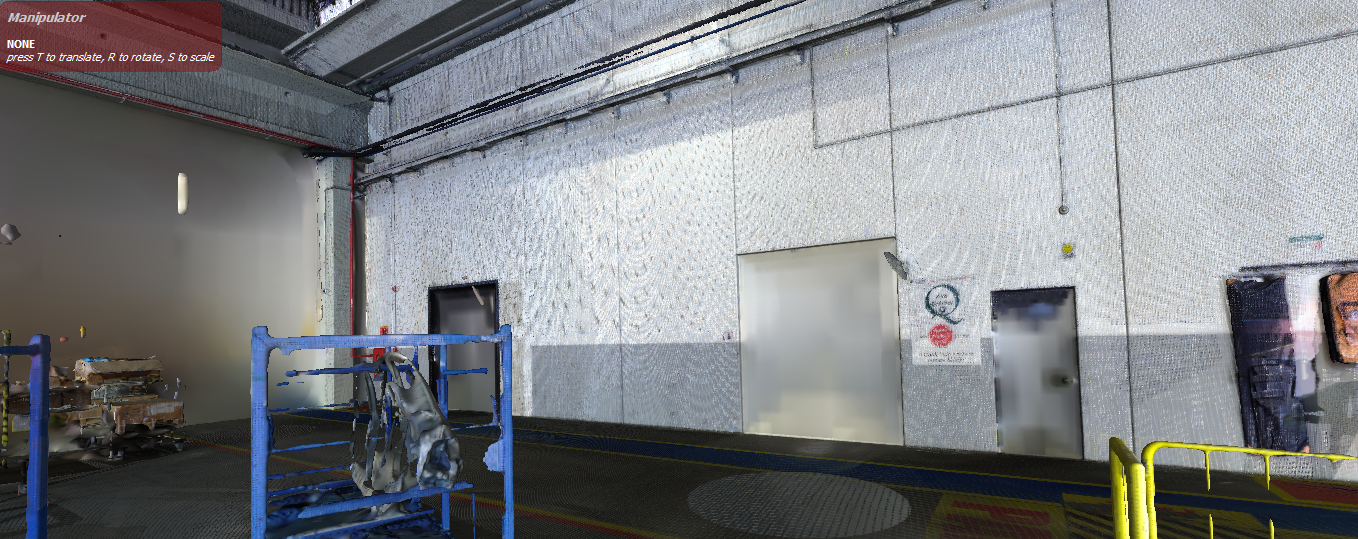}
	\end{figure}

	Our final level of abstraction is creating a Python script that starts the docker container that has access to the ptx point clouds and the two scripts, runs the desired script (desired approach) and kills the container upon completion.\\
	This is easily achievable by using the Python library \textit{docker}. It allows to automate any docker commands in Python. Naturally, the user has to give the arguments they wish to execute in the docker scripts to this script. This script will then pass these arguments to the scripts inside the docker. Please see Figure \ref{Figure:surfacerecon} for a visualization of the full process and Figure \ref{Figure:brazilmesh} for a reconstruction results of the example layed out in Figure \ref{Figure:brazilpc}.

	\section{Data Sampler}\label{datasampler}
	When the surface reconstruction is done and we have a fully constructed mesh, we can feed it to the data sampler tool that we have developed for the purpose of generating the IRIS format data set that will finally be imported into the 3D Labeling tool.\\
	The data sampler tool is built also in Unity3D. The idea behind it is to import a mesh and then the user can start recording what the camera is seeing and generating rgb and depth images while going through the mesh. In other words, it is a mimic of how the data for the 3D labeling tool is usually collected in real environments.
	
	\subsection{Recording and Generating the RGBD Data}
\begin{figure}[!hbt]
		\centering
		\includegraphics[width=\linewidth]{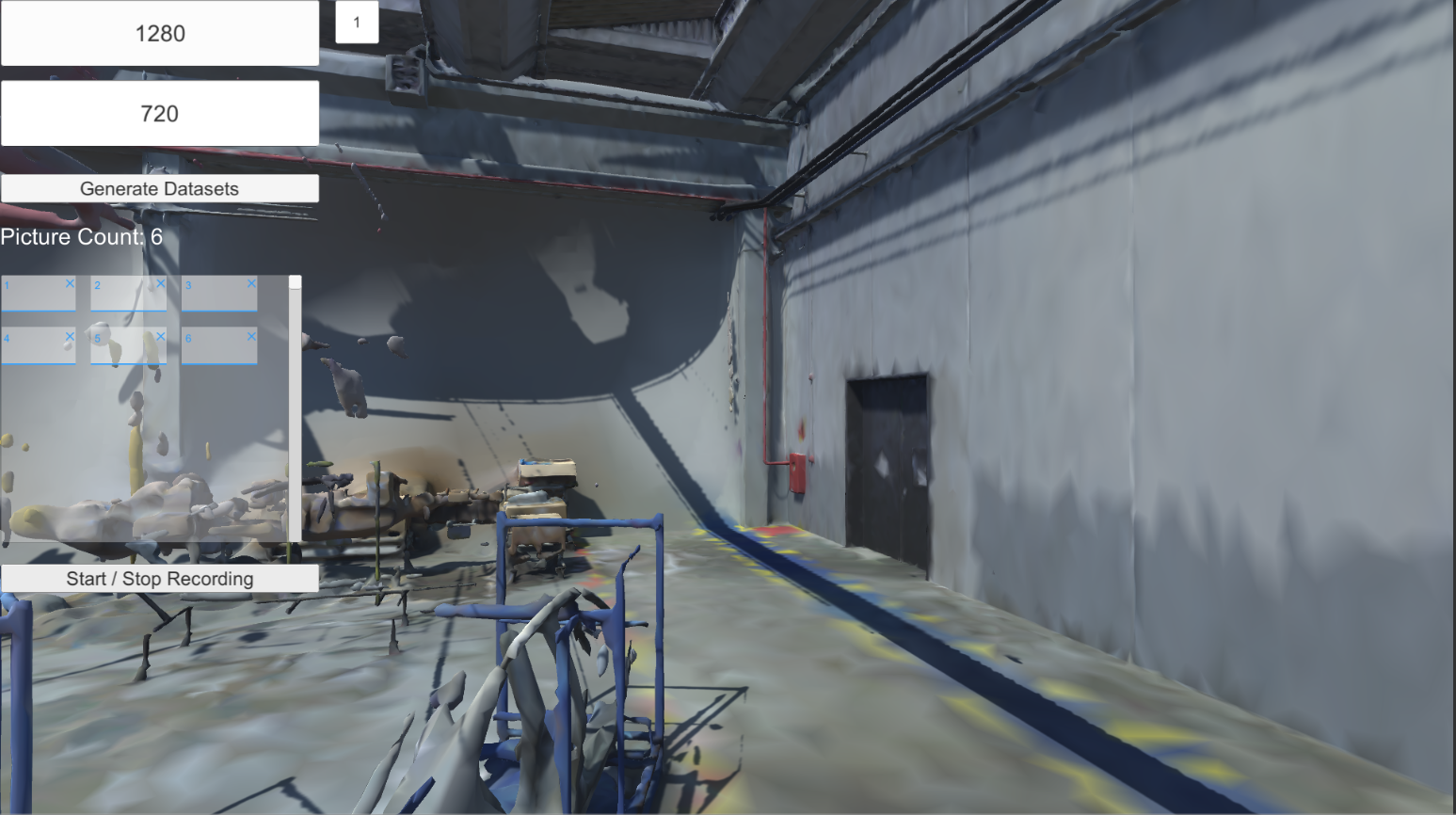}
		\caption{Data Sampler UI Example}
		\label{fig:datasamplerUI}
	\end{figure}

	When the mesh is imported to the data sampler tool, the user can navigate around it using the arrow keys and the mouse that basically changes the positions and rotations of the main camera. Before doing so, to start recording and capturing the RGB images from the mesh, they click on the start/stop recording button (see Figure~\ref{fig:datasamplerUI}). When the recording is on, the camera's transform (position and rotation) is saved on each frame (fps can be set by the user) in a list.\\
	When the user clicks on the generate datasets button, a screenshot is taken on each camera position that has been recorded. However, to capture what the camera is seeing we had to add a target render texture on the camera with a resolution that is set by the user (The top left values in Figure~\ref{fig:datasamplerUI}). Then, using Unity's \textit{ReadPixels} function, everything that is read by the main screen is projected into the texture that is finally an RGB image stored on the disk to form the first part of the IRIS format.\\
	In addition, for the depth images, we used Unity's depth shader script and applied a post processing effect on the rendered texture to obtain a depth map which is a depth image stored in IRIS format. Note that taking the depth image is done with the same method as of the RGB images but using another camera that is created as a child of the main camera to preserve the main camera's position and rotation for the purpose of having for the same RGB image the corresponding depth image.
	
	\subsection{Generating the Camera Information}
	To complete the IRIS format creation, we still need the camera's extrinsic list that defines all the positions of the camera in the world and the direction that is pointing at. Along with the camera's extrinsic we also need the intrinsic that represent the camera's calibration and how the 3D camera coordinates are transformed to a 2D image coordinates.\\
	First, the camera extrinsic are the positions stored when the user is recording. Hence, using these positions and transforming them into world space with Unity's \textit{transform.localToWorldMatrix} function, we obtain a 4x4 transformation matrix that is added to a list which is finally JSON serialized and stored in the ``extrinsic.json" file of the IRIS format.\\
	Second, to generate the camera's calibration we had to calculate the Unity camera's focal length. Knowing the camera's field of view, we used the following equation to extract the focal length:
	
	\begin{equation}
			focal length = \dfrac{resolution.height}{2*\tan(fieldofview*(\pi/360))}
	\end{equation}
	
	\noindent
	In addition to the focal length, we added to the intrinsic matrix the resolution's width and height divided by 2. Finally, we got the following intrinsic matrix serialized into JSON and written on a file called ``intrinsic.json" in the IRIS format directory:
	\begin{equation*}
	\centering
	\quad
	\begin{bmatrix}
	focallength & 0 & width / 2 \\
	0 & focallength & height / 2 \\
	0 & 0 & 1 \\
	\end{bmatrix}
	\quad
	\end{equation*}

	\begin{figure}[!hbt]
		\centering
		\includegraphics[width=\linewidth]{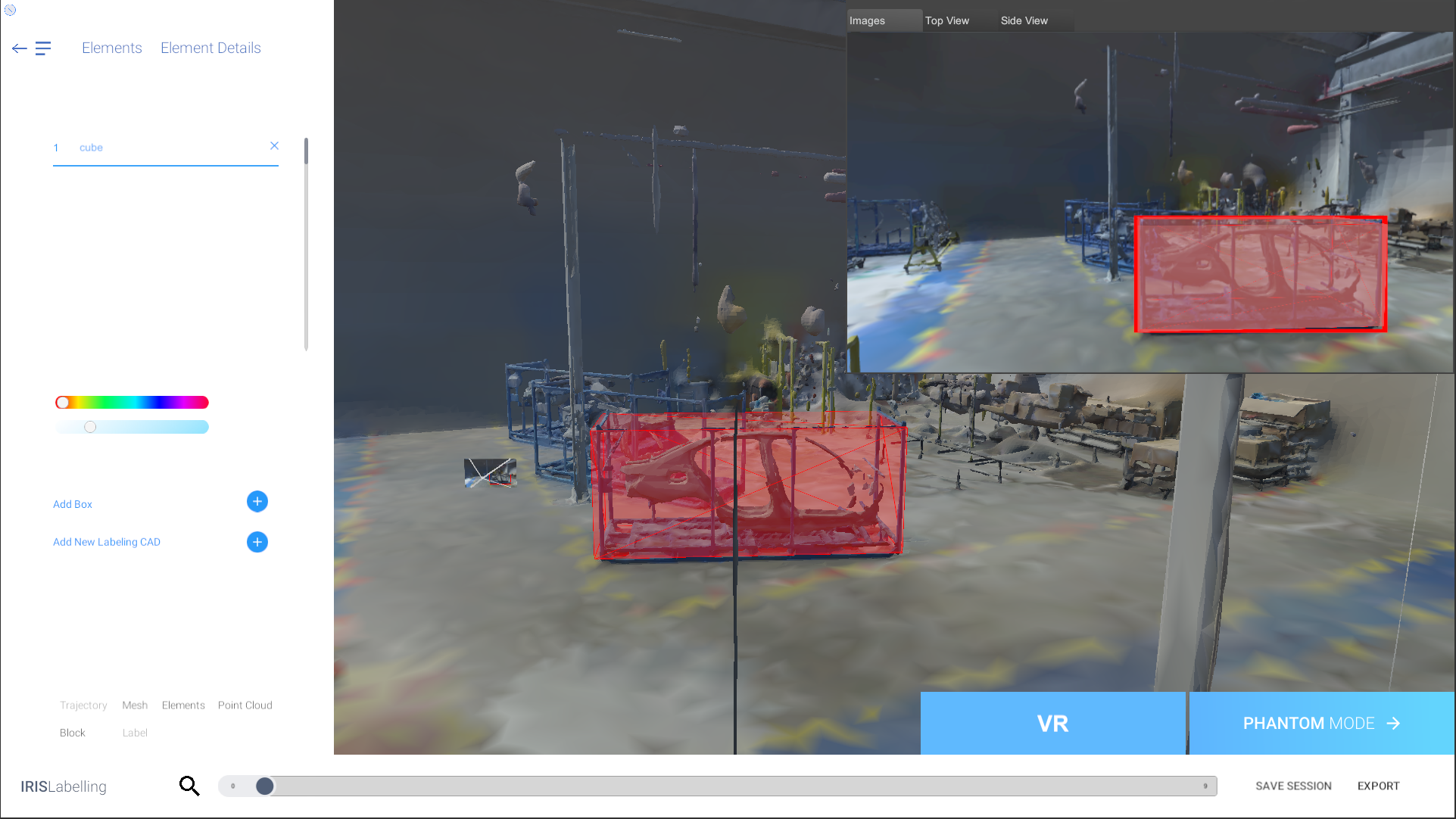}
		\caption{Sampler Generated Data Imported in 3DLT}
		\label{fig:datasampler}
	\end{figure}
	
	\noindent
	In summary, the generated data set from this tool is stored as an IRIS format which later on can be imported to the 3D labeling tool (see Figure~\ref{fig:datasampler}). That, gives the user the ability to label any data set, whether it is only coming from a large point cloud or a standard IRIS format data set.

	\chapter{Snapping}
	
	\section{Introduction}
	In the previous chapters we showed the various features that the user can benefit from to ensure an intuitive labeling experience. However, in some cases, the placement of the labeling elements on the mesh in the right position and rotation can be time consuming.\\
	As a result, we introduced a new feature that snaps the labeling element in its convenient place if it is more or less close to it and the labeling object has the same size as the object to be labeled.\\
	This functionality is based on the concept of registering the labeling element onto its correct place in the mesh. Before applying the registration algorithm, two depth images should be sent to the server along with the tool's main camera intrinsic and extrinsic. We will begin by examining how the depth images are captured and what do they represent. Then, we will present the server side of this feature and finally we will highlight the results of this enhancement.
	
	\section{Data Pre-Processing}
	The snapping is launched when the user selects a certain labeling element in the scene and clicks on the button "S" on the keyboard. Before the snapping is applied, the data is prepared in the Unity side by forming a message that contains two encoded depth images and the main camera's intrinsic and extrinsic. This message is sent to the server through websockets to ensure the corresponding computations and to produce a rigid transformation that is applied on the labeling element.
	
	\subsection{Depth Camera Setup}
	First, a camera is added as a child of the main camera's game object to ensure that both have the same transformation. This camera is going to play the role of a depth camera that takes a screenshot of what it is rendering the moment the snapping is applied (Same concept as mentioned in Chapter 8). Moreover, the depth property is given to this camera by enabling the ``DepthTextureMode.Depth" that generates a depth texture. To calculate the depth map in values and encoding it into an RGBA format, we used the ``Shader" language in Unity and applied it as a post rendering effect on the camera. This is done using Unity's \textit{OnRenderImage} and \textit{Graphics.Blit} that is given the source and destination textures along with the material that will affect the destination texture. It is the texture to encode and send to the server.\\
	Furthermore, in the shader calculations, the depth is received from the Depth Texture of the camera that is already enabled. However, the depth value returned from this texture is not linear; therefore, we used the \textit{Linear01Depth} function to set the depth value between $0$ and $1$, in the case where $1$ is the maximum value corresponding to the far clipping plane that is set to $65$ meters in our case. Note that the far clipping plane is the farthest point that the camera can render.\\
	Finally, the linear depth value is encoded on $4$ channels R, G, B and A on a scale of $[0, 1]$ in the ``Shader" script. Knowing that there's $256$ color in each channel, we multiply the depth value by $256^{i}$ (i ranging between $1$ and $4$, representing the values of R, G, B and A) while limiting its value to 256 using \textit{modulo}. In addition, to enclose the values between the range $[0, 1]$ we divide each one by $255$. As a result, we obtain a vector of $4$ values (RGBA) representing the encoded depth value.
	
	\subsection{Capturing the Depth Images}
\begin{figure}[!hbt]
		\centering
		\includegraphics[width=\linewidth]{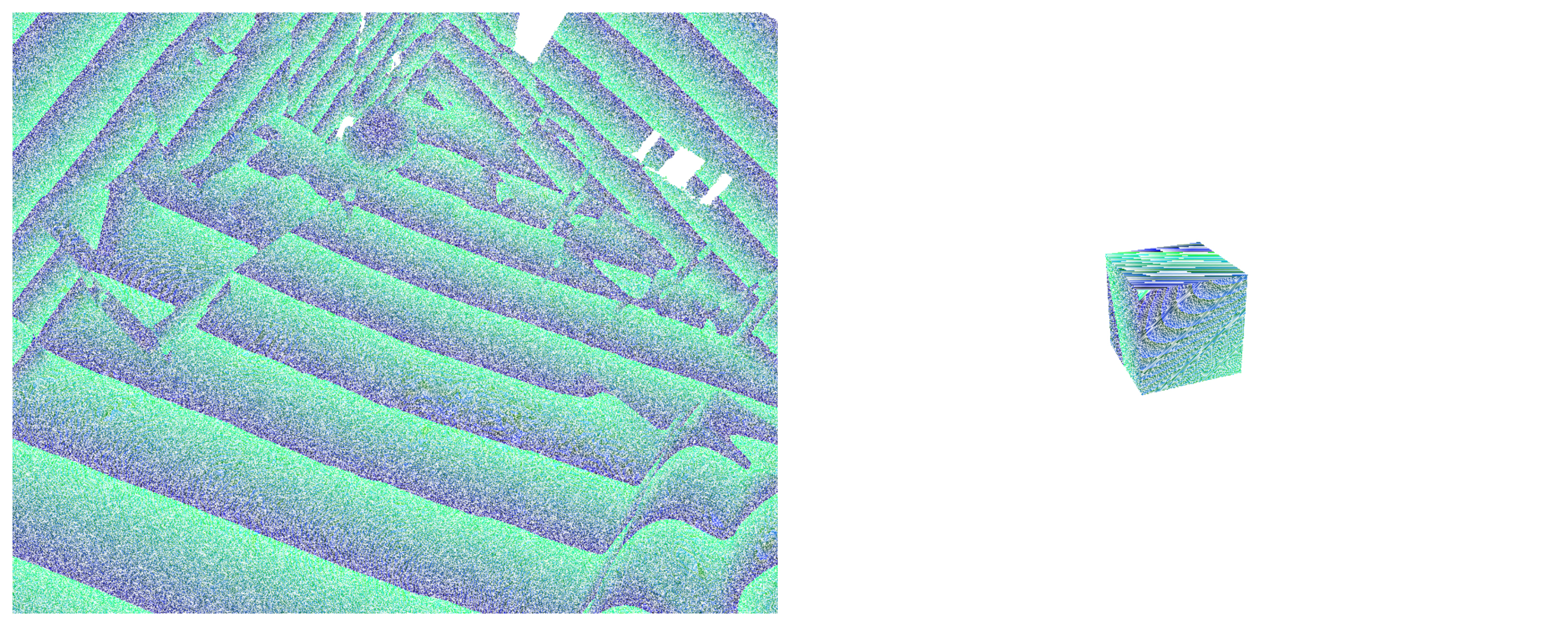}
		\caption{Depth Images Example}
		\label{fig:depthimages}
	\end{figure}

	To apply the ``Shader" script mentioned above, we have to attach it to a material on which the camera should render the texture to ensure the post processing effect and generate the depth map. To differ between the labeling element and the mesh, we assigned a culling layer for each one, then we added a culling mask to the depth camera for each layer. As a result, the camera captures one image that contains the labeling element and then changes its culling mask and captures another image containing only the mesh.\\
	The images in Unity are 2D textures where we specified the resolution to $256\times144$ for the purpose of decreasing the number of pixels and by that optimizing the time complexity of the registration algorithm. Finally, to capture the images, we used the \textit{ReadPixels} function in Unity that reads the pixels from the screen into the saved texture with the corresponding resolution (see Figure~\ref{fig:depthimages}). Then, we encode the following texture into an array of bytes using Unity's \textit{EncodeToPNG} function. Note that, this is done twice, once for the image that contains only the labeling element and the other for the image that contains the mesh; however, when we encode to PNG, the RGBA values are then transformed from the range $[0, 1]$ to a range $[0, 255]$.
	
	\subsection{Generating the Camera Intrinsic and Extrinsic}
	As part of the data preparation, we generated the depth camera's intrinsic and extrinsic in the same way we did in Chapter 8 in which we formed the camera's extrinsic using Unity's \textit{transform.localToWorldMatrix} property that returns a transformation matrix in the world space of the game object. On the other hand, the intrinsic matrix was constructed after calculating the camera's focal length and the texture's resolution.
	\subsection{Sending the Data}
	In order to send the data to the Python server, we created a class that contains two byte arrays one for each image and two matrices for the camera's intrinsic and extrinsic. This class is serialized into a JSON message and sent to the server through websockets.
	
	\section{Server Side}
	The Python server receives a total of four inputs: (1) Encoded depth image containing the labeling object, (2) Encoded depth image containing the mesh, (3) Intrinsics of the camera and (4) Extrinsics of the camera. In this section we will describe how we calculate the optimal transformation matrix that transforms the labeling object onto its correct spot on the mesh.
	\subsection{Extracting depth values}\label{extractingdepth}
\begin{figure}[h]
		\caption{u,v image coordinate axes}
		\label{Figure:imagecoordinates}
		\centering
		\includegraphics[width=0.5\textwidth]{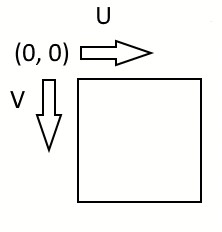}
	\end{figure}

\begin{figure}[h]
		\caption{Resulting matrix containing three columns, [u, v, z]}
		\label{Figure:uvz}
		\centering
		\includegraphics[width=0.75\textwidth]{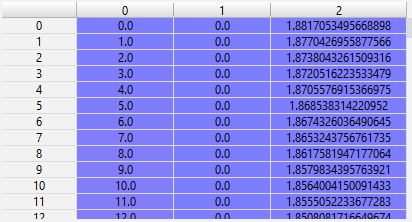}
	\end{figure}

	Decoding the received images is simply performing the reverse steps that were used to encode them on the Unity side.	First, to extract the RGBA values from the byte array sent from Unity, we use the Python library \textit{Pillow}'s Open function. Evidently, we will get a 3-dimensional array of shape heightxwidthx4. In other words, it is essentially a heightxwidth matrix with each entry being a 4 element vector representing the four values red, green, blue and alpha on a scale of [0, 255]. We divide all of the values by 255 reducing the range to [0,1]. Second, we perform the inner product of each entry with the vector $[1, \frac{1}{256}, \frac{1}{256^2}, \frac{1}{256^3}]$ which will result in the z value of each pixel on a scale from 0 to 1. Third, we multiply this matrix by $65$ (because we know the max depth that the Unity camera can calculate is 65) to receive the actual z value in world coordinates. Finally, we construct a $(width*height)X3$ matrix in which the first two columns are the $(u,v)$ coordinates of each pixel on the image starting from the top left and row by row, see Figure \ref{Figure:imagecoordinates}. The third column is the depth(z in meters) value of each pixel in world coordinates, see Figure \ref{Figure:uvz}.
	\subsection{Calculating World Space Coordinates}
	Since we know the depth coordinate of each pixel, we can use single view geometry to calculate each pixels' position in world space. The projection matrix, formed by multiplying the intrinsics by the extrinsics; see Equation \ref{projectionmatrix}, is used to transform points in world space into points in image space; see Equation \ref{projectt}.
	What we require is the opposite of the latter equation, we need to find the point's coordinates in world space knowing both it's coordinates in image space and its depth in world space. By left multiplying equation \ref{projectt} by the inverse of P we will get the following equation:
	\begin{equation}
		X = P^{-1}x
	\end{equation}
	We apply this equation on all of the points in the matrix we calculated in section \ref{extractingdepth}, then we will get the world coordinates of each pixel in the original images.
	\subsection{Reducing Mesh Points}
	Since the mesh will contain a lot more pixels than the labeling object, we should remove points that are very far from the object so as to increase the accuracy and speed of the registration algorithm. To do this, we first calculate the minimum and maximum coordinate of the labeling object on each axis. That would leave us with 6 values (3 minimums and 3 maximums). Each pair will serve as a bound on a certain axis. Because the labeling object will be in an awkward position, the bounds we choose should not be very strict. Therefore, we add/subtract a certain offset from the maximums/minimums. Choosing the correct offset value for all cases is impossible, with our testing we found that 0.15 (15 cm) yielded the best results. We then simply remove any points in the mesh that lie outside the bounds we created on each axis.
	\subsection{Registration}
	We have already covered registration extensively in a previous chapter. Our needs in this problem are different from the previous one though. In this problem there is a lot of outliers and noise. As we have previously said, the registration algorithms that are the most robust against these two artifacts are the stochastic registration methods. We have reviewed 4 state of the art stochastic methods.
	\subsubsection{Choosing the Best Algorithm}
	Even though it would seem like CPPSR \cite{7780570} would be the obvious choice since it leverages the possibility of using the pixel's colors to stregthen registration accuracy. Unfortunately, the reconstructions usually lose color accuracy while CAD objects retain a very accurate real world estimate of the colors. Henceforth, we in our case colors would be a detriment rather than a benefit. Additionally, we will not use IPDA \cite{7759602} since its main focus is registering a dense point cloud to a sparse one. However, since we are capturing the same resolution from the same camera, then the pixels will be uniformly distributed on both point sets and hence both would have the same \textit{denseness}. Although VBPSM \cite{7439870} achieves promising results, it's cubic time complexity prevents it from being used in any real time application. Finally, SVR \cite{7410845} presents itself as the best candidate as it is the best compromise between registration accuracy and computational complexity(O(mn)/where m and n are the number of components in GX and GY
	respectively).
	\subsubsection{Modification}
	We have used a modified version of SVR in which we use fast Gauss Transform to approximate the trained SVMs to a Gaussian Mixture Model instead of the one implemented by the authors, thus reducing the computational complexity to O(n+m).
	\subsubsection{Output}
	We provide the algorithm with the labeling object points as the source point set and the reduced mesh points as the target point set. As for the algorithm's hyperparameters, according to the authors changing them will barely affect the registration result and therefore we left the ones the authors used in their paper. Naturally, the algorithm will output the optimal rigid transformation matrix (containing only rotation and translation) that transforms the labeling object onto the mesh. This transformation matrix is sent to Unity via the same websockets connection to be applied on the labeling object.
	
	\section{Results}
\begin{figure}[!hbt]
		\centering
		\includegraphics[width=\linewidth]{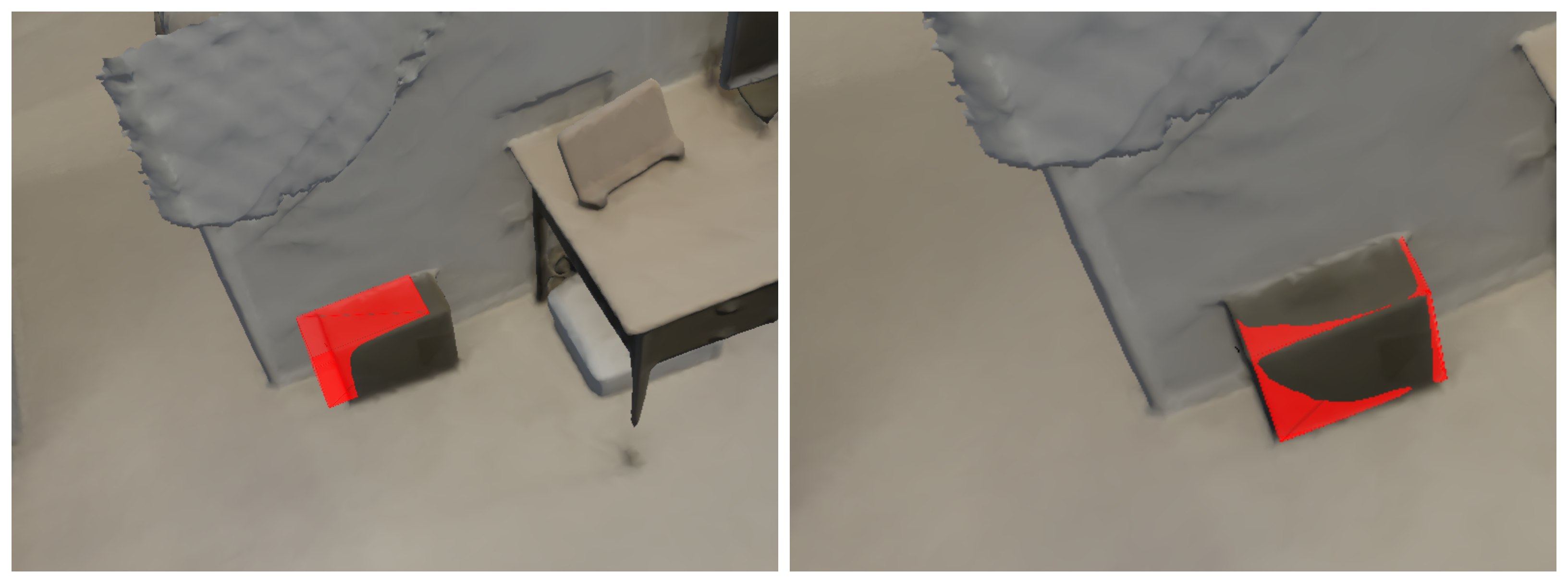}
		\caption{Snapping Example}
		\label{fig:snappingexample}
	\end{figure}

	On the unity side, a transformation matrix is received that defines the translation and rotation that will be applied on the labeling element to ensure the snapping into the correct position on the mesh. However, if we apply the transformation on the labeling object directly it will lead to wrong results because of the difference in the coordinate system. In other words, the center of rotation will be the center of the labeling element and not the center of the world. The transformation that should be applied is relative to the world space, to fix this issue we added an empty game object on the zero position (Unity's world space position) and made the labeling object a child of the created game object. Therefore, when applying the transformation on the new center of rotation, the labeling element will move accordingly to its parent (see Figure~\ref{fig:snappingexample}).
	
	\chapter{Labeling in Virtual Reality}
	\section{Introduction}
	In this chapter we introduce the concept of labeling in virtual reality as an added feature to the tool. The idea is to place the user in front of the mesh and allow labeling using VR controllers. This feature provides an immersive experience and exploits the 3D aspect of the tool to its maximum limits.
	
	\section{First Integration into Unity}\label{sec:firstint}
\begin{figure}[!hbt]
		\centering
		\includegraphics[width=0.5\linewidth]{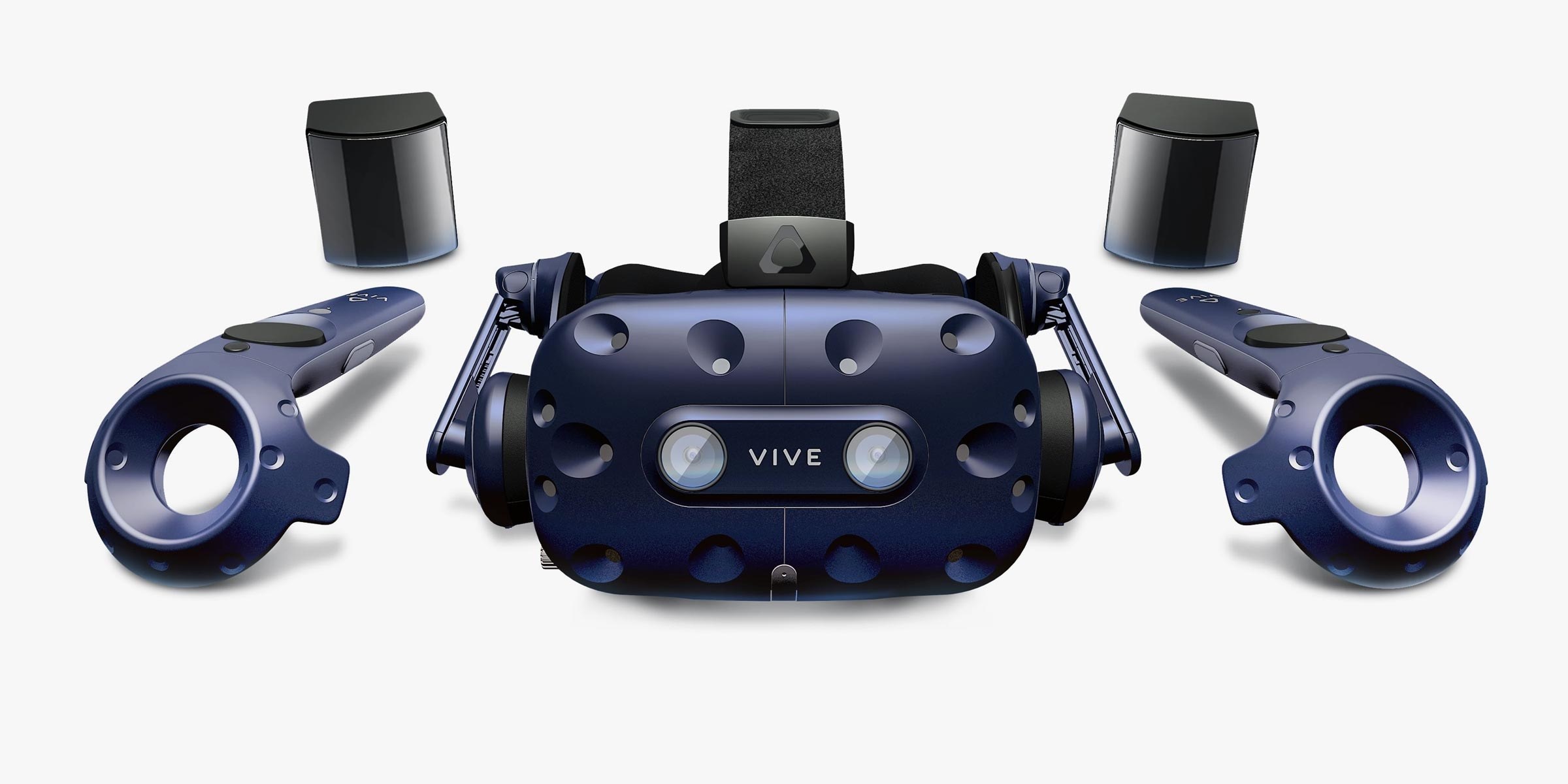}
		\caption{HTC Vive Pro}
		\label{fig:htcvivepro}
	\end{figure}

	The first objective of this integration was to install the \textit{SteamVR} module and add the \textit{OpenVR} SDK to Unity. \textit{SteamVR} is the software that will pair the VR headset and the controllers to the user's machine. Hence, the user must setup his platform and download \textit{SteamVR} as a prerequisite for the application to run in virtual reality mode. Note that during the development of this feature we used the \textit{HTC Vive Pro} hardware package (see Figure~\ref{fig:htcvivepro}). 
	
	\subsection{Teleportation in VR}
\begin{figure}[!hbt]
		\centering
		\includegraphics[width=\linewidth]{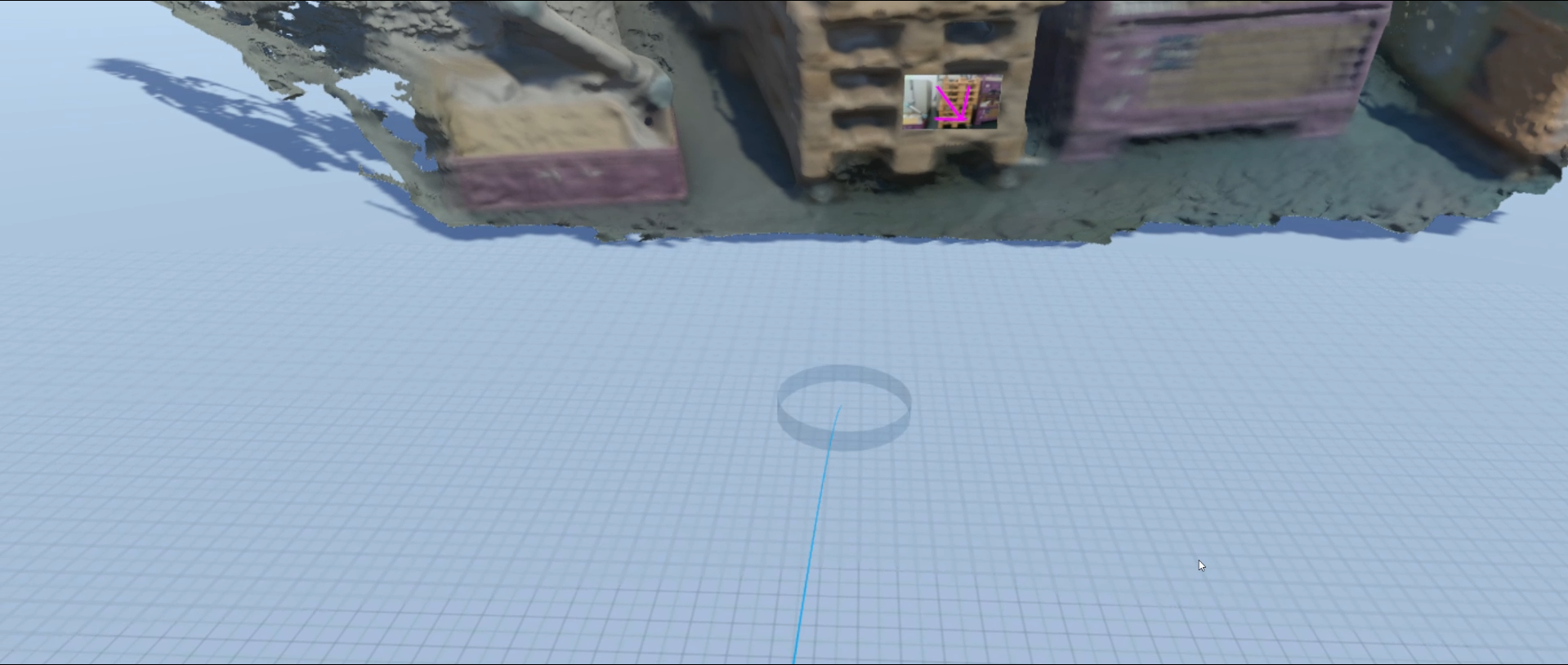}
		\caption{Teleportation Example}
		\label{fig:teleportation}
	\end{figure}

	To start developing the VR application, we imported demo scenes from the \textit{SteamVR} asset that got us familiar with the capabilities of the virtual reality features in Unity. One of these features is the teleportation that allows the user to change its position in the scene by clicking the corresponding button. It is represented by a line and a target spot on the floor that indicates the final position of the user (see Figure~\ref{fig:teleportation}). This functionality is useful because the headset is linked using a cable which limits the movement of the user.
	
	\subsection{Switching from 2D to VR}
	When the user clicks on the ``VR" button in the main window of the tool, all of the 2D user interface panels are switched off and a floor (plane) is added on the origin of the world (it is the grid shown in Figure~\ref{fig:teleportation}). However, if the user does not have the ``SteamVR" module then the virtual reality functionality is not activated.
	
	\subsection{Basic Object Manipulations Using \textit{SteamVR}}
	After setting up the mesh and the VR environment, we started to test simple object creation, deletion and manipulation that are included in the \textit{SteamVR} library.
	\subsubsection{Simple Object Creation}
\begin{figure}[!hbt]
		\begin{subfigure}[b]{0.4\textwidth}
			\centering
			\includegraphics[width=\linewidth]{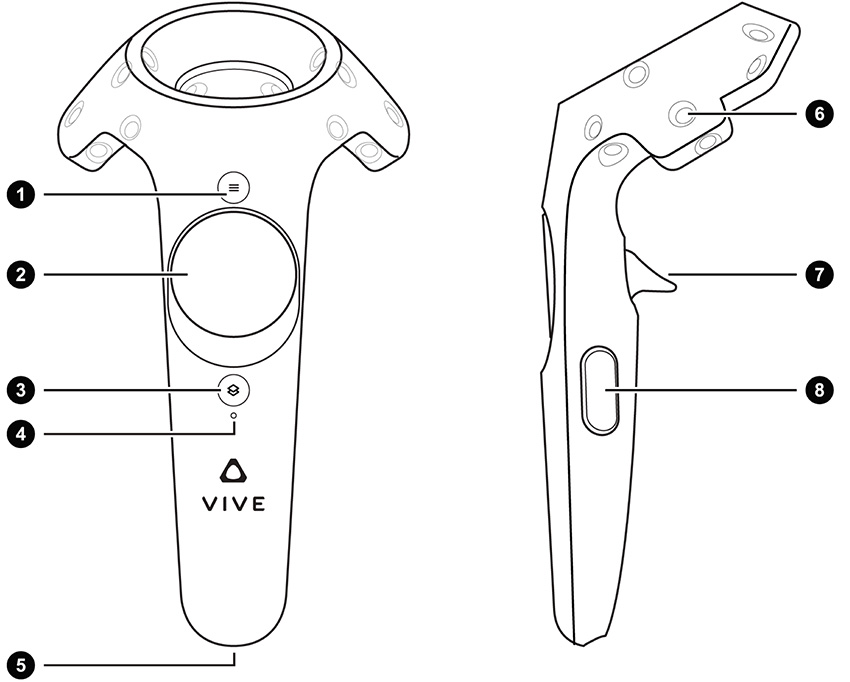}		
		\end{subfigure}
		\begin{subfigure}[b]{0.4\textwidth}
			\centering
			\includegraphics[scale=0.8]{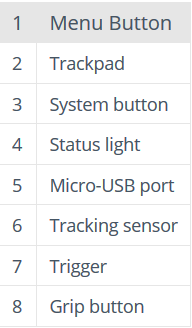}
		\end{subfigure}
		\centering
		\caption{HTC VIVE Controllers}
		\label{fig:controllers}
	\end{figure}

	This simple object creation is based on adding a cube in the middle of both controllers when the user clicks on their trigger button (see Figure~\ref{fig:controllers} for the \textit{HTC VIVE} controllers' key bindings terminology). Knowing the controllers' distance to each other in the scene, we calculated the center and added the object in that position.
	
	\subsubsection{Grabbing Objects}
\begin{figure}[!hbt]
		\centering
		\includegraphics[width=\linewidth]{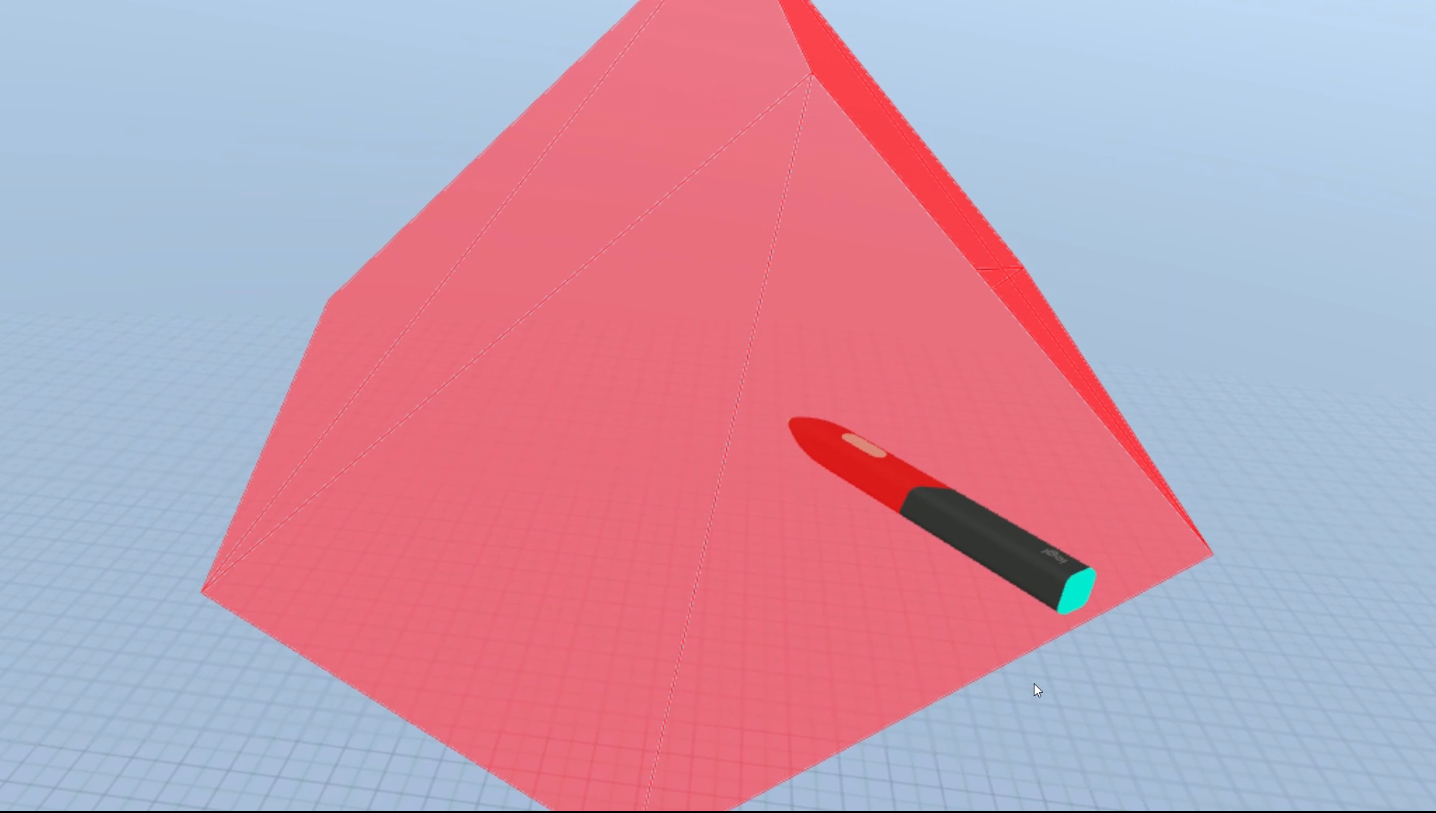}
		\caption{Grabbed Object Example}
		\label{fig:grabbingobjects}
	\end{figure}

	\textit{SteamVR} library provides also a useful functionality that could be added to any game object so it can be grabable whenever the controller and the object are colliding and the `Grab" button is clicked. The grabbed object becomes a child of the controller's game object and it will move relatively to the parent game object (see Figure~\ref{fig:grabbingobjects}). This is an essential feature in the use case of this tool because the user can now create the 3D object, grab it and place it on the mesh to label a certain object.

	\section{Collaboration with \textit{Logitech}}
	During the development of this feature in the tool, \textit{Logitech} contacted our supervisor at BMW Group and suggested to collaborate with them in the virtual reality part of the tool. They were launching a new product called \textit{VR Ink} stylus pen and they were willing to integrate it in the 3D labeling tool. 
	
	\subsection{Logitech's Stylus Pen}
\begin{figure}[!hbt]
		\begin{subfigure}[b]{0.4\textwidth}
			\centering
			\includegraphics[width=\linewidth]{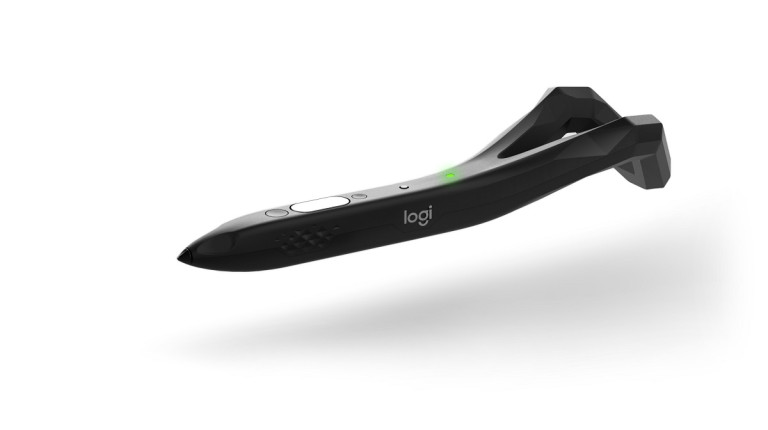}		
		\end{subfigure}
		\begin{subfigure}[b]{0.4\textwidth}
			\centering
			\includegraphics[width=\linewidth]{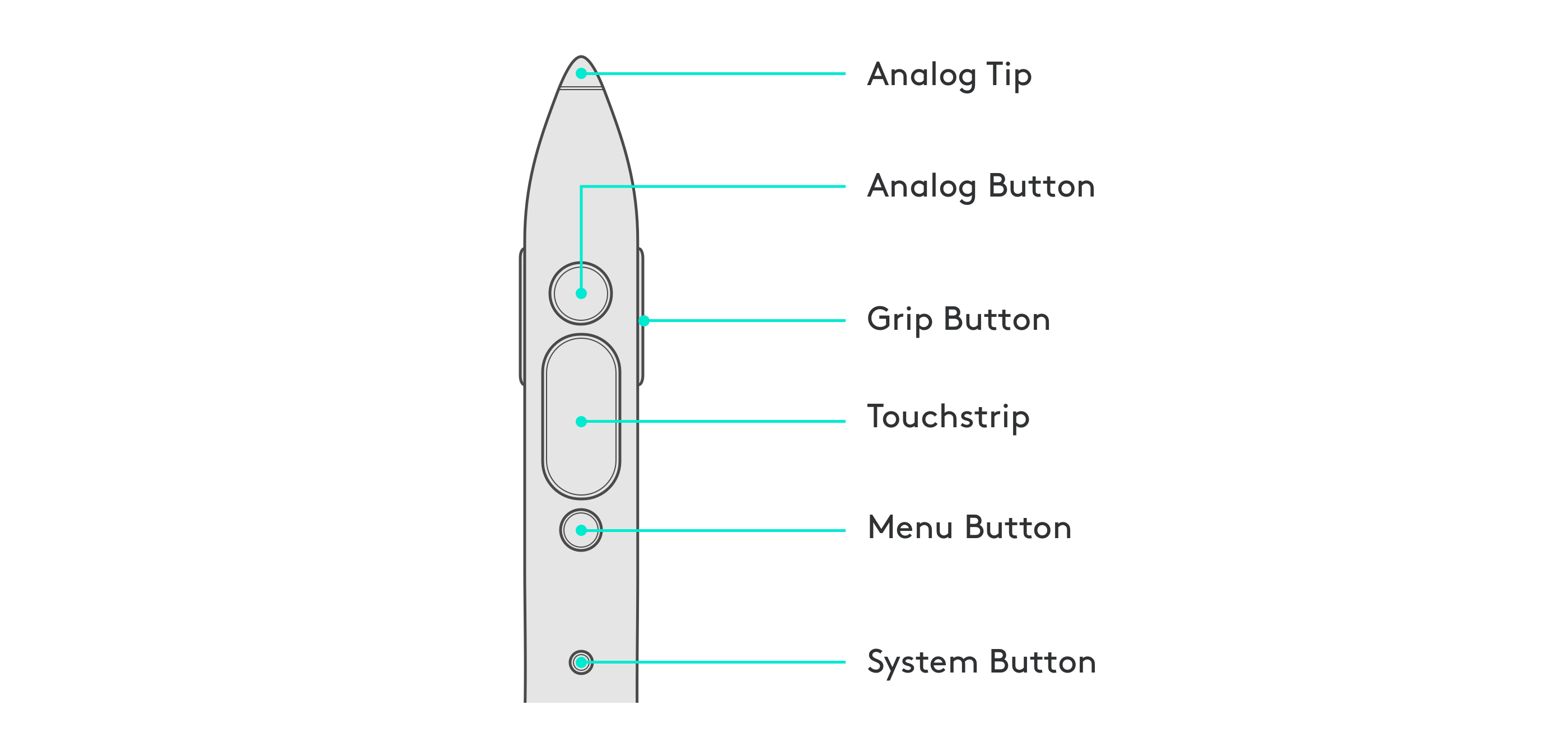}
		\end{subfigure}
		\centering
		\caption{Logitech Stylus Pen}
		\label{fig:styluspen}
	\end{figure}

	This pen is an input device for VR that allows the creation of controlled lines and drawing on virtual and physical surfaces. It has the same key bindings as the normal \textit{HTC Vive} controller but with an added analog tip that could be used with the analog button. For example, it could be used to adjust the thickness of a line when drawing (see Figure~\ref{fig:styluspen}).
	
	\subsection{Integration into the 3D Labeling Tool}
\begin{figure}[!hbt]
		\centering
		\includegraphics[width=\linewidth]{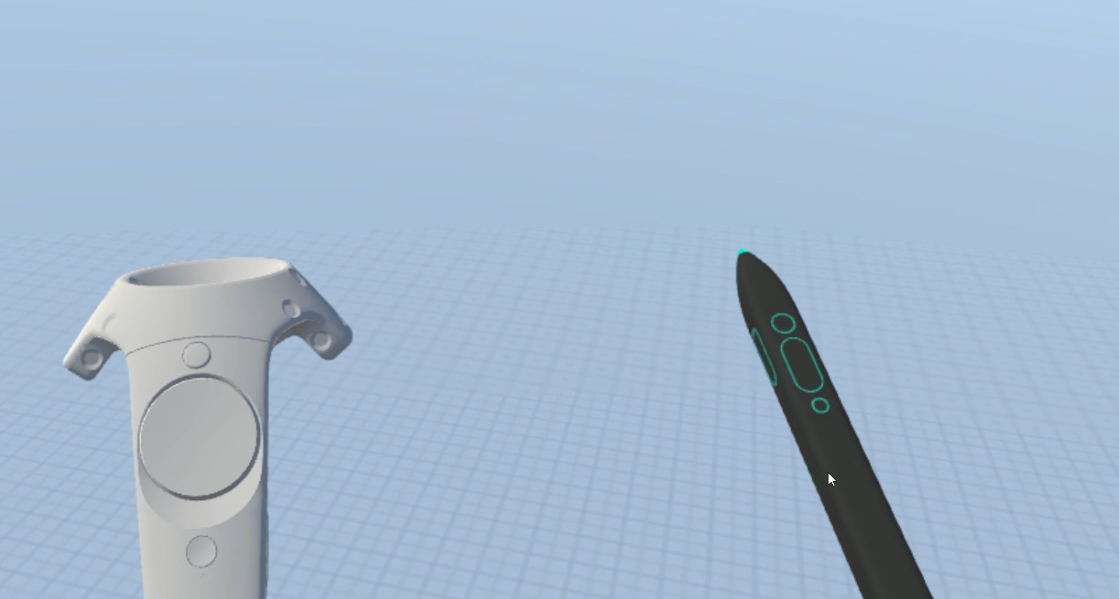}
		\caption{Stylus And HTC Models in VR}
		\label{fig:stylusandhtc}
	\end{figure}

	With the help of \textit{Logitech}'s team, we managed to integrate the corresponding game objects in our project that are responsible to enable the camera (The \textit{HTC VIVE} headset view) along with the model of both controllers (one of them is the \textit{HTC VIVE} controller and the other one is a model of the stylus as showing in Figure~\ref{fig:stylusandhtc}). They also developed some functionalities for cube creations that we later on upgraded to a 3D aspect that can be useful for the tool (Section~\ref{sec:labobjcreation}).\\
	Nevertheless, the user does not have to acquire this stylus pen to be able to use the VR option in the tool. The key bindings and the pairing mechanism is the same as a normal \textit{HTC VIVE} controller.\\
	
	\section{Added Features}
	In this section we are going to discuss in details the added features to \textit{Logitech}'s already implemented API that can benefit the user experience of the labeling process in VR.
	\subsection{User Interface}\label{sec:ui}
\begin{figure}[!hbt]
		\centering
		\includegraphics[width=\linewidth]{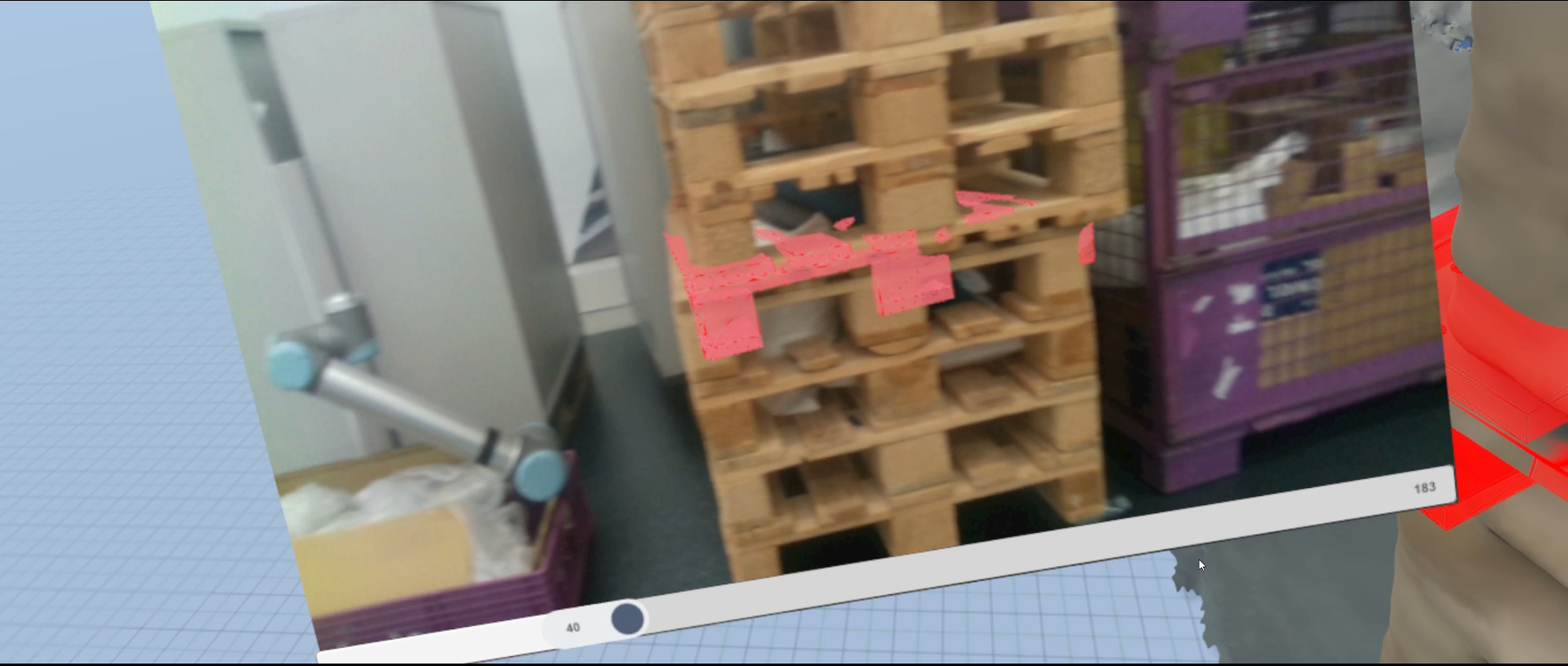}
		\caption{VR User Interface}
		\label{fig:charmouta}
	\end{figure}

	The user has to continuously check if the labeling objects that they are placing on the mesh are well projected to the images. Therefore, we added a user interface that consists of a series of panels that pops up on the non-dominant controller (see Figure~\ref{fig:charmouta}). To implement this canvas, we have added it as a child game object on the controller after aligning it in a way that is convenient for the user (size and location).\\
	Moreover, this canvas includes the RGB images to be labeled along with the slider filter to go through the images, the CAD model inventory and a settings menu to enable or disable world scale (Section~\ref{sec:worldscale}). In addition, for the user to enable this panel, they have to click on the "Menu Button" on the non-dominant hand, which is also the same button to switch between menus.\\
	\textit{Logitech} already implemented a raycaster program that generates a ray from the pen to a \textit{UI Element} in the scene. Thus, it can be used to trigger the buttons or the images from a distance without having the need to make the pen touch the user interface elements.\\
	
	\subsection{Labeling Objects Creation}\label{sec:labobjcreation}
	We gave the user 3 options to create a 3D object in the world:
	\subsubsection{Drawing a Cuboid}
\begin{figure}[!hbt]
		\centering
		\includegraphics[scale=0.8]{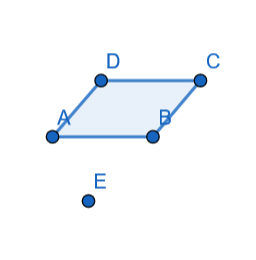}
		\caption{Plane Projection to Add Depth Example}
		\label{fig:projectionexample}
	\end{figure}

\begin{figure}[!hbt]
		\centering
		\includegraphics[width=\linewidth]{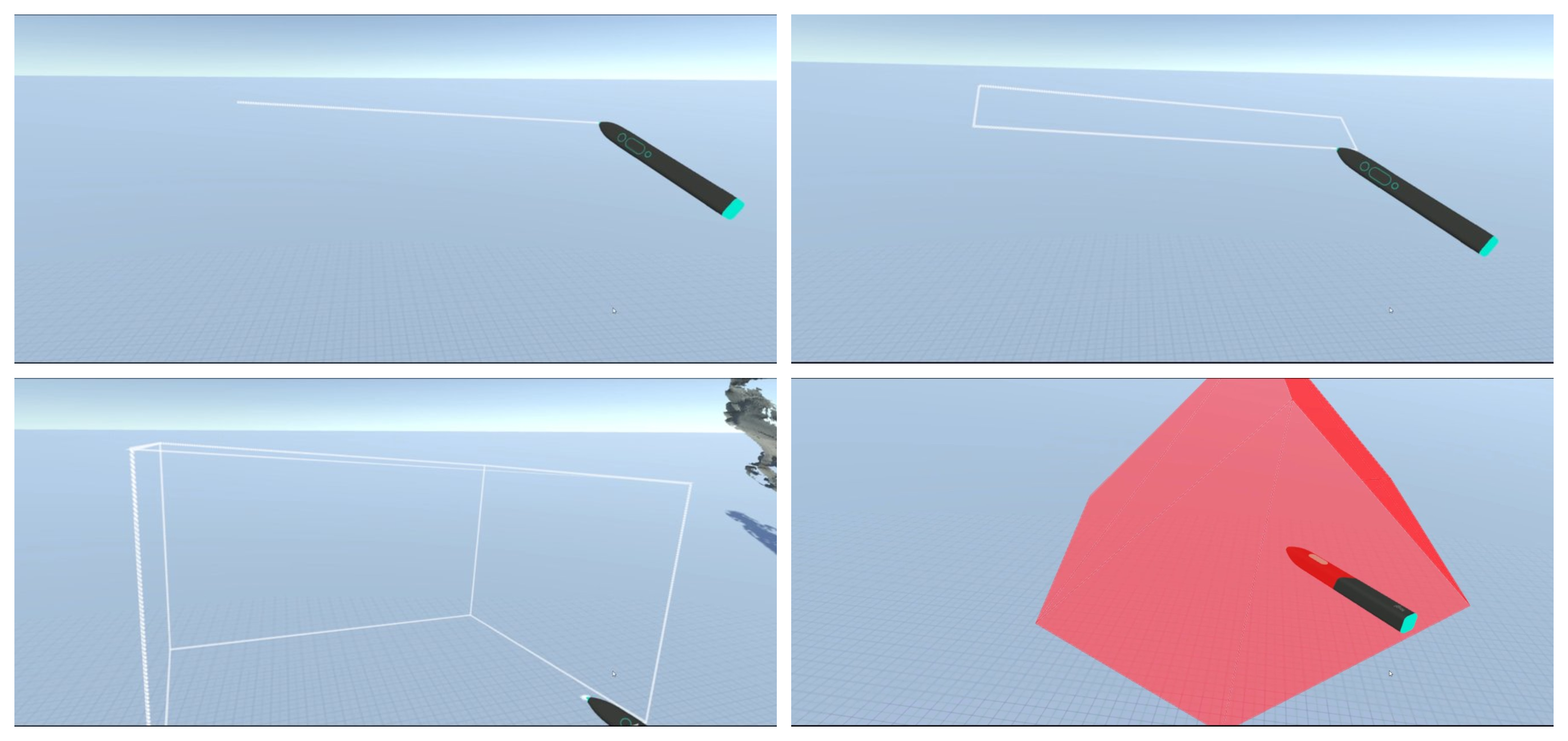}
		\caption{Drawing a Cuboid Example}
		\label{fig:cuboiddrawing}
	\end{figure}

	For the user to draw a cuboid using the tracker, they have to choose 2 points and draw a line, then drag the controller to create a rectangle and finally drag again to add depth for the rectangle so it becomes a cuboid (see Figure~\ref{fig:cuboiddrawing}).\\
	The idea behind this process is to project the position of the controller on the planes created by the first two points and their normal vectors passing by these points. As a result of these 2 projections, we obtain 2 points that will finish drawing the rectangle as shown in the second photo in Figure~\ref{fig:cuboiddrawing}.\\
	Moreover, adding the depth was based on the same idea of projecting to a plane but we had to do it twice. For example, in Figure~\ref{fig:projectionexample} we have the rectangle already drawn and we want to add the depth based on the trackers position which is the point E. First, we project E to the plane formed by A and the vector AB, then the result of this projection is then projected to the plane formed by B and the vector BC. To complete the drawing, this process is done 4 times for the remaining planes and points.\\
	To create the labeling object we used the first 3 points calculated using the projection to the plane and a fourth point that represents the depth of the cuboid. Furthermore, using the vectors composed by these 4 points, we managed to calculate the position, rotation and scale of the cuboid.\\
	In conclusion, this method enhances the user experience by giving the user the opportunity to draw a custom cuboid that suits the object that they want to label in the mesh.
	
	\subsubsection{Convex Hull to Cuboid}
\begin{figure}[!hbt]
		\centering
		\includegraphics[width=\linewidth]{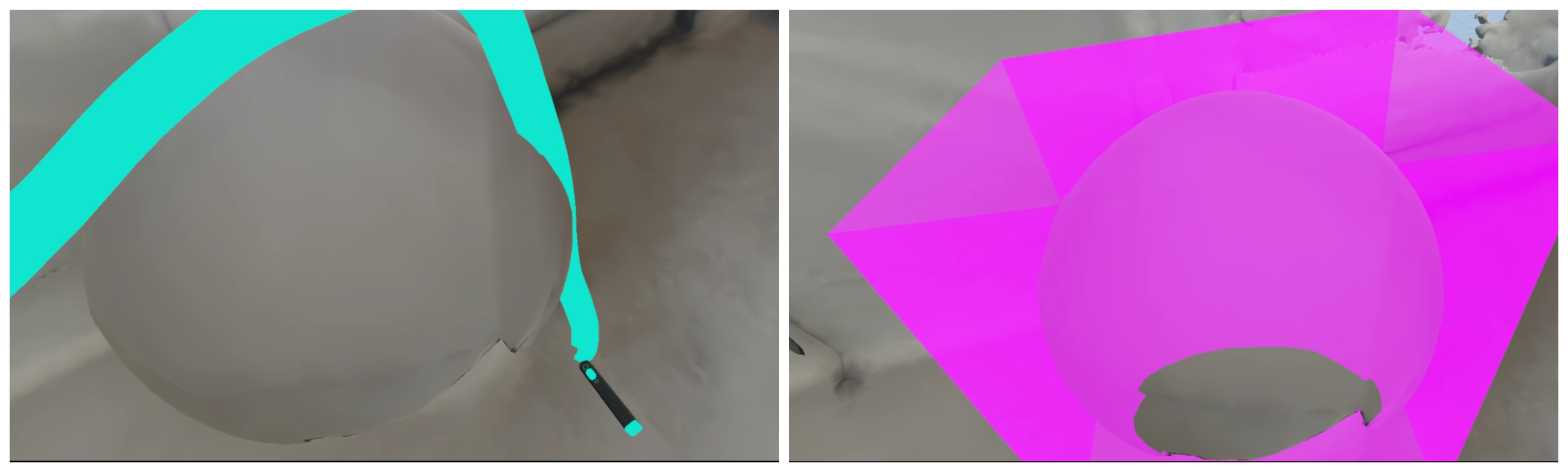}
		\caption{Convex Hull to Cuboid Example}
		\label{fig:convexhullexp}
	\end{figure}

\begin{figure}[!hbt]
		\centering
		\includegraphics[width=\linewidth]{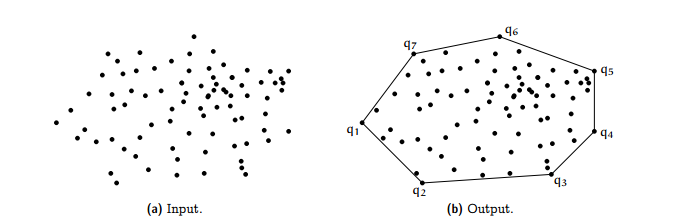}
		\caption{Convex Hull Example}
		\label{fig:convexhull}
	\end{figure}

	In this method, the user presses a button on the controller and starts painting a line as if they were scratching on the mesh the object to be labeled. As a result, they obtain the minimum box that envelopes the created line (see Figure~\ref{fig:convexhullexp}).\\
	This feature is based on the concept of \textit{Convex Hull} which is an approach in mathematics used to envelope a set of points in the Euclidean space in the smallest convex shape (see Figure~\ref{fig:convexhull}). To implement it in Unity we used~\cite{convexhull} that already has implemented the method to produce a convex hull game object from an array of points.\\
	For the user experience to be intuitive we decided to let the user draw a line and not place points in the scene. We used the \textit{Line Renderer} Unity class to show the line on the screen yet extract the points from it and assign it to the convex hull algorithm.\\
	To transform the convex shape obtained from the previous step to a cuboid, we decided to take advantage of Unity's box collider properties. When this collider is assigned to a game object, it automatically creates the minimum bounding box around it. Therefore, the next step was to add the box collider on the convex hull object and retrieve its vertices that will help produce the cuboid using the ``Mesh" class in unity.\\
	Finally, after creating the box and rendering it to the scene (using the \textit{Mesh Renderer} component), we made sure that it has the same transformation as the box collider that encapsulates the convex hull. Then, we delete the convex hull game object and keep the box as a labeling element.

	\subsubsection{CAD Model Importer}
\begin{figure}[!hbt]
		\centering
		\includegraphics[width=\linewidth]{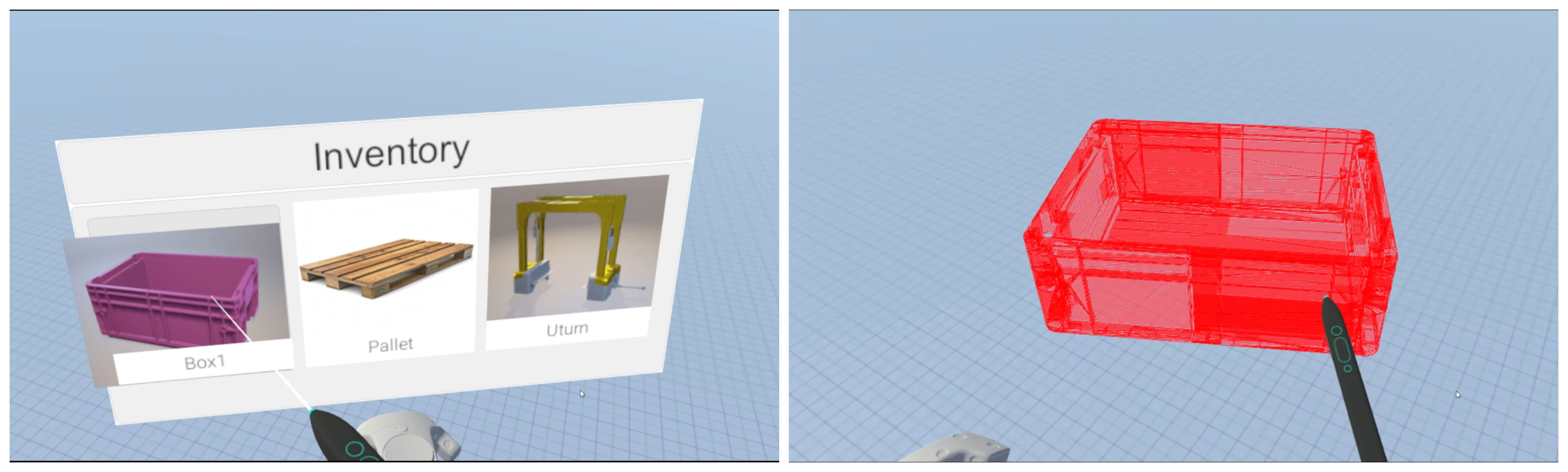}
		\caption{Cad Objects in VR}
		\label{fig:cadobjvr}
	\end{figure}

	This feature is based on the same one that is used in the 2D mode of the tool. However, we added a CAD model inventory panel in the user interface part of this mode to give the user the option to load a CAD object into the environment when labeling in VR (see Figure~\ref{fig:cadobjvr}).
	
	\subsection{Objects Manipulation}\label{sec:objectsmanipulation}
	After creating the labeling element, the user wants to manipulate it: change its position, copy, delete and scale. As already mentioned, \textit{Logitech} uses the same grabbing mechanism as the one of \textit{SteamVR} which is responsible for changing the position and rotation of the labeling element in relative to the movement of the dominant controller.\\
	To copy the labeling element, we duplicate the object that the controller is colliding with by instantiating the collided game object. In addition, to delete a labeling element, the user must be pointing or colliding with the object and press the \textit{Trigger} button. Note that when deleting the object it does not mean to only destroy the game object but also we had to remove it from the list that contains all the other labeling objects.\\
	Furthermore, we added the feature of grabbing the object from afar using a raycaster program that is issued from the dominant hand controller in case it is pointing at a labeling element. This implementation caused some issues when trying to use it with the user interface raycaster program (see Section~\ref{sec:ui}). Hence, to solve that problem, we managed to synchronize both of the raycaster programs using boolean variables and giving the priority for the user interface raycaster program in case there's an object behind the panel.\\
	Finally, the user has the ability scale the labeling element if they are colliding with it and pressing on the non dominant hand trigger button. The object will be scaled along the distance between the two controllers. In other words, the idea behind this type of scaling is to transform the controllers game object along with the labeling object to the camera's coordinate system. Then, we calculate the difference (Vector) between both of the controllers and transform it to the object's coordinate system. This allows the user to scale the object on its axis regardless of its rotation relative to the user's view. In addition, to be able to determine the axis on which the object has to scale, we calculate the dominant axis; in other terms, we check the position of the controllers and see if it is dominant on the x, y or z axis. For example: if the user's controllers are on top of each other than the dominant axis is the y axis and the object should be scaled on that axis while taking into account the distance between the controllers.
	\begin{figure}[!hbt]
		\centering
		\includegraphics[scale=0.5]{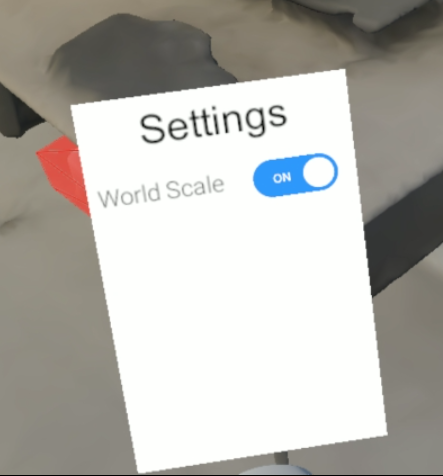}
		\caption{World Scale Setting in UI}
		\label{fig:worldscale}
	\end{figure}
	\subsection{World Scale}\label{sec:worldscale}
	The last functionality that we implemented in VR was the ability to scale the entire scene/world. This came as a solution for an issue that we faced when we were trying to label meshes that are too high or too large for the user to physically reach.\\
	This solution consists of toggling the \textit{World Scale} setting in the user interface panel (~\ref{fig:worldscale}), which adds an empty game object that parents all the labeling objects available along with the mesh. Therefore, when the \textit{Grab Button} is triggered on the controllers, the \textit{World} game object is transformed along them. Then, the mesh and the labeling objects are transformed accordingly. Note that the scaling in this case is uniform and it differs from the scaling that was discussed in Section~\ref{sec:objectsmanipulation}. In addition, when the world scale option is enabled, we made sure that the copy and object creation is done according to the world game object. In other words, these objects are added directly as children of the world object. Nevertheless, this will not disrupt the labeling process, because when the world scale option is off or the user turns off VR mode and gets back to the 2D mode, all of the features of that option are reverted and the labeling objects will preserve their alignment with respect to the mesh.

	\chapter{Experimental Results}
	In the final phase of this project, we conducted a series of experiments to test the run time efficiency of the tool, compared to the other state of the art labeling tools presented in Chapter 2.\\
	These tests were performed on Windows 10 Home Edition running on a personal computer hosting an NVIDIA GTX 1060 GPU, 16GB of RAM and a 6 cores 2.2 GHz i7 CPU.\\
	The experiments were performed in two different ways. First, we compare the labeling tools (see Chapter 2) on three distinct data sets: (1) shots taken by a robotic arm of a box containing door handles (92 images), (2) shots of a kitchen (98 images) and (3) images of pallets, trollies and boxes in a factory (184 images). Second, we compared the methods based on the number of objects to be labeled in a data set. Note that the timer starts when the labeling process begins and stops when the annotations are exported.
	\section{Experiment 1}
\begin{figure}[h]
		\centering
		\includegraphics[width=\linewidth]{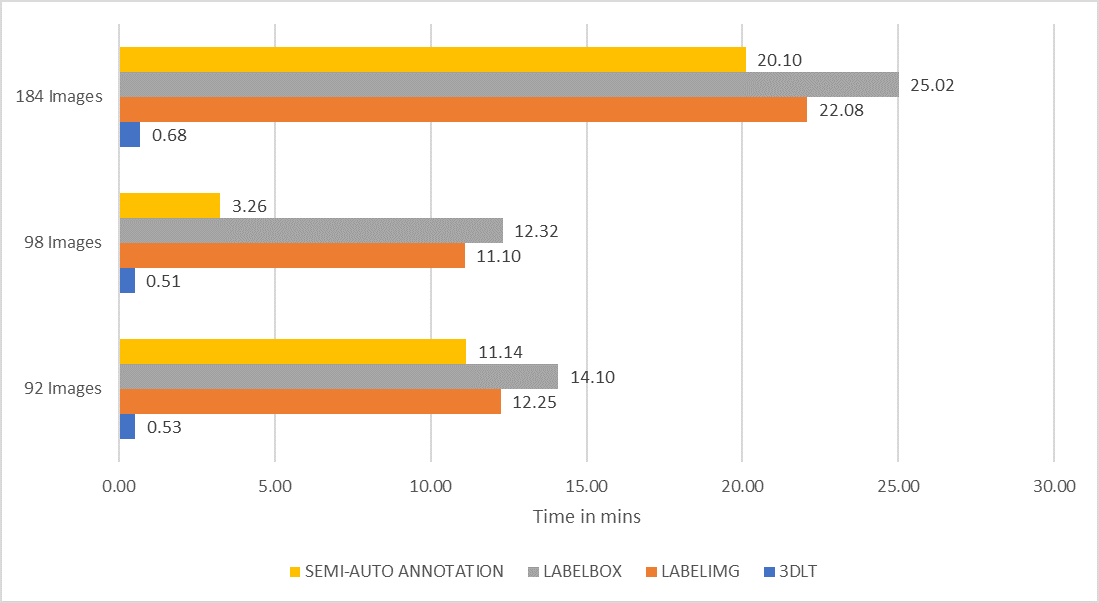}
		\caption{Comparison on Different Data Sets}
		\label{fig:comparison1}
	\end{figure}

	The first chosen labeling tool is an offline application called \textit{LabelImg}~\cite{labelimg}. It achieves 2D labeling by drawing a rectangle on each imported image and generates the annotations in XML format. As shown in Figure~\ref{fig:comparison1}, the data set that contains 92 images took more time to label than the one with 98 images. This is correlated with the position of the object in the images when drawing the bounding box around it (object on the image's edges or behind other objects). Also, when the number of images increased to 184, the time to finish the labeling almost doubled from 12.25 to 22.08 minutes.\\
	The second tool that was chosen is the \textit{LabelBox}~\cite{labelbox} web application on which the 2D labeling process is done in the same fashion as \textit{LabelImg} but online rather than on premise. Figure~\ref{fig:comparison1} displays how this method struggles in run time comparing to the previously mentioned tool. It is caused by the delays in the server's response time when the annotations are drawn on the image. Therefore, this method is not reliable because its performance depends on the user's activity and the network's congestion when annotating the images.\\
	The third and final tool that was mentioned in Chapter 2 is the \textit{Semi-Auto Annotation Tool}~\cite{annomage} that basically enhances the idea of the previous 2D labeling tools. It adds the ability to detect the object before drawing the 2D bounding box on it accordingly. However, as shown in Figure~\ref{fig:comparison1}, this method only shows its best performance on the kitchen data set because it is already trained to detect a fridge which is present in this data set. As a result, the fridge in each image is detected and annotated directly by a bounding box, which made the labeling faster (3.26 minutes). However, the other two data sets do not hold any object that is present in the pre-trained model of this tool, hence resolving to approximately the same run-time as the previous tools by drawing the rectangle on the objects manually for each image. 
	\newpage
	\section{Experiment 2}
\begin{figure}[h]
		\centering
		\includegraphics[width=\linewidth]{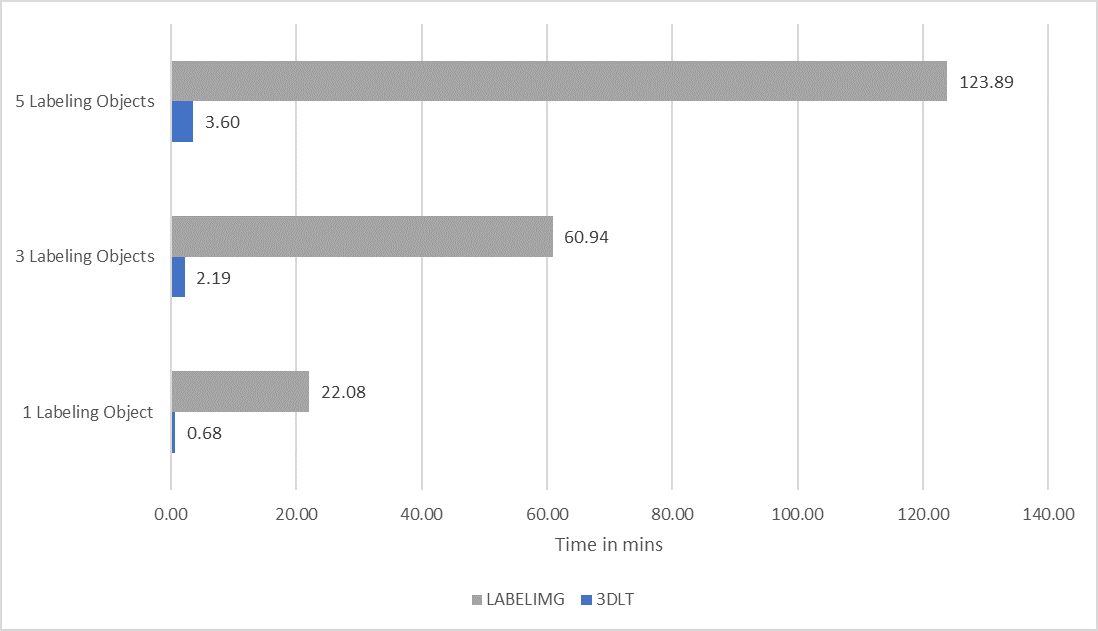}
		\caption{Comparison on Different Number of Objects}
		\label{fig:comparison2}
	\end{figure}

	This experiment is performed on the robot arm data set in which we compared the 3DLT with the \textit{LabelImg tool}. As you can see in Figure~\ref{fig:comparison2}, the same data set is imported 3 times, where we increased the number of objects to be labeled in an image at each distinct import. The 2D labeling tool's time to label showed an exponential increase in time when expanding the number of objects because the user has to draw a 2D bounding box on each one in every image.
	\newpage
	\section{Discussion}
	When comparing any labeling tool mentioned in Chapter 2 to our tool, we notice that in the first experiment, the time taken to label any of the three different data sets in the 3DLT is the same constant time for each data set, and is highly faster than any of the others (see Figure~\ref{fig:comparison1}) in each of the tests. It is mainly because the act of placing one object on the mesh is not affected by the number of images in the data set. In contrast, the run time of the other tools depends on the number of images because the user annotates each image to generate the ground truth labels.\\
	In the second experiment, the act of placing multiple objects in our 3D labeling tool and generating the annotations is almost 34 times faster than a 2D annotation software (see Figure~\ref{fig:comparison2}). As a result, this shows how time consuming the 2D labeling process is, and how the 3D Labeling tool that we developed is much more efficient for any data set or any type of objects to be labeled.

	\chapter{Conclusion}
	In this report we introduced a novel labeling tool, the 3D Labeling Tool (3DLT), that greatly reduces time spent on preparing training data for object detection machine learning models. The tool's user interface is easy to use and intuitive. It is a cross platform tool that can run on Windows, MacOS and Linux that has a client-server architecture with a Unity Client and a Python server.\\
	We used the recorded RGBD images in ORB-SLAM to compute the camera trajectory and in Open3D to extract a 3D triangular mesh. We also defined a special data format to standardize our data preparation process, the IRIS format. Moreover, we showed how any raw point cloud can be labeled using 3DLT. We compared many state of the art surface reconstruction techniques and selected the Screened Poisson Reconstruction. We surpassed hardware limitations set on reconstruction resolution by creating a workflow that splits the point cloud into multiple pieces, reconstructs each piece separately and merges them all back together. We containerized the aforementioned reconstruction workflow in Docker to make it portable and devoid of requirements. The reconstructed mesh can then be imported into a Unity application that we developed, Surface Sampler (SS). This application allows the user to sample data from the mesh by navigating around it and taking RGBD images. The collected data is exported into the IRIS format and can therefore be loaded in 3DLT.\\
	To calculate the 2D bounding rectangle, we projected the 3D labeling object to its equivalent position on the 2D images using the camera's projection matrices. We then used Python to calculate the smallest possible axis-aligned bounding rectangle that contains the projected labeling object. In the cases where a certain part of the labeling object was blocked by another, we carefully selected the unsupervised outlier detection algorithm, rPCA, to allow for the computation of smarter bounding rectangles. This algorithm has linear time complexity allowing it to preserve the tool's ``real-time" property.\\
	Since manually defining the position of a labeling object in 3D is non trivial, we simplified the process by allowing the user to select corresponding points on the labeling object and the mesh. Then the tool calculates the optimal transformation parameters that map the former point set to the latter. We compared and categorized many nonrigid registration algorithms, ultimately selecting TPS-RPM. Additionally, we parallelized it to perfect correspondence calculation and restrict its transformation parameters to 9 degrees of freedom.\\
	In the cases where the user knew the size of the object to be labeled, we introduced two ways to simplify the process. First, we gave the user the ability to select the labeling object's corresponding points on the 2D image instead of the mesh. We do this by reducing the computation of the desired labeling object position to a system of non linear equations. We solve it by using the DOG\_BOX method's output as a starting estimate to the Newton-Raphson method. Allowing the user to label using the images broadens the tools capabilities and removes the requirement of a mesh in the input data. Second, we implemented a feature that allows the user to snap the labeling object onto the desired object as long as it is relatively close to it. We use Unity's shader program to apply post-processing effects on the camera's renders in order to retrieve their depth map as a 2D texture and decrease the frequency of outliers. The rendered texture is encoded and sent as point sets to the Python server. After that, we use Support Vector Registration to find the optimal rigid registration between these point sets.\\
	Moreover, we introduced Virtual Reality as a practical add-on to the 3D labeling tool. It allows the user to label any importable data set in VR using the appropriate controllers with intuitive mechanisms for creating and manipulating labeling elements using Convex Hulls and simple geometry concepts.\\
	Finally, we performed a series of experiments to compare our tool against current state of the art methods. The experiments were performed in order to highlight how tedious it is to label thousands of images by drawing 2D bounding boxes. The performance	of the tools was benchmarked based on both the number of images in each data set and the number of objects to be labeled in a single data set. The 3D labeling tool excelled in all experiments displaying superior results compared to the 2D labeling tools.\\
	Even though the tool can be built on the most popular operating systems, we think developing a WebGL version and abstracting all heavy computation to the server level is the next logical step.	Even though this is possible in Unity, we need to redesign the architecture of the tool to handle importing data from the user's machine onto the WebGL application in an efficient manner. Additionally, another area of possible improvement is taking cues from Anno-Mage and allowing the tool to predict bounding boxes for pre-trained object classes. Contrastingly to Anno-Mage, our tool would predict 3D bounding cubes instead of 2D rectangles. Therefore we cannot use the established RetinaNet as a suggestion engine. However, PointNet \cite{Charles2017} has recently revolutionized deep learning on point sets for 3D classification and we think it is worth exploring as a potential suggestion engine.

\clearpage

\printbibliography

\end{document}